%% file: acl_latex.tex
\documentclass[11pt]{article}

\usepackage[final]{acl}

\usepackage{times}
\usepackage{latexsym}

\usepackage[T1]{fontenc}

\usepackage[utf8]{inputenc}

\usepackage{microtype}

\usepackage{inconsolata}

\usepackage{graphicx}
\usepackage{booktabs}
\usepackage{adjustbox}
\usepackage{array}
\usepackage{subcaption}
\usepackage{amssymb} 
\usepackage{pifont}  
\usepackage{multirow}
\usepackage{makecell}
\usepackage{hyperref}
\usepackage{svg}
\usepackage[most]{tcolorbox}
\usepackage{booktabs}
\usepackage{multirow}
\usepackage{graphicx}
\usepackage[table]{xcolor}
\title{Characterizing Web Search by Conversational LLM Agents:\\ From Search Decisions and Strategies to Results and Responses}

\author{
  \textbf{Mahsa Amani\textsuperscript{1,3}},
  \textbf{Seungeon Lee\textsuperscript{1}},
  \textbf{Abhisek Dash\textsuperscript{1}},
  \textbf{Asmaa El Fraihi\textsuperscript{4,*}},
\\
  \textbf{Yunah Jang\textsuperscript{6,*}},
  \textbf{Elisabeth Kirsten\textsuperscript{2}},
  \textbf{Qinyuan Wu\textsuperscript{1,3}},
  \textbf{Krishna P. Gummadi\textsuperscript{1}},
\\
  \textbf{Manish Gupta\textsuperscript{5}},
  \textbf{Abhilasha Ravichander\textsuperscript{1}},
  \textbf{Muhammad Bilal Zafar\textsuperscript{2}},
  \textbf{Soumi Das\textsuperscript{1}}
\\
\\
  \textsuperscript{1}Max Planck Institute for Software Systems,
  \textsuperscript{2}Ruhr University Bochum,
  \textsuperscript{3}Saarland University,
\\
  \textsuperscript{4}Laboratoire d'Informatique de l'École Polytechnique,
  \textsuperscript{5}Microsoft,
  \textsuperscript{6}Seoul National University
\\
  \textsuperscript{*}\small Work conducted during an internship at the Max Planck Institute for Software Systems.
}

\newcommand{\dataset}[1]{\textsc{InVivoGPT}}
\newcommand{\replay}[1]{Replay}

\newcommand{\invivo}{{\tt invivo}}
\newcommand{\invitro}{{\tt invitro}}
\begin{document}

\maketitle


\begin{abstract}
Conversational LLM agents increasingly rely on Web search, yet the end-to-end lifecycle of agentic search remains poorly understood. We present the first study of Web search across four major conversational platforms (ChatGPT, Claude, Grok, and DeepSeek), combining real-world user interactions (\invivo{}) with controlled experiments using the same platform's models by their APIs (\invitro{}). We investigate the quality of agentic decisions to invoke Web search, their strategies to formulate queries, the potential domain preferences in the search results they receive, and the choices they make when transforming search results into grounded responses. 
We find that Web-search decisions vary substantially across platforms and models, while more frequent Web-search invocation does not necessarily yield better response quality. We further show that conversational agents employ different complex querying strategies and that platform specific search engines return search results from their preferred domains. Finally, although responses are largely grounded in search results, some claims rely on uncited search results, raising concerns about attribution and reliability. Our findings have important implications for the design of future AI agents and Web search tools optimized for conversational retrieval.
\end{abstract}


\input{introduction}
\input{related}
\input{sec-setup-revise}

\input{sec1-revise}

\input{sec2-revise}

\input{sec3-revise}
\input{conclusion}
\input{storyline}
\input{limitations}
\input{ethics.tex}
\bibliography{custom}
\newpage
\appendix
\input{appendix}

\clearpage





\end{document}

%% file: introduction.tex
\section{Introduction}
\label{Sec: Intro}

\begin{figure}
    \vspace{-4mm}
    \centering
    \includegraphics[width=1\columnwidth]{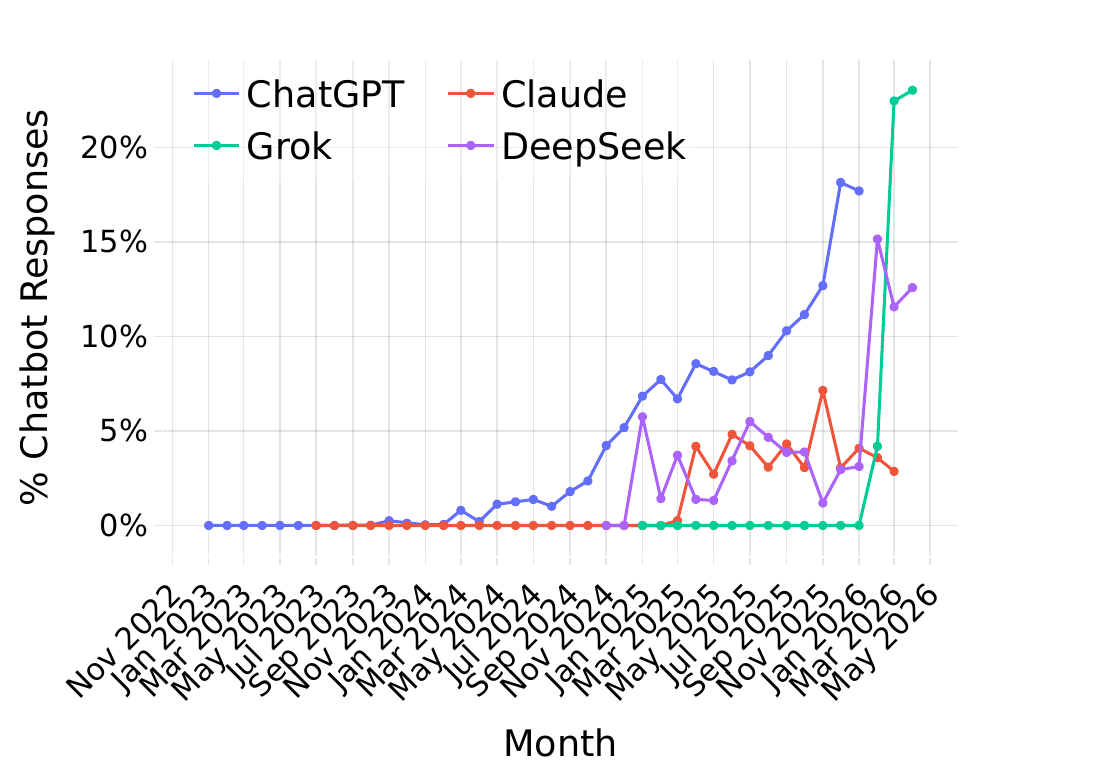}\vspace{-2mm}
    \caption{Datasets from four popular chatbot platforms analyzed in this work show a consistent trend of increasing use of Web search over time. Appendix Figures \ref{fig:tooly_turns_other_platforms} and \ref{fig:web_call_trend_by_model_other_platforms} further show that search dominates all tool calling and that newer reasoning-oriented models invoke search at higher rates, across all platforms.}
    \label{fig:num_web_call_over_time}
    \vspace{-4mm}
\end{figure}

The ability of agents powered by large language models (LLMs) to autonomously search the Web is fundamentally altering how we access online information~\citep{chatterji2025people}.
Today, millions of users rely on conversational AI platforms like ChatGPT~\cite{openai_chatgpt}, Claude~\cite{anthropic_claude}, Grok~\cite{xai_grok}, and DeepSeek~\cite{deepseek_chat} as Web search assistants, trusting them to navigate the Web and generate accurate answers. 
Figure~\ref{fig:num_web_call_over_time}, plotted using datasets analyzed in this paper, shows how a growing fraction of all chatbot responses by ChatGPT, Claude, Grok, and DeepSeek involve calling Web search tools. 
Yet, the life cycle of agentic search -- from the user prompt to the final generated response -- is neither well studied nor understood.

Web search is a complex cognitive task even for human users. 
Users first need to perceive their knowledge deficit, then reason how to query the search tool -- including issuing multiple queries both simultaneously and iteratively expanding the queries based on prior tool responses, and finally, update their prior knowledge using the retrieved information.
The life cycle of agentic search (see Figure~\ref{fig:overview}) similarly begins with the model recognizing and deciding that its own parametric (internal) knowledge is not sufficient to answer a user's prompt.
The next stage involves querying a search tool, both in parallel and in sequence, reformulating the queries based on retrieved results.  
The final stage involves synthesizing the search results from multiple (potentially contradicting) sources with the model's own parametric knowledge to generate a coherent response. 
Accordingly, our investigation here is motivated by the following three high-level research questions: \\
(1) {\it How do agents decide when to call Web search?} \\
(2) {\it How do agents query Web search tools?} and \\
(3) {\it How do agents use search results to synthesize responses?}  

To conduct our study, we adopted the data collection strategy described in \citep{karnam2026bowling} and collected conversational data donated by real-world users of four chatbot platforms, namely ChatGPT, Claude, Grok, and DeepSeek.
To our knowledge, our work is the first to investigate Web search by multiple chatbots \invivo{}, i.e., when users are interacting with them for their normal, daily conversations.
We carefully designed an experimental setup, where we query using selected user conversations and compare the performance of different model families \invitro\, i.e., we study how the LLM agent responses compare and change, when accessed via their platform APIs.     
To our knowledge, our work is the first to systematically study how the choice of model family impacts Web search.
%

We summarize our key findings below:\\
\noindent {\bf 1.} {\it On decisions to call Web search:} LLM agents differ significantly in when they chose to call Web search. While LLM agents are good at deciding when they do not need Web search, their decisions to call Web search do not always improve their responses. 

\noindent {\bf 2.} {\it On querying strategies and search results:} LLM agents employ differing complex querying strategies, issuing multiple queries, often iteratively reformulating them using information from retrieved sources and verbose user conversational history. The platform-specific search engines that LLM agents query, return limited number of search results with strong preferences (biases) towards specific domains.

\noindent {\bf 3.} {\it On grounding of claims in response citations:} LLM agents exhibit strong preferences towards popular URLs, when citing them. While several claims in responses are well grounded in cited URLs, a non-trivial fraction of claims can be attributed to returned search URLs that are not cited. Also, the factuality of ungrounded claims likely generated from models' parametric knowledge remains poor.

Taken together, these findings have important implications for both users and designers of conversational AI agents. For users, invoking Web search does not guarantee higher-quality responses. Moreover, Web-grounded responses can reflect platform-specific preferences for particular sources, rather than a comprehensive view of the Web. For designers, the substantial variation across search-calling decisions, querying strategies, and citation practices highlights the need to optimize and evaluate the entire search life cycle. In particular, designers should pay attention to source attribution, as agents may use information from those Web sources without giving appropriate credit to them. 

We release the code and resources for our analyses in our GitHub repository.\footnote{
\url{https://github.com/mahsaama/AgenticSearchLens/}}
While the donated conversational data cannot be publicly released,
we provide mock data to illustrate the data format and facilitate
use of the code.

%% file: related.tex
\subsection{Related Work}

\textbf{Autonomous information-seeking systems.}
LLMs transformed from static generators into autonomous information-seeking systems. Early approaches taught browsing via imitation~\citep{nakano2022webgptbrowserassistedquestionansweringhuman} and interleaved reasoning with tool use~\citep{YaoZYDSN023}; recent systems internalise this loop through end-to-end reinforcement learning~\citep{jin2025search,song2025r1,wu2026webdancer,DBLP:journals/corr/abs-2509-06501,zheng-etal-2025-deepresearcher}. \textit{Yet strong end-to-end performance leaves open how these models behave in real user-agent interactions, whether they reason soundly at each step or simply land on the right answer.}

\noindent\textbf{User-agent interactions in the real world.} 
Existing datasets fall short in three ways: they cover only ChatGPT under hashed identifiers~\citep{zheng2023judging, zhao2024wildchat}, span 25 models but skew toward open-source systems served through arenas~\citep{zheng2024lmsys}, or rely on opt-in share URLs that capture only self-selected snippets~\cite{yan2026sharechatdatasetchatbotconversations}. Our dataset (Table~\ref{tab:dataset-stats}) provides authentic multi-turn sessions from over 600 users across ChatGPT, Claude, Grok, and DeepSeek, each paired with its full search trace. \textit{To the best of our knowledge, this is the first analysis of agentic information seeking grounded in observable search trajectories across frontier commercial systems.}

\noindent\textbf{Analysing the systems' behavior.}
Prior work largely synthesizes benchmarks to study \textit{when} models invoke retrieval tools~\cite{kale2025lookupanalysinginternal,wu2026call,wei2025browsecompsimplechallengingbenchmark,futuresearch2025deepresearchbenchevaluating, jang2026askmattersadaptiverag}, \textit{how} conversational prompts become search queries~\cite{mo-etal-2023-convgqr,itercqr, zhu-etal-2025-convsearch,pezzuti2026pictureagenticsearch,xi2025infodeepseekbenchmarkingagenticinformation,liu2026veriwebverifiablelongchainweb}, and \textit{whether} responses faithfully ground in cited evidence~\cite{ye-etal-2024-effective,huang-etal-2024-learning,shao2026do,gou2026mindweb}. \textit{We instead study these questions on real-world traces from deployed commercial systems, observing how tool invocation, query formulation, and citation grounding play out across models and over time.}

%% file: sec-setup-revise.tex
\section{Datasets and Experimental Setups}
\label{sec:setup}

\begin{figure}[t]
    \centering
    \includegraphics[width=1\columnwidth]{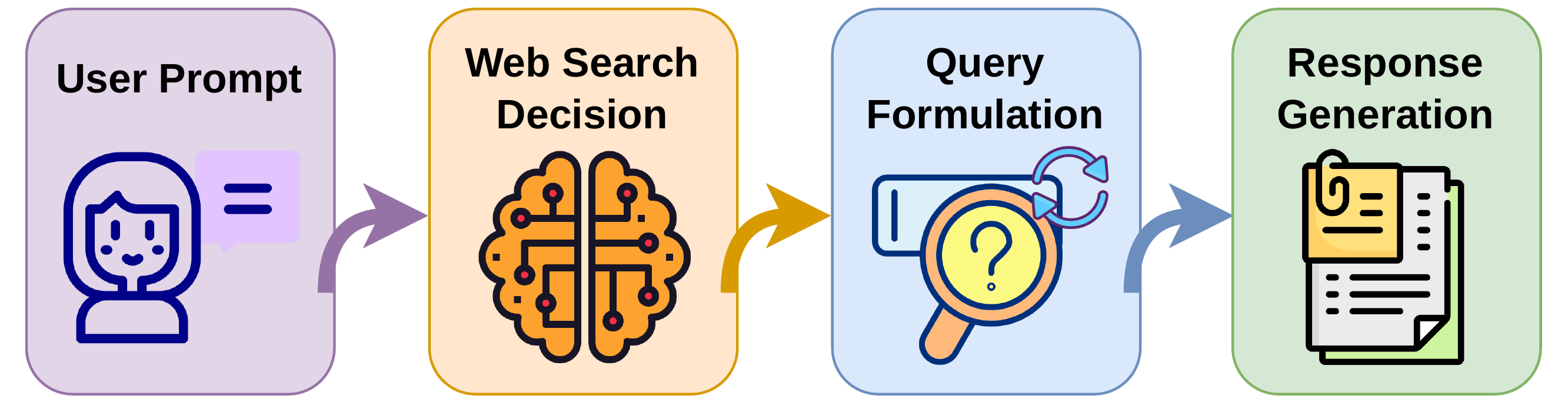}
    \caption{Life cycle of agentic Web search.}
    \vspace{-2mm}
    \label{fig:overview}
\end{figure}

We study the end-to-end lifecycle by which a conversational agent transforms a user prompt into a Web-grounded response (Figure~\ref{fig:overview}). Given a user prompt, the lifecycle consists of three stages: (i) \emph{Web-search decision}, where the agent decides whether Web search is needed; (ii) \emph{query formulation}, where one or more Web queries are generated and iteratively refined; and (iii) \emph{response generation}, where the agent synthesizes information from search results together with parametric knowledge.

\paragraph{Invivo Dataset.}
We analyze conversational traces donated by users of four conversational AI platforms: ChatGPT \cite{openai_chatgpt}, Claude \cite{anthropic_claude}, Grok \cite{xai_grok}, and DeepSeek \cite{deepseek_chat}. These traces are collected through GDPR-compliant data donations following the methodology of \textsc{InvivoGPT}~\cite{karnam2026bowling}. Our ChatGPT dataset extends \textsc{InvivoGPT} with conversations from 10 additional users. Overall, our dataset comprises 171,264 conversations from 613 users with fine-grained logs of user prompts, Web queries, search results, cited URLs, and generated responses (Table~\ref{tab:dataset-stats}). A \emph{turn} consists of a user prompt and the corresponding model response, some of which involved web search. Throughout the paper, we refer to these naturally occurring \textit{interface-based interactions} as the \invivo\ setting. The data collection procedure is described in Appendix \ref{sec:data-collection}.

\begin{table}[t]
\centering
\scriptsize
\tabcolsep2pt
\begin{tabular}{llrrrr}
\toprule    
\textbf{Company} &
\textbf{Platform} &
\textbf{\# Users} &
\textbf{\# Convs} &
\textbf{\# Turns} &
\begin{tabular}[c]{@{}r@{}}\textbf{\# Turns With}\\ \textbf{Web Search}\end{tabular}\\
\midrule
OpenAI & ChatGPT   & 310 & 143,730 & 690,754 & 42,273\\
Anthropic & Claude    & 102 & 9,267   & 64,354  & 1,696\\
xAI & Grok      & 100 & 9,005   & 53,840  & 3,004\\
DeepSeek & DeepSeek  & 101 & 9,262   & 36,020  & 1,730\\
\bottomrule
\end{tabular}%
\caption{\invivo{} statistics across different platforms.}
\label{tab:dataset-stats}
\end{table}

\begin{table}[t]
\centering
\scriptsize
\tabcolsep1.5pt
\begin{tabular}{llp{2.2cm}p{1cm}p{1cm}}
\toprule
\textbf{Company} &
\textbf{Model} &
\textbf{API} &
\textbf{Release Date} &
\textbf{Know. Cutoff} \\
\midrule
OpenAI    & GPT-5.3-chat & Responses API  & Mar 2026    & Aug 2025 \\
Anthropic & Claude Sonnet 4.6 & Messages API   & Feb 2026 & Aug 2025 \\
xAI       & Grok-4.3   &  OpenAI-compatible   Responses API    & May 2026       & Dec 2025 \\
DeepSeek  & DeepSeek-v4-flash &  Anthropic-compatible Messages API  & Apr 2026   & Unknown \\
\bottomrule
\end{tabular}
\caption{\invitro{} details across different models.}
\label{tab:insitu_stats}
\end{table}

\paragraph{Invitro Setup.}
To complement the observational analysis, we conduct controlled experiments using the frontier models listed in Table~\ref{tab:insitu_stats}. We use the same set of 1,000 user prompts (using only the first user message) that exclude personal information through the platforms' APIs, allowing us to compare Web-search behavior under identical inputs. The details of excluding personal information are provided in Appendix~\ref{app:prompt_selection}. 
This setup enables controlled comparisons both across providers (OpenAI, Anthropic, xAI, and DeepSeek) and across OpenAI models (\texttt{GPT-5.3-chat}, \texttt{GPT-4.1-mini}, \texttt{GPT-o4-mini}) while keeping the prompts fixed. Throughout the paper, we refer to this \textit{API-based setting} as the \invitro\ setup.


%% file: sec1-revise.tex
\section{Analyzing Search Calling Decisions}
\label{sec:webtoolcall}

%
Web search is invoked when LLM agents {\it decide} that they need additional (external) knowledge, i.e., their parametric (internal) knowledge is insufficient, to respond to a user prompt. 
So we begin by analyzing agents' decisions to search the Web. 
%

\subsection{Search Calling Behaviors}

\begin{figure}[ht]
    \centering
    \begin{subfigure}[b]{1\columnwidth}
        \centering
        \includegraphics[width=1\linewidth]{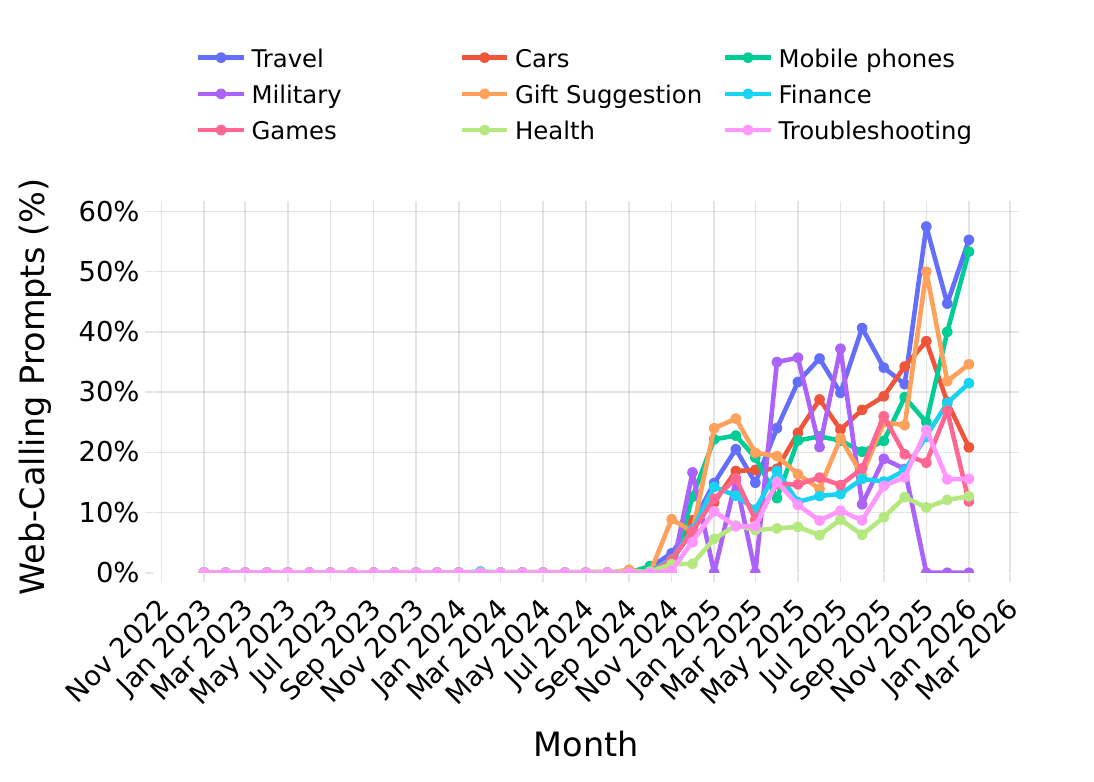}
    \end{subfigure}
    \caption{The fraction of user prompts that result in Web search varies based on topic, even as they show a generally rising trend over time across all topics.}
    \label{fig:prompts_and_web_call_per_topic}
\end{figure}

As shown in Figure~\ref{fig:prompts_and_web_call_per_topic} for ChatGPT (and Figure~\ref{fig:topic_web_call_rate_over_time_all_platforms} in the Appendix for the other platforms), the fraction of user prompts resulting in Web search has steadily increased over time across conversational platforms.
This upward trend is also observed across different conversation topics, although the frequency of Web-search invocation varies substantially by topic; for example, \textit{Travel}-related prompts trigger Web search more frequently than \textit{Health}-related prompts.
%

\textbf{Search calling behaviour differs across LLM agents.} It is hard to directly compare the search calling decisions by different LLM agents in \invivo~, due to differences in user prompts across platforms.
So, we compare search calling decisions by the different LLM agents in \invitro~, querying the same 1000 user prompts. 
Table~\ref{tab:replay} reveals substantial differences in the number of prompts invoking Web search across the agents -- while GPT calls Web search for only 140 (14\% of) prompts, Claude calls Web search for 825 (83\% of) prompts.
Clearly, LLM agents are making very different decisions on which prompts they invoke Web search.

\subsection{Harness Instructions vs. Model Decisions}

\begin{table*}[t]
\centering
\Large
\setlength{\tabcolsep}{3.5pt}

\begin{adjustbox}{max width=0.8\linewidth}
\begin{tabular}{lcccccccccc}
\toprule
\multirow{2}{*}{\textbf{Model}} &
\multirow{2}{*}{\makecell{\textbf{\# Web}\\\textbf{Calling}\\\textbf{Samples}}} &
\multicolumn{3}{c}{\textbf{\makecell{Web-calling\\Samples}}} &
\multicolumn{3}{c}{\textbf{\makecell{Web-calling Samples\\in No-Web Mode}}} &
\multicolumn{3}{c}{\textbf{\makecell{No Web-calling\\Samples}}} \\
\cmidrule(lr){3-5}
\cmidrule(lr){6-8}
\cmidrule(lr){9-11}
& & \textbf{F} & \textbf{C} & \textbf{R}
  & \textbf{F} & \textbf{C} & \textbf{R}
  & \textbf{F} & \textbf{C} & \textbf{R} \\
\midrule

GPT-5.3-chat
& \cellcolor{yellow!25}140
& \makecell[c]{3.50\\{\small(3.31, 3.69)}}
& \makecell[c]{3.52\\{\small(3.34, 3.71)}}
& \makecell[c]{4.06\\{\small(3.90, 4.21)}}
& \makecell[c]{2.81\\{\small(2.61, 3.01)}}
& \makecell[c]{2.56\\{\small(2.39, 2.74)}}
& \makecell[c]{3.26\\{\small(3.06, 3.46)}}
& \makecell[c]{3.97\\{\small(3.90, 4.03)}}
& \makecell[c]{4.08\\{\small(4.02, 4.15)}}
& \makecell[c]{4.54\\{\small(4.49, 4.59)}} \\

Claude Sonnet 4.6
& \cellcolor{yellow!25}825
& \makecell[c]{3.29\\{\small(3.22, 3.35)}}
& \makecell[c]{3.47\\{\small(3.40, 3.55)}}
& \makecell[c]{3.72\\{\small(3.67, 3.78)}}
& \makecell[c]{3.41\\{\small(3.33, 3.49)}}
& \makecell[c]{3.30\\{\small(3.21, 3.39)}}
& \makecell[c]{3.85\\{\small(3.77, 3.92)}}
& \makecell[c]{4.36\\{\small(4.22, 4.50)}}
& \makecell[c]{4.21\\{\small(4.04, 4.37)}}
& \makecell[c]{4.45\\{\small(4.32, 4.57)}} \\

Grok-4.3
& \cellcolor{yellow!25}766
& \makecell[c]{3.81\\{\small(3.75, 3.87)}}
& \makecell[c]{4.13\\{\small(4.07, 4.20)}}
& \makecell[c]{4.24\\{\small(4.18, 4.29)}}
& \makecell[c]{3.30\\{\small(3.22, 3.38)}}
& \makecell[c]{3.57\\{\small(3.48, 3.65)}}
& \makecell[c]{4.18\\{\small(4.11, 4.25)}}
& \makecell[c]{4.42\\{\small(4.30, 4.53)}}
& \makecell[c]{4.45\\{\small(4.32, 4.57)}}
& \makecell[c]{4.61\\{\small(4.50, 4.71)}} \\

DeepSeek-v4-flash
& \cellcolor{yellow!25}584
& \makecell[c]{3.13\\{\small(3.04, 3.21)}}
& \makecell[c]{3.50\\{\small(3.40, 3.59)}}
& \makecell[c]{3.74\\{\small(3.66, 3.81)}}
& \makecell[c]{2.50\\{\small(2.41, 2.59)}}
& \makecell[c]{2.91\\{\small(2.81, 3.01)}}
& \makecell[c]{3.48\\{\small(3.40, 3.56)}}
& \makecell[c]{3.85\\{\small(3.75, 3.96)}}
& \makecell[c]{4.07\\{\small(3.97, 4.17)}}
& \makecell[c]{4.26\\{\small(4.17, 4.34)}} \\

\bottomrule
\end{tabular}
\end{adjustbox}
\caption{Web calling decisions and quality of responses on querying the same 1000 samples in \invitro{} setting across models under different Web-calling modes. F, C, and R denote factuality, completeness, and relevance. Values are reported as mean scores, with 95\% bootstrap confidence intervals shown in parentheses.}
\label{tab:replay}
\end{table*}


\begin{table}[ht]
\centering
\begin{adjustbox}{max width=\columnwidth}
\begin{tabular}{lccc}
\toprule
& \multicolumn{3}{c}{\textbf{Developer Prompt Model}} \\
\cmidrule(lr){2-4}
\textbf{Replay Model} & \textbf{GPT-5.3-chat} & \textbf{GPT-4.1-mini} & \textbf{o4-mini} \\
\midrule
o4-mini      & 478 & 640 & 634 \\
GPT-4.1-mini & 407 & 417 & 440 \\
GPT-5.3-chat      & 140 & 145 & 169 \\
\bottomrule
\end{tabular}
\end{adjustbox}
\caption{Effect of harness instructions and models' self decisions in web-calling decisions over the same 1000 samples. Both the factors play a major role with self decisions dominating.}
\label{tab:dev_prompt_vs_model}
\end{table}

The striking differences in Web search behaviors by our LLM agents could potentially be attributed to two main factors: their \textit{harness instructions} and \textit{models' self-decisions}.
By harness instructions, we refer to explicit guidelines provided to models, often as part of their \textit{system prompt}, on when they should invoke specific tools such as Web search.
Models interpret and follow these instructions, when making their web calling decisions.

\textbf{Both harness instructions and models' decisions impact Web search behaviour.} To better understand the role of harness instructions, we examined three models from OpenAI - \texttt{GPT-5.3-chat}, \texttt{GPT-4.1-mini}, and \texttt{GPT-o4-mini} -- whose system prompts have been reverse-engineered and published on the Web. 
Appendix \ref{app:diffs_in_system_prompts} provides details of their system prompts.
To study the impact of the instructions for Web search calling, we copied the instructions from the system prompt of one GPT model as a developer prompt to another GPT model and queried for the same 1000 user prompts.
Table~\ref{tab:dev_prompt_vs_model} shows how Web search invocations vary across different GPT models, when using their own (native) and other models' harness instructions.
The diagonal cells show that when using native harnesses, GPT-5.3-chat is the most conservative in invoking Web search for 140 prompts, followed by GPT-4.1-mini for 417 prompts and GPT-o4-mini for 634 prompts. 
The off-diagonal cells show that changing harness instructions via developer prompt to those of other models can impact Web calling decisions in a significant way -- GPT-o4-mini reduces the number of Web searches from 634 prompts to 478 prompts.
However, \textit{models' self-decisions} still dominate -- across the different harnesses, GPT-o4-mini still issues far more calls than GPT-5.3-chat.

\subsection{Need for Search Calling}

Next, we investigate whether the observed differences in Web search behavior by LLM agents are justified, i.e., whether they result in improvement in quality of their final responses.
We evaluate the response quality using three complementary measures namely, factuality, completeness, and relevance, which measure the correctness, coverage, and how well the response addresses the user's request, respectively (on a scale up to 5). Human validation shows 67-77\% within-1 agreement between the judge and human annotations.
Appendix~\ref{app:replays_eval_criteria} describes the evaluation metrics and protocol in greater detail.
In Table \ref{tab:replay}, we compare the quality of LLM agent responses for prompts under three settings: (i) prompts for which the model autonomously invoked Web search (\textit{Web-calling samples}), (ii) the same prompts with Web search disabled (\textit{Web-calling samples in No Web mode}), and (iii) prompts for which the model chose not to invoke Web search (\textit{No Web-calling samples}). 
We highlight three observations from Table~\ref{tab:replay}: 
(i) \textit{\textbf{Web search generally improves response quality}} for prompts for which LLM agents autonomously invoked Web search, although the magnitude and direction of the effect vary across models and evaluation metrics. 
Overall, Web search is clearly beneficial for GPT-5.3-chat, Grok-4.3, and DeepSeek-v4-flash, while Claude-Sonnet-4.6, despite invoking Web searches extensively, exhibits at best mixed effects. 
Paired bootstrap significance tests confirm that nearly all observed differences are statistically significant ($p < 10^{-3}$).
(ii) \textbf{\textit{Interestingly, prompts for which models do not invoke Web search achieve higher scores}} across all quality measures, suggesting that LLMs are quite good at judging when prompts can be answered using parametric knowledge alone.
(iii) Overall, the response quality for prompts needing Web search is still lower than those not needing Web search, suggesting that \textbf{\textit{while Web search is useful, it is still not sufficiently effective at plugging the gaps in models' parametric knowledge.}}

\noindent {\bf Summary:} Across conversational AI platforms, there is a consistent trend towards increasing use of Web search when responding to user prompts. However, there are significant differences in search calling behaviors based on topics and across LLM agents. An LLM agent's decision to search is influenced both by its harness instructions and the underlying model itself. LLM agents are quite good at judging when Web search is {\it not needed}. While Web searches do tend to improve the response quality in general, they are {\it not guaranteed} to improve and the improvements {\it do not} fully compensate for the missing parametric knowledge.

%% file: sec2-revise.tex
\section{Analyzing Search Queries and Results}
\label{sec:webqueries}

Once an LLM agent decides to invoke Web search, it transforms a verbose user prompt into one or more Web queries. 
The queries are then answered by platform-specific search engines, which return Web pages (URLs) relevant to the queries, which can be used to both refine the queries and finally, generate the response. 
Accordingly, we first analyze the agents' querying strategies and then investigate the results from their local search engines.
%


\subsection{Querying Strategies}

Unlike traditional Web search, where users typically issue a single query and manually refine it if needed, conversational agents can autonomously orchestrate the querying process by generating multiple Web queries, which can be issued both in parallel as \textit{fan-out} queries and in sequence across \textit{iterations}.
The agents' freedom to orchestrate multiple queries is typically specified in their harness instructions.
%

\begin{figure}[t]
    \centering

    \begin{subfigure}[b]{0.48\columnwidth}
        \centering
        \includegraphics[width=\linewidth]{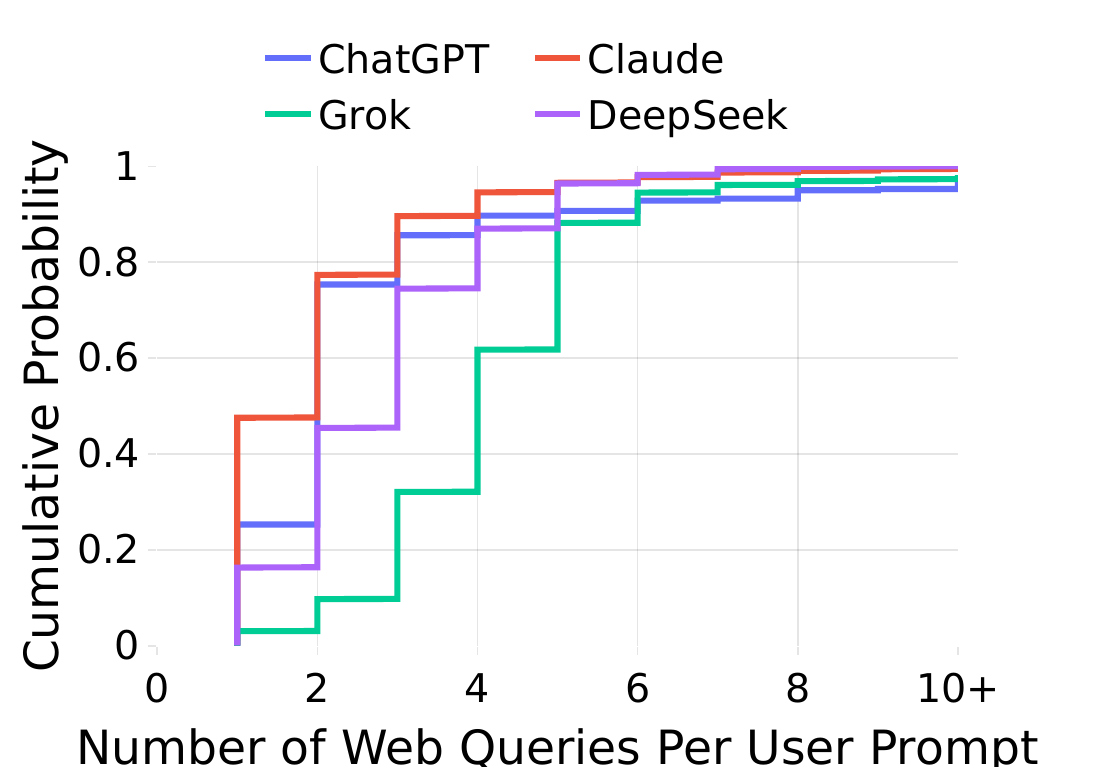}
        \caption{Web queries per prompt}
        \label{fig:total_Web_queries_rev}
    \end{subfigure} 
    \begin{subfigure}[b]{0.48\columnwidth}
        \centering
        \includegraphics[width=\linewidth]{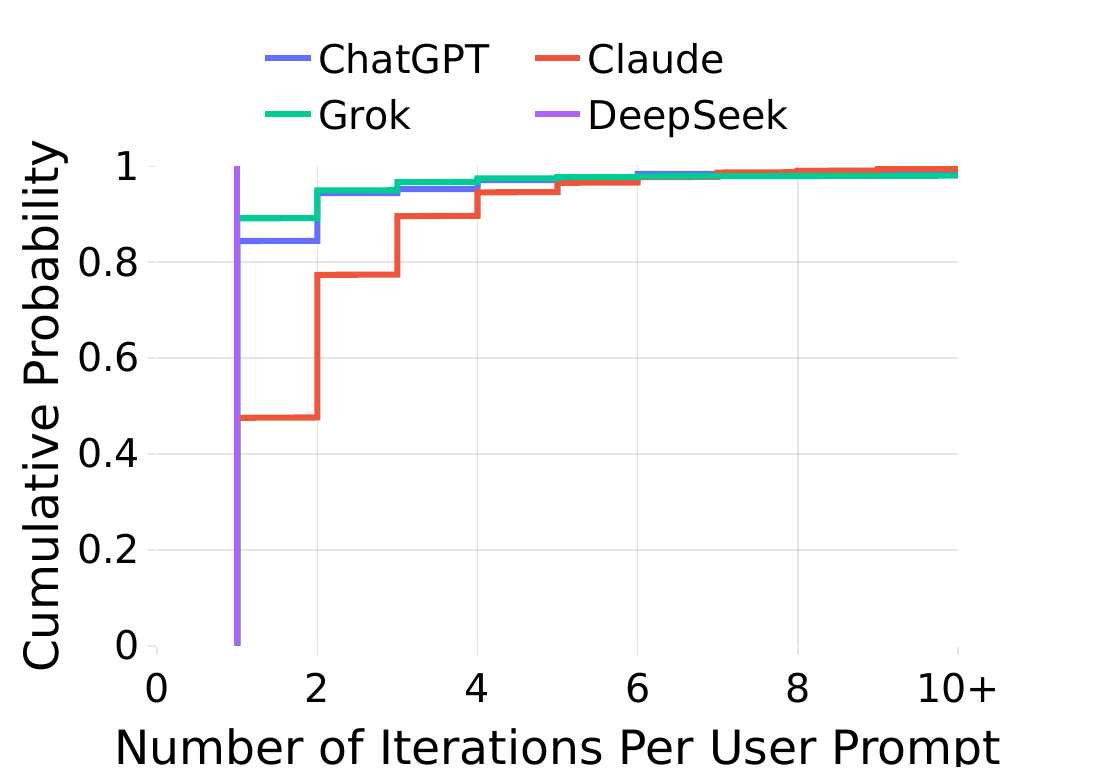}
        \caption{Iterations per prompt}
        \label{fig:iterations_per_prompt_rev}
    \end{subfigure}

    
    \caption{Query complexity increases in different platforms over total number of web queries and iterations.}
    \label{fig:query_reformulation_cdfs_rev}
\end{figure}

    
  

\textbf{Different platforms and models have different querying strategies.} Figures~\ref{fig:total_Web_queries_rev}  and \ref{fig:iterations_per_prompt_rev} show the cumulative distributions of the number of Web queries issued per prompt and the number of iterations over which they are spread. 
The distributions of fan-out queries per iteration and over time are provided in Appendix Figures~\ref{fig:number_of_query_reformulations_and_parallel_queries_over_time} and~\ref{fig:query_comp_longitudinal}.
In our \invivo{} dataset, we observe that LLM agents frequently issue multiple queries per prompt, with the median number varying between 2 (Claude) and 4 (Grok), with the maximum number going much higher than 10.
Interestingly, while ChatGPT, Grok, and DeepSeek generate queries in parallel fan-outs at each iteration, Claude only generates one query at each iteration, resulting in much deeper (over many iterations) search for Claude.
Our \invitro{} experiments with the same platform models via APIs (which can be subjected to different harness) reveal different strategies in Figures~\ref{fig:total_Web_queries_insitu_rev} and \ref{fig:iterations_per_prompt_insitu_rev}.
There, we find that only GPT-5.3 issues fan-out queries, while Grok, DeepSeek, and Claude issue only one query per iteration.
While it is likely that these differing strategies are the result of differing harness instructions, all strategies are fairly involved with no single dominant strategy.

\subsection{Query Formulation and Reformulation}


Next, we investigate how user prompts are reformulated into Web queries, including how iterative queries are reformulated to narrow down the search space, potentially using results from prior queries.

\begin{figure}[t]
    \centering
    \begin{subfigure}[b]{0.48\columnwidth}
        \centering
        \includegraphics[width=\linewidth]{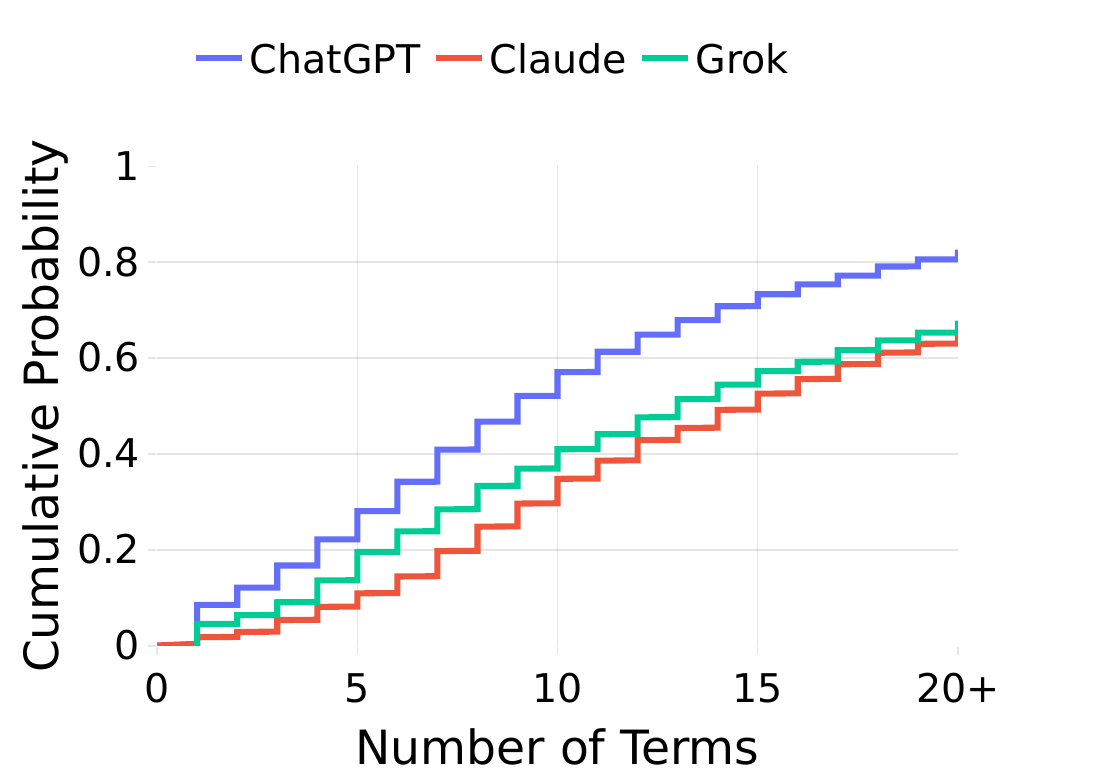}
        \caption{User prompt terms}
        \label{fig:user_prompt_terms_rev}
    \end{subfigure}
    \begin{subfigure}[b]{0.48\columnwidth}
    \centering
    \includegraphics[width=\linewidth]{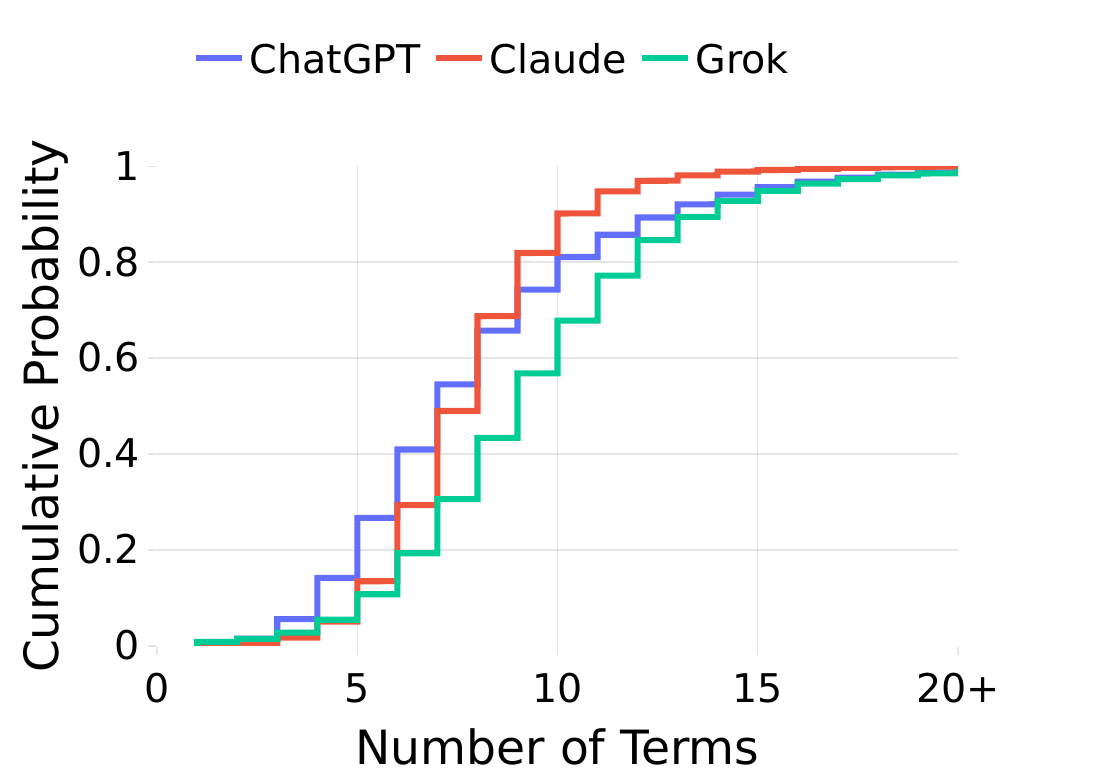}
    \caption{Web query terms}
    \label{fig:Web_query_terms_cdf_rev}
    \end{subfigure}
    \caption{User prompt and Web query complexity across different platforms over number of terms. Web queries contain far fewer terms than user prompts.}
    \label{fig:query_terms}
\end{figure}

\begin{figure}[ht]
    \centering
    \begin{subfigure}[b]{0.48\columnwidth}
        \centering
        \includegraphics[width=1\linewidth]{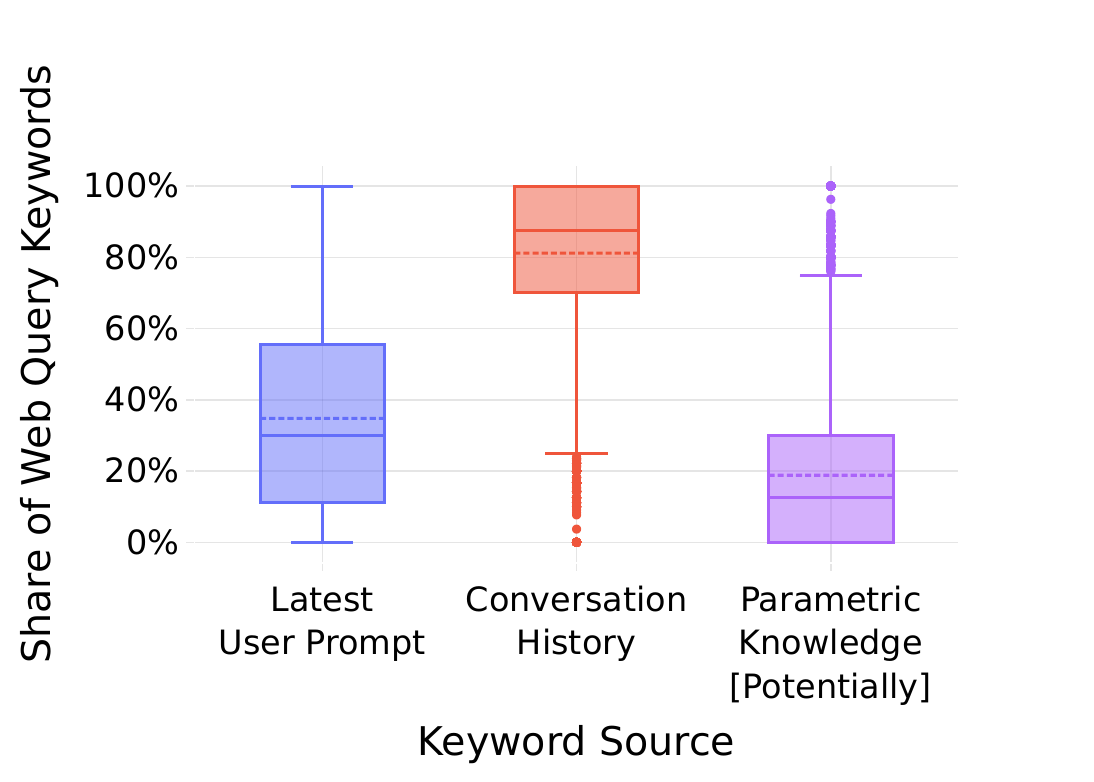}
        \caption{First iteration.}
        \label{fig:Web_query_token_source_detection_1_loop_rev}
    \end{subfigure} 
    \begin{subfigure}[b]{0.48\columnwidth}
        \centering
        \includegraphics[width=1\linewidth]{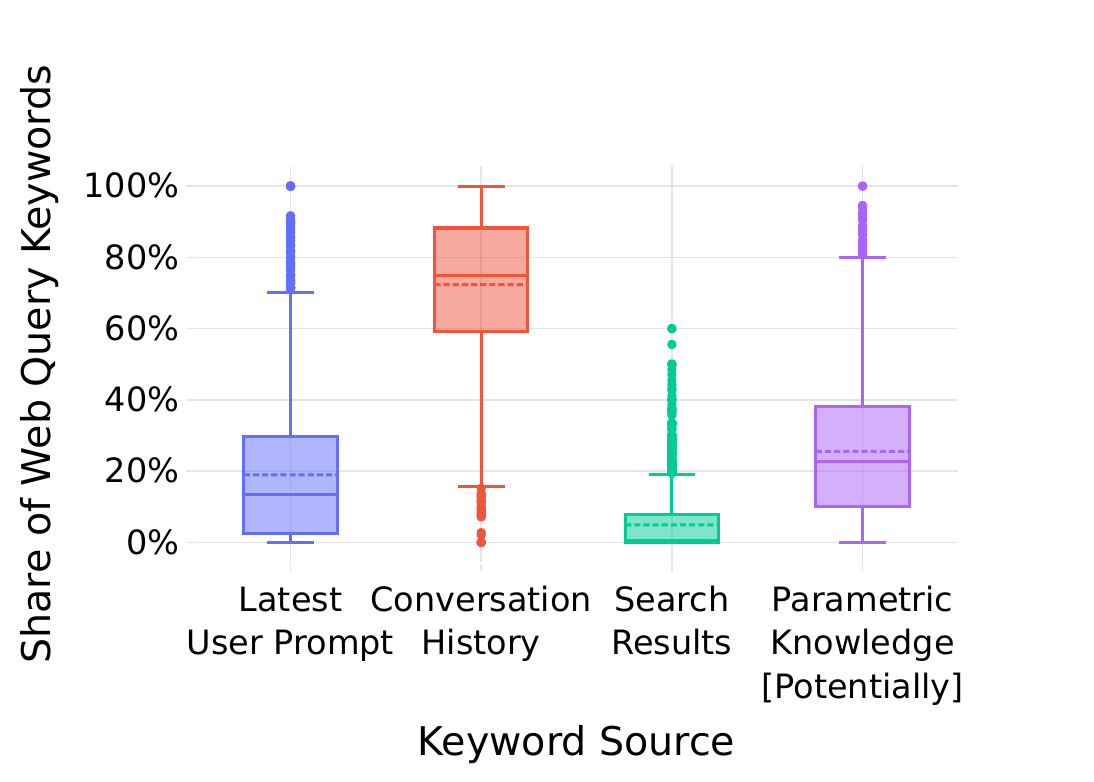}
    \caption{Multi-iteration.}
    \label{fig:Web_query_token_source_detection_multi_loop_rev}
    \end{subfigure}
    \caption{Origin of new query terms in Web queries is conditioned on conversation history and search results from previous iterations.}
    \label{fig:Web_query_token_source_detection_rev}
\end{figure}

\textbf{Web queries are more concise than user prompts but longer than traditional queries.} Figures~\ref{fig:user_prompt_terms_rev} and~\ref{fig:Web_query_terms_cdf_rev} compare the lengths of user prompts and the corresponding Web queries. 
Rather than re-issue verbose user prompts verbatim as a query, the agents rewrite them into concise queries with fewer terms. 
While nearly 80\% of user prompts contain more than 20 terms, almost all generated Web queries contain fewer than 10 to 15 terms.
Despite this compression, LLM generated Web queries remain substantially longer than traditional human generated queries (which are typically 2 to 4 terms~\cite{bendersky2009analysis}).
%

\textbf{New query terms get derived from the conversational context and prior search results.}
Figure~\ref{fig:Web_query_token_source_detection_rev} traces the provenance, i.e., the origins, of query terms introduced during query (re)-formulation in our \invivo{} dataset. 
Terms in the first iteration queries are derived primarily from the latest user prompt (30\%) together with the conversation history (80\%), demonstrating that query construction is conditioned on the broader conversational context. 
As search progresses across iterations, queries increasingly incorporate information from previous search results.
While the overall fraction of query terms that originate from the model's own parametric knowledge remains small, it increases with iterations, suggesting the role of model's parametric knowledge in processing prior search results when re-issuing queries.

\begin{figure}[ht]
    \centering
    \begin{subfigure}[b]{0.48\columnwidth}
    \includegraphics[width=\linewidth]{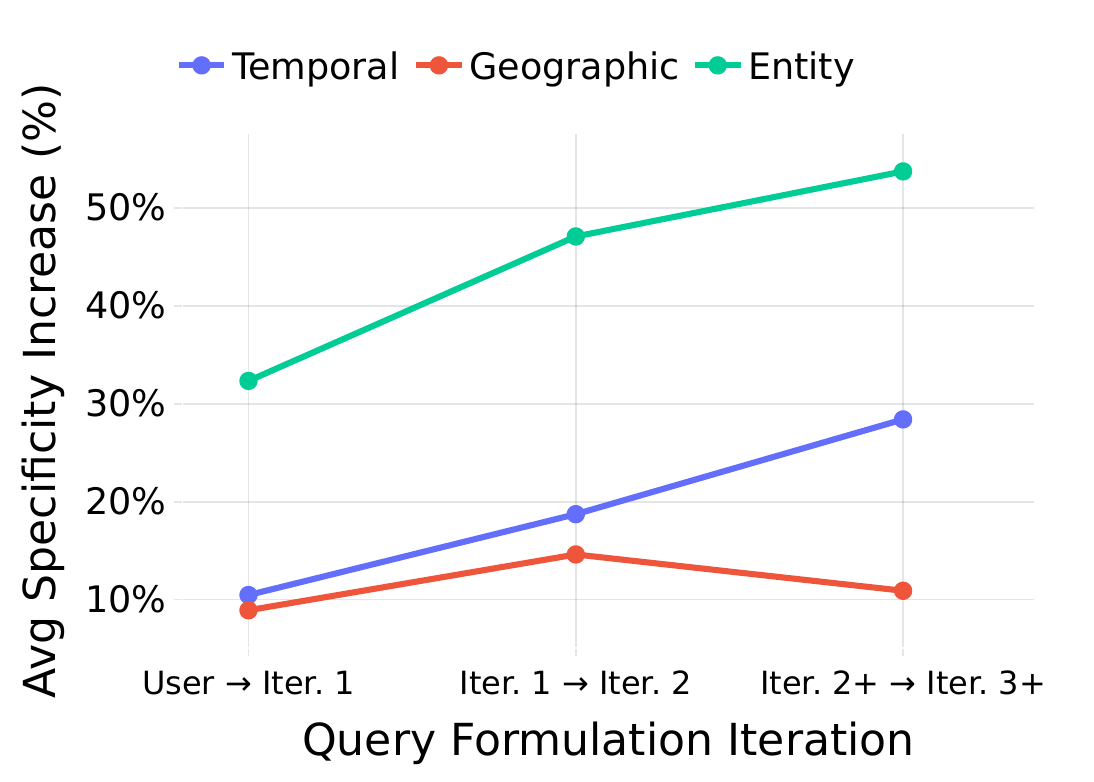}
    \caption{ChatGPT}
    \end{subfigure}
    \centering
    \begin{subfigure}[b]{0.48\columnwidth}
    \includegraphics[width=\linewidth]{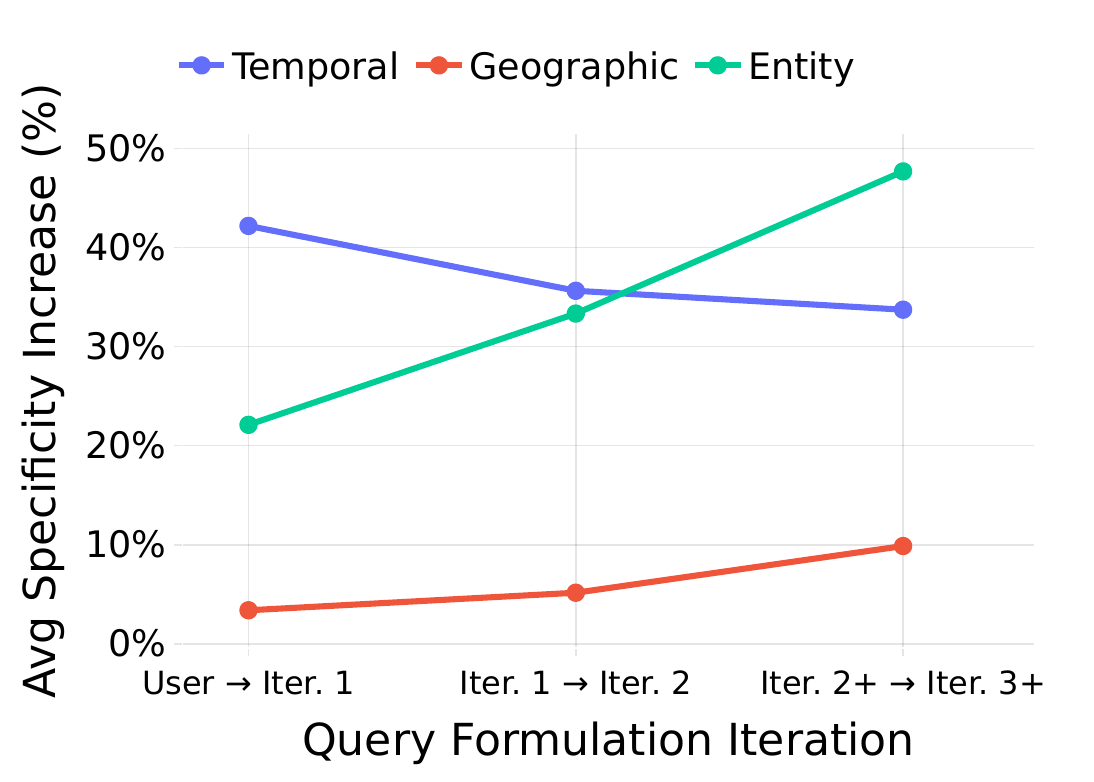}
    \caption{Claude}
    \end{subfigure}
    \caption{Query Specificity (temporal, geographic, and entity) increases over iterations in \invivo{}.}  
    \label{fig:query_specificity_distribution_by_iteration}
\end{figure}

\textbf{Web queries keep getting more specific over iterations.} Figure~\ref{fig:query_specificity_distribution_by_iteration} quantifies query refinement by evaluating the query \textit{specificity} along three complementary dimensions - time, geography, and entity (using a 5-point Likert scale). 
Across successive iterations, Web queries consistently become more specific (leading to higher Likert scores) by introducing constraints like exact dates, locations, and named entities. 
The detailed strategy for computing specificity is in Appendix \ref{app:query_specificity}. 
Human annotators also reached >90\% agreement with judge annotations.
Rather than repeatedly issuing similar searches, conversational agents progressively reduce ambiguity and narrow the search space, enabling better search results.

\subsection{Domain Preferences in Search Results}

Queries issued by LLM agents are processed by platform-specific search engines, which return search results, i.e., relevant Web pages (URLs) that are then used by LLMs to generate final response.
We now investigate the domain preferences of the search engines employed by ChatGPT, Claude, Grok, and DeepSeek. 
%

\begin{table}[t]
\centering
\begin{adjustbox}{max width=\columnwidth}
\begin{tabular}{lrrrr}
\toprule
\textbf{Platform} &
\makecell{\textbf{\#Web} \\ \textbf{Queries}} &
\makecell{\textbf{Avg. \#Web Queries} \\ \textbf{/User Prompt} \\
\textbf{(95\% CI)}} &
\makecell{\textbf{Avg. \#URLs} \\ \textbf{/Web Query} \\
\textbf{(95\% CI)}} &
\makecell{\textbf{Avg. \#URLs} \\ \textbf{/User Prompt} \\
\textbf{(95\% CI)}} \\
\midrule
ChatGPT
& 48,226
& 3.07 {\color{black}\scriptsize(2.98--3.16)}
& 14.06 {\color{black}\scriptsize(13.98--14.14)}
& 43.24 {\color{black}\scriptsize(42.52--43.99)}
 \\

Claude
& 2,936
& 1.80 {\color{black}\scriptsize(1.72--1.88)}
& 9.19 {\color{black}\scriptsize(9.12--9.25)}
& 16.61 {\color{black}\scriptsize(15.94--17.31)} \\

Grok
& 12,371
& 4.55 {\color{black}\scriptsize(4.40--4.71)}
& 8.95 {\color{black}\scriptsize(8.91--8.99)}
& 40.68 {\color{black}\scriptsize(39.18--42.32)} \\

DeepSeek
& 4,092
& 2.63 {\color{black}\scriptsize(2.56--2.71)}
& 5.46 {\color{black}\scriptsize(5.27--5.65)}
& 14.37 {\color{black}\scriptsize(13.74--15.02)}
 \\









\bottomrule
\end{tabular}
\end{adjustbox}
\caption{Web-search querying and search results vary highly across platforms in \invivo, for turns with available Web queries.}
\label{tab:queries-retrieved}
\end{table}


\textbf{Search engines built for LLMs return few results per web query and vary from one platform to another.} Table~\ref{tab:queries-retrieved} summarizes the number of results (URLs) returned by the different search engines per query. 
We observe substantial differences across the platforms. 
On average, search engines on Claude and Grok return around 10 results per query, while DeepSeek returns far fewer (around 5) and ChatGPT returns far more (around 15).
Note that all these search engines return far fewer URLs than a typical Google search return.
When we account for the number of queries issued per user prompt, we find that ChatGPT and Grok  accumulate the most number of results per user prompt (around 40), while DeepSeek and Claude accumulate the least (around 14 to 16).
%
Figure \ref{fig:retrieved_urls_per_web_query_cdf} shows the distributions of search results in greater detail across both \invivo\ and \invitro{}.
%
%
Further details are provided in Appendix~\ref{app:search-results_insitu}.
%

    
    

\begin{figure}[ht]
\centering
\begin{subfigure}{1\linewidth}
    \centering
    \includegraphics[width=1\linewidth]{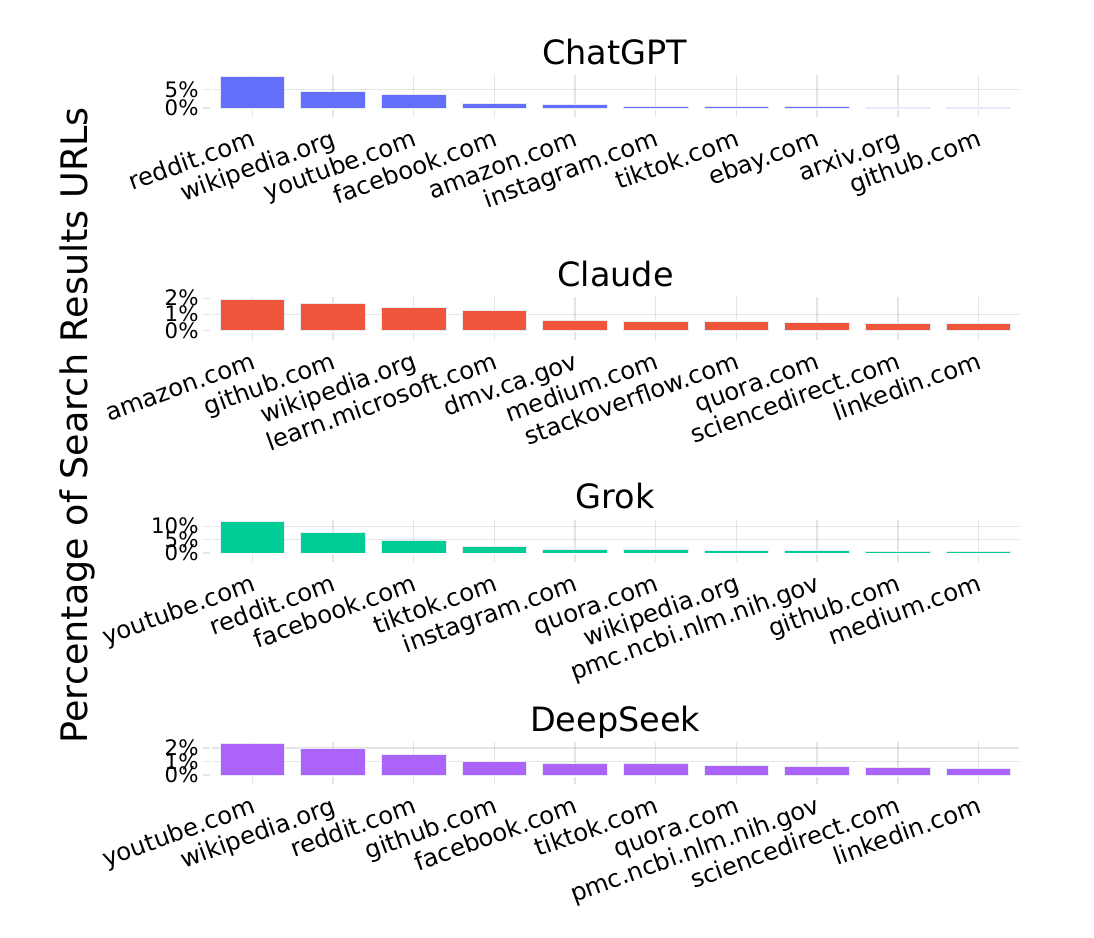}
    \label{fig:top_retrieved_domains_invivo}
    \end{subfigure}
    %
    
\caption{Top 10 domains preferred by search engines across platforms, exhibiting substantial differences in domain preferences.}
\label{fig:top_retrieved_domains_other_platforms_insitu_sec2}
\end{figure}

\textbf{Search engines show preference towards domains that vary from one platform to another.} Given the limited number of search results returned per query, we examine whether search engines on different platforms exhibit different preferences towards specific domains.
To this end, we analyzed the domains from which the search engines are returning most of their results.
We find striking differences in the domains preferred by the search engines as shown in Figure~\ref{fig:top_retrieved_domains_other_platforms_insitu_sec2} (and Figure \ref{fig:top_retrieved_domains_other_platforms_insitu_web} for \invitro{}).
%
Looking at the top-ranked domains (based on number of URLs returned as search results), we highlight two findings: 
(i) the top-domains account for a substantial fraction of all search results, particularly for ChatGPT and Grok. In our \invivo\ ( \invitro\ ) setting the top-10 domains account for 21.3\% (47.8\%) and 32.3\% (29\%) of all returned search results, respectively. 
(ii) different search engines exhibit strong preferences for different domains -- for example, \texttt{reddit.com}  and \texttt{youtube.com} dominate search results returned by ChatGPT and Grok and also rank highly in DeepSeek search, but they are \textit{completely absent} in Claude search! 
Similarly, results from \texttt{xiaohongshu.com}, a Chinese social media and e-commerce platform, are prominent in DeepSeek search, but \textit{completely absent} from other platforms. 

{\bf Summary:}  Our analysis of querying strategies employed by different LLM agents reveals considerable complexity in how web queries are formulated and iteratively reformulated based on returned search results. However, there does not appear to be a dominant search strategy, suggesting that much work remains to be done in evaluating which strategy works best. Our analysis of platforms' search engines results, not only reveals significant differences across platforms, but it also raises several concerns. Compared to traditional search engines, search engines built for LLM agents return only a small number of results (typically, less than 30) and these results are frequently concentrated in a few domains and these preferred domains vary from one search engine to another. It is very unclear if users of LLM agents today are aware of such preferences and if they potentially constitute unwanted/harmful biases on the platforms.






%% file: sec3-revise.tex
\section{Analyzing Search Responses}
\label{sec:assembly_model_responses}


We now analyze how conversational agents synthesize the final response for user prompts using the returned search results from different sources together as well as their own parametric knowledge.
A characteristic feature of LLM agent responses involving Web search is the grounding of the claims by citing supporting external Web pages (URLs). 
To understand how search results are  incorporated into responses,
we decompose responses into atomic claims.
We first analyze how agents \textit{select / generate citations} from the search results and incorporate them into the responses,
and then investigate the extent to which the atomic claims from the response are \textit{grounded} in cited and retrieved sources.

\subsection{Selection / Generation of Citations}



Table~\ref{tab:grounding-rate-all-rev} shows that across platforms and our \invivo\ and \invitro\ setups, responses by LLM agents include, on average, between 1 and 8 citations.

\begin{table}[t]
\centering
\Large
\renewcommand{\arraystretch}{1.15}
\begin{adjustbox}{max width=\columnwidth}
\begin{tabular}{p{1.5cm}lrrrrr}
\toprule
\textbf{Category} &
\makecell{\textbf{Platform}\\\textbf{/ Model}} &
\makecell{\textbf{\#Search Results}\\\textbf{URLs}} &
\makecell{\textbf{\#Cited}\\\textbf{URLs}} &
\makecell{\textbf{\#Cited Search}\\\textbf{Results}} &
\cellcolor{yellow!20}\makecell{\textbf{Citation}\\\textbf{Rate (95\% CI)}}\\
\midrule

\multirow{4}{*}{\invivo}
& ChatGPT
& \textbf{1,196,238}
& \textbf{168,510} 
\normalsize{(4.0)} 
& \textbf{159,580}
& \textbf{13.3\%} {\normalsize(13.1,13.6)} \\

& Claude
& \textbf{27,162}
& \textbf{5,668} 
\normalsize{(3.5)} 
& \textbf{5,409}
& \textbf{19.9\%} {\normalsize(17.2,22.9)} \\

& Grok
& \textbf{110,684}
& \textbf{1,950} 
\normalsize{(0.7)}
& \textbf{1,943}
& \textbf{1.8\%} {\normalsize(1.4,2.2)} \\

& DeepSeek
& \textbf{22,338}
& \textbf{7,610} 
\normalsize{(4.9)}
& \textbf{7,610}
& \textbf{34.1\%} {\normalsize(31.6,36.7)} \\

\midrule

\multirow{4}{*}{\invitro}
& GPT-5.3-chat
& \textbf{4,940} 
& \textbf{534} 
\normalsize{(3.8)}
& \textbf{526}
& \textbf{10.6\%} {\normalsize(9.3,12.0)} \\

& Claude Sonnet 4.6
& \textbf{17,086} 
& \textbf{6,790} 
\normalsize{(8.2)} 
& \textbf{6,790}
& \textbf{39.7\%} {\normalsize(37.1,42.6)} \\

& Grok-4.3
& \textbf{17,090}
& \textbf{4,406} 
\normalsize{(5.7)}
& \textbf{4,364}
& \textbf{25.5\%} {\normalsize(24.2,26.9)} \\

& DeepSeek-v4-flash
& \textbf{10,582}
& --
& --
& -- \\
\bottomrule
\end{tabular}
\end{adjustbox}

\caption{Citation rates (reported with 95\% bootstrap confidence intervals) across platforms in \invivo\ and models in \invitro{} show high variability.}
\label{tab:grounding-rate-all-rev}
\end{table}


\begin{figure}[ht]
    \centering
    \begin{subfigure}[b]{0.48\columnwidth}
        \centering
        \includegraphics[width=1\columnwidth]{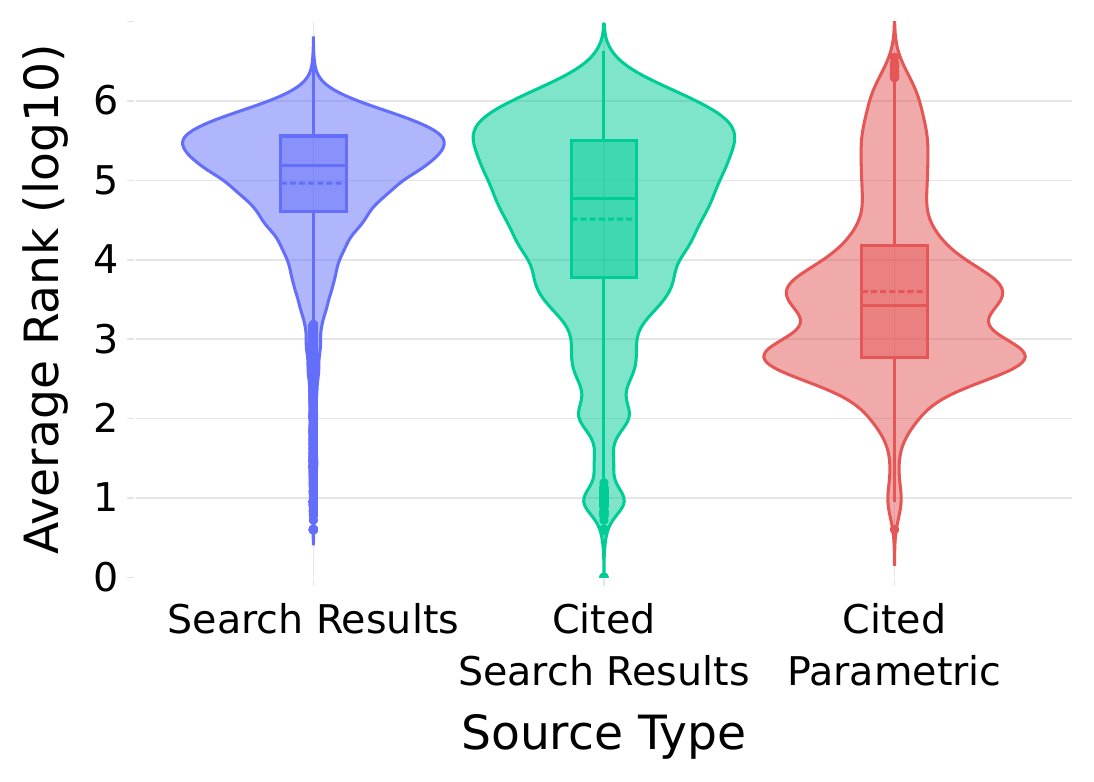}
        \caption{ChatGPT}
        \label{fig:source_rank_violinplot_split_cited_openai}
    \end{subfigure}
    \begin{subfigure}[b]{0.48\columnwidth}
        \centering
        \includegraphics[width=1\columnwidth]{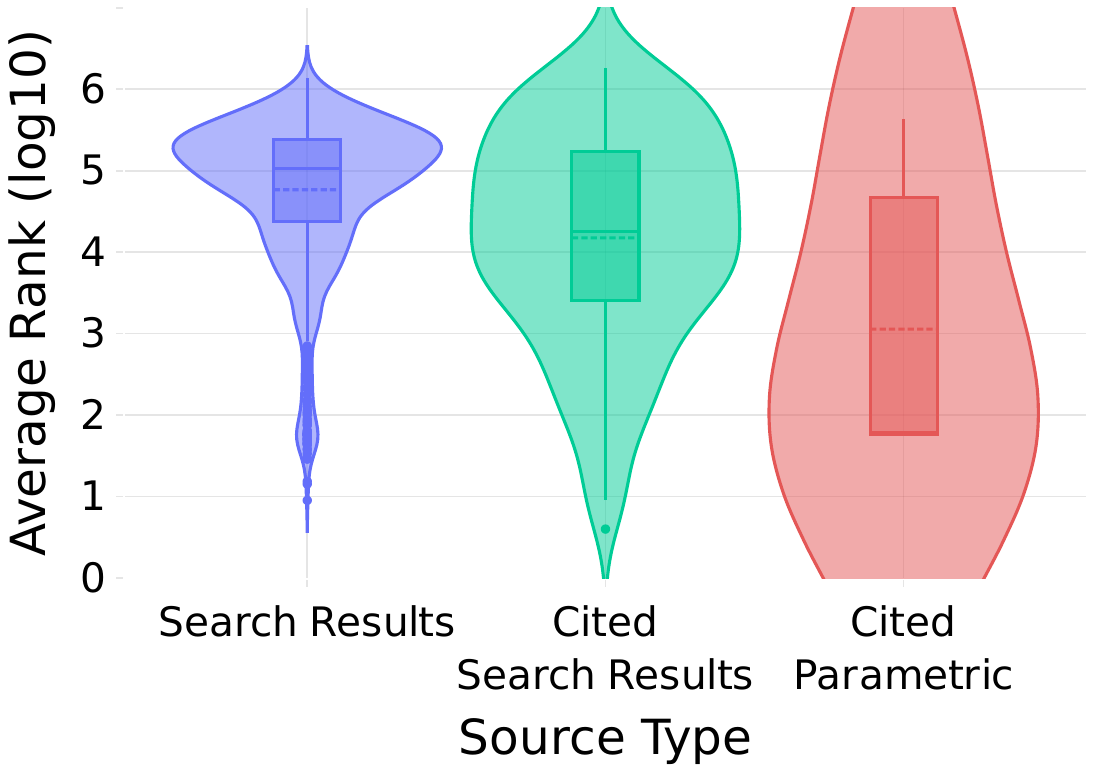}
        \caption{Grok}
        \label{fig:source_rank_violinplot_split_cited_grok}
    \end{subfigure}
    
    \caption{Tranco ranks of domains of URLs \invivo{}.}
    \vspace{-1mm}
    \label{fig:source_rank_violinplot_tranco_rev}
\end{figure}


\textbf{Agents primarily cite URLs returned by Web search.}
We begin by quantifying citation selection using the \emph{citation rate}, defined as the fraction of search result URLs that are eventually cited in the response:
\textit{Citation Rate}=$\frac{N_{\text{Cited\&Search results}}}{N_{\text{Search results}}}$.
Table~\ref{tab:grounding-rate-all-rev} shows that LLM agents cite a non-trivial fraction of all URLs returned as search results to users, with citation rates ranging from 1.7--34.0\% across platforms in the \invivo\ traces and 10.6--40.0\% in the \invitro\ setup. Appendix Table~\ref{tab:model-grounding-rates} shows that this variation is smaller across models from the same platform
(The estimated citation rates are stable across bootstrap resamples, with narrow 95\% confidence intervals). 
Table~\ref{tab:grounding-rate-all-rev} also shows that most of the URLs cited in responses (across both \invivo\ and \invitro\ settings) are selected from URLs returned as search results. 
Across the platforms, only a small fraction (less than 1\% to 5\%) of the cited URLs are generated by the model's parametric knowledge.
Appendix Table~\ref{fig:citation_comp_longitudinal} shows these rates over time.
%
We also investigate their relative reliability. 
Following Rao et al.~\cite{rao2026detecting}, we categorize cited URLs as valid, unverifiable, hallucinated, or dead. 
Across both the \invivo\ and \invitro\ settings, hallucinated citations remain rare (1--2\%; Appendix Tables~\ref{tab:url_status_invivo_platform} and~\ref{tab:url_status_insitu_platform}).
Interestingly, hallucinations occur among both citations selected from search results and generated using parametric knowledge.


\textbf{Agents preferentially cite more authoritative domains.}
Next, we examined whether URLs from certain domains are \textit{preferentially} selected over others.
%
Figure~\ref{fig:source_rank_violinplot_tranco_rev} compares the Tranco ranks of returned search result URLs and cited URLs (\textit{that can come from both search result URLs and parametric domains}), where lower Tranco ranks correspond to more authoritative domains.
We can observe that cited URLs (especially from parametric domains) are consistently drawn from higher-ranked domains than the overall search result URLs, indicating that conversational agents preferentially cite more authoritative sources. 
The same can be observed for other platforms on \invivo\ and \invitro\, as shown in Appendix Figures~\ref{fig:source_rank_violinplot_tranco_other_platforms} and~\ref{fig:source_rank_violinplot_tranco_insitu_rev}, respectively.

\subsection{Grounding of Claims}
%
%
%
Having examined how agents select citations from retrieved search results, we next investigate the extent to which claims are grounded, i.e., supported by evidentiary content in URLs cited and/or returned as search results.
%
For every response, we first decompose the response into atomic claims and then evaluate the grounding of each claim by checking whether it is \textit{entailed} by content scraped from cited and/or retrieved URLs using platform-specific judge models. Details of the judge models and the prompts for claim extraction and entailment evaluation are provided in Appendix~\ref{sec:prompts_claim_nli}. 
We validated the entailment judgments with a human annotation study over randomly sampled examples and observed 93\% agreement between human annotations and the automated judgments. 
%
%

\begin{figure}[ht]
\centering
\begin{subfigure}[b]{0.48\columnwidth}
\centering
\includegraphics[width=\linewidth]{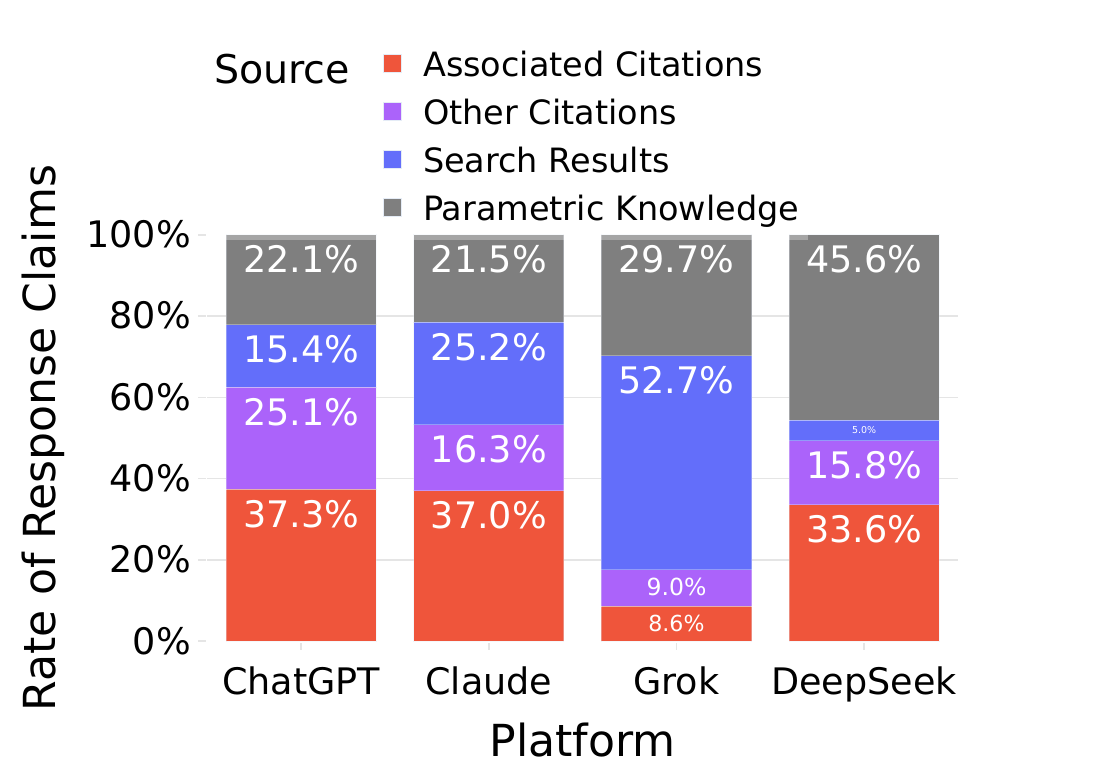}
\caption{Entailment in \invivo}
\label{fig:response_source_nli_sentence_based_judge_summary_claim_invivo_rev}
\end{subfigure}
\begin{subfigure}[b]{0.48\columnwidth}
\centering
\includegraphics[width=\linewidth]{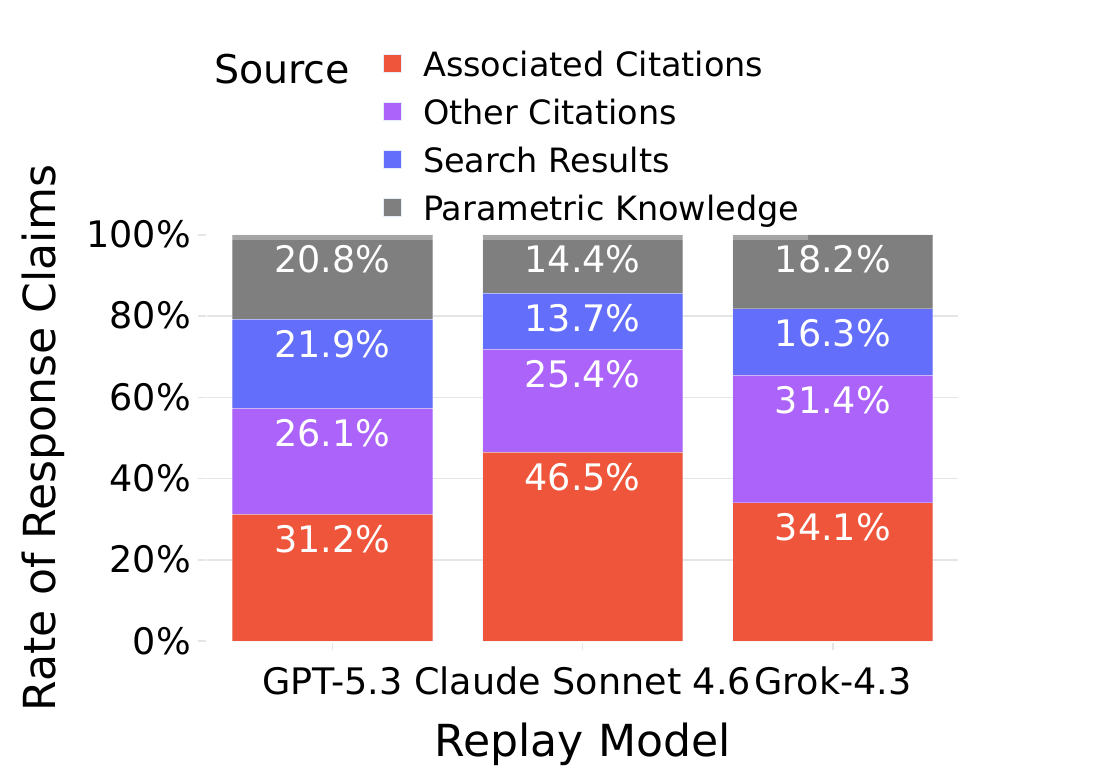}
\caption{Entailment in \invitro}
\label{fig:response_source_nli_sentence_based_judge_summary_claim_insitu_rev}
\end{subfigure}
\caption{Response entailment evaluation shows responses are a mixture of cited evidence, search results, and parametric knowledge.}
\label{fig:response_source_nli_rev}
\end{figure}


\textbf{Most grounded claims are supported by cited sources.}
We begin by evaluating the fraction of response claims grounded in \texttt{associated citations}, i.e., the citations immediately following the claim, followed by all the remaining citations in the response (\texttt{other citations}).
Figure~\ref{fig:response_source_nli_rev} shows that \emph{associated} citations account for the largest fraction of grounded claims across both the \invivo\ and \invitro\ settings (30--46\%), indicating that conversational agents generally place citations close to the evidence supporting the generated claims.
Nevertheless, a substantial fraction of claims are instead supported by \emph{other} cited URLs within the same response, suggesting that claims are not always strictly attributed to citations listed immediately following them. 
%

\textbf{A substantial fraction of claims rely on uncited search results.}
For the claims that are not grounded in cited URLs, we evaluated whether they can be grounded in URLs returned as search results, but remain \textit{uncited}. 
Figure~\ref{fig:response_source_nli_rev} further reveals that a non-trivial fraction of claims are supported by URLs in search results, but which are never cited in the final response. 
Across ChatGPT, Claude, and Grok platforms, these uncited, yet searched URLs, consistently account for 14\% to 53\% of all claims across both \invivo\ and \invitro\ settings. 
%
This behavior raises concerns about whether LLM agents are properly attributing credit to the source URLs.
%

\textbf{Ungrounded claims are less factual than grounded claims.}
After our attempts to ground claims in URLs that are cited and returned as search results, we are left with claims that remain \textit{ungrounded} -- these are unexplained claims that are (potentially) generated from models' parametric knowledge. 
Figure~\ref{fig:response_source_nli_rev} shows that upwards of 15\% to 20\% of all claims remain ungrounded.
Finally, we evaluated and compared the factuality of claims that are grounded vs. ungrounded 
(Appendix Tables~\ref{tab:claim_factuality} and \ref{tab:claim_factuality_insitu}). 
We find that claims supported by \textit{associated citations and other cited sources} achieve the highest factuality scores (3.47, 3.35 in ChatGPT), followed by claims supported by \textit{search results that are not cited} (3.33). 
In contrast, ungrounded claims (potentially generated by parametric knowledge) consistently receive the lowest factuality scores (2.84) across platforms. 
%

{\bf Summary:} Our analysis of responses synthesized by LLM agents using search results reveals a preference towards citing URLs from more popular domains.
While many claims in the response are well-grounded in cited URLs, a non-trivial fraction of claims are (a) grounded in uncited URLs returned during search, suggesting potentially poor attribution practices, and (b) ungrounded and potentially generated from parametric knowledge. Such ungrounded claims score poorly on factuality, raising reliability concerns. 

%% file: conclusion.tex
\section{Conclusion}

Conversational LLMs are increasingly becoming Web-search agents that must decide when to search, how to search, and how to incorporate retrieved information into their responses.
By combining real-world conversational traces with controlled experiments across four major platforms, we characterize this end-to-end search lifecycle. 
We find substantial differences across agents at each stage: Web-search invocation varies considerably and does not uniformly improve response quality; agents engage in distinct, increasingly specific querying strategies; retrieved results exhibit strong platform-specific domain preferences; and, while responses are largely grounded in retrieved evidence, a non-trivial fraction of claims rely on uncited search results or remain unsupported by the observable search trace.
These findings show that conversational Web search should be understood and evaluated as a sequence of interconnected decisions rather than as a single retrieval step. Building more reliable search agents will therefore require improving not only retrieval itself, but also when agents search, how they seek evidence, and how faithfully they attribute that evidence in their responses.

%% file: limitations.tex
\section*{Limitations}

Our study has several limitations that should be considered when interpreting the results.

First, our analysis relies on observational conversational traces collected from users of four popular chatbot platforms. Although these datasets provide valuable insight into real-world agentic Web-search behavior, they do not expose the full internal reasoning processes, hidden chain-of-thought traces, ranking algorithms, or proprietary retrieval pipelines used by the platforms. As a result, some observed behaviors may reflect platform-specific harness implementations rather than the underlying language models alone.

Second, while our datasets span multiple platforms and millions of conversational turns, the data may not fully represent the broader population of chatbot users or all deployment settings. User populations, geographic regions, languages, subscription tiers, and platform-specific product features may influence Web-search behavior. Additionally, replay experiments were conducted on sampled conversations and may not capture all real-world conversational dynamics.

Third, several analyses in this work depend on automated judge models for labeling conversational situations, query reformulation types, entailment, and response quality. Although we performed human validation and observed high agreement rates, judge-based evaluation remains imperfect and may inherit biases or inconsistencies from the evaluation models themselves.

Fourth, our provenance analysis of query reformulation cannot perfectly distinguish whether newly introduced query terms originate from retrieved evidence, latent parametric knowledge, or implicit reasoning traces. Similarly, identifying ``parametric citations'' is inherently approximate because the internal memory and pretraining sources of models are inaccessible.

Finally, our study primarily focuses on English-language interactions and Web-search-enabled text generation. We do not analyze multimodal retrieval, non-English retrieval behavior, personalized memory systems in depth, or downstream effects such as user trust, persuasion, or behavioral influence. Future work could extend our framework to multilingual, multimodal, and fully agentic environments involving planning, browsing, and long-horizon tool use.

%% file: ethics.tex
\section*{Ethical Considerations}

This study was conducted in accordance with the Ethical Review Board (ERB) guidelines of our institution. All datasets are obtained through GDPR-compliant data donations, where participants provided explicit informed consent to share their conversational data for research purposes. Due to the sensitive nature of conversational traces, the collected datasets are stored on secured institutional servers, were not shared with any third party, and will not be publicly released. The donated data will be permanently deleted within three years following the completion of this study.

Several analyses in this work rely on LLM-as-a-judge evaluations using GPT-4o-mini for OpenAI data to scale annotation and evaluation. We chose GPT-4o-mini because it belongs to the same platform family as the original OpenAI conversational traces analyzed in this work. All evaluations were conducted through OpenAI’s EDU workspace, which provides strict data-protection guarantees and prohibits the use of submitted data for model training.

For analyses involving Claude, Grok, and DeepSeek, we conducted evaluations with Claude-Haiku-3.5, Grok-3-mini, and Deepseek-chat models directly through the respective platforms rather than routing conversational data through external systems. This design choice was made to minimize unnecessary cross-platform data transfer and reduce potential risks of conversational data leakage across providers.

%% file: appendix.tex
{\Large\textbf{Appendix Overview}}

\begin{itemize}
    \item Appendix~\ref{sec:data-collection}: Data Collection Strategy
    \item Appendix~\ref{sec:llm-usage}: LLM Usage
    \item Appendix~\ref{sec:replay}: \invitro{} Setup and Evaluations 
    \item Appendix~\ref{app:query_strategy}: Additional Analysis of Querying Strategies
    \item Appendix~\ref{app:search_res_and_citations}: Citation Selection and Search Result
    \item Appendix~\ref{sec:hallucinated_url_detection}: Hallucinated URL Detection
    \item Appendix~\ref{sec:app-nli}: Grounding and Entailment Analysis
    \item Appendix~\ref{sec:app-other-platforms}: Additional \invivo{} Results
    \item Appendix~\ref{sec:app-replay-platforms}: Additional \invitro{} Results
    \item Appendix~\ref{sec:human_validation}: Human Annotation and Judge Validation
    \item Appendix~\ref{sec:prompts}: Prompts
    \item Appendix~\ref{app:disclaimer}: Disclaimer for annotators

\end{itemize}

\input{data_collection}
\input{llm_usage}
\input{replay_setup}

\input{longitudinal}
\input{hallucinated_urls}
\input{nli}

\input{other_platform_res}
\input{replay_platforms}

\clearpage
\newpage
\input{human_validation}

\input{prompts}
\input{disclaimer}

%% file: data_collection.tex
\section{Data Collection Strategy}
\label{sec:data-collection}

We build on the data collection strategy introduced by \textsc{InVivoGPT}~\cite{karnam2026bowling} and adapt it to collect conversation histories, including Web search traces, from four AI platforms.
The strategy follows the paradigm of GDPR-empowered data donations \cite{karnam2025setting,zannettou2024analyzing, dash2026algorithmic}.
Under Article~15(3) of the GDPR \cite{EU2016GDPR}, users have the right to obtain a copy of personal data processed by online platforms.
This paradigm enables participants to request exports of their own chatbot interactions and subsequently donate these exports for research.
Following this approach, we recruited participants through the crowd-sourcing platform Prolific~\cite{prolific2025prolific}. Participants were asked to exercise their data access rights and obtain exports of their interaction histories from four conversational AI platforms: ChatGPT, Claude, Grok, and DeepSeek. 
We then parsed the donated exports into a common representation suitable for cross-platform analysis. 
In total, we collected data from $613$ users across platforms, resulting in $171{,}264$ tracked conversations. 
Table~\ref{tab:dataset-stats} reports the breakdown by platform.

This study focuses on Web search usage. Specifically, we use the donated traces to study when and how web search is invoked by AI systems, and how search-grounded responses are integrated into conversations.

\paragraph{Conversation Traces.}
The exported conversation archives differ across ChatGPT, Claude, Grok, and DeepSeek. For each platform, we parse the provider-specific export format to recover (i) user prompts, (ii) assistant responses, (iii) Web-search tool invocations, (iv) generated Web queries, (v) search result URLs, and (vi) citations included in the final response, whenever exposed by the platform. We then normalize these platform-specific representations into a common schema consisting of conversations, turns, Web queries, search result URLs, cited URLs, and associated metadata. Platform-specific fields unavailable in an export (e.g., Web queries in DeepSeek traces) remain missing in the normalized representation.
The collected traces contain conversation-level and message-level records. We represent a conversation as a sequence of turns, where a turn consists of a user prompt and the corresponding model response.
For our analysis, we further extract platform-specific indicators of Web search behavior, including whether web search was invoked, the search queries issued by the system, search results, and cited URLs where available, and any metadata linking results of web search to the generated response.
Because providers expose different formats and levels of detail, we normalize all data into a shared schema. 
We refer to the four per-platform traces as \textit{\textbf{Invivo traces}}.

\paragraph{Ethical Considerations.}
Our study was conduced with careful attention to ethical considerations and is approved by the ERB in our institution.
Because conversation histories may contain personal, sensitive, or identifying information, participants were informed about the nature of the data donation and the risks associated with it.
Participants provided explicit informed consent before uploading any data.
All donated data were stored on secure servers with access restricted to authorized members of the research team. Before analysis, participant identifiers were replaced with anonymized IDs, and personally identifying information was removed or minimized where possible, with the same procedure as~\cite{karnam2026bowling}. We do not share raw donated conversation histories with third parties.

\paragraph{\invitro\ experiments.} Before conducting these experiments, all sampled conversations underwent the same automated PII-sanitization pipeline introduced by~\cite{karnam2026bowling}. Specifically, conversations were filtered using LLM-based prompts designed to identify personally identifiable information (PII), including names, contact information, addresses, account identifiers, credentials, and other user-specific information. Conversations flagged as containing PII were excluded from replay. The remaining prompts were subsequently reviewed manually by the authors to verify that no identifying or sensitive information remained before being submitted to the commercial APIs. Consequently, only sanitized user prompts were transmitted during replay experiments. The PII-identification prompts used in the filtering pipeline are provided in Appendix~\ref{app:pii_id_prompt}.

%% file: llm_usage.tex
\section{LLM Usage}
\label{sec:llm-usage}
In this paper, we leverage LLMs for the following purposes:

\begin{enumerate}
\item \textbf{Data Annotation}: LLMs were used to support parts of the data annotation process.
\item \textbf{Text Improvement}: Used to correct grammatical errors and provide feedback on writing.
\item \textbf{Code Writing}: LLM-based copilots assisted in generating some portions of code.
\item \textbf{Related Work Discovery}: In addition to traditional search methods, we employed AI2 Paper Finder and OpenAI Deep Research to identify relevant literature.
\end{enumerate}

%% file: replay_setup.tex
\section{\invitro{} Setup \& Evaluations}
\label{sec:replay}

\subsection{Prompt Selection}
\label{app:prompt_selection}

For \invitro\ setup, we constructed the dataset by sampling 1,000 user prompts from the \invivo{} dataset, ChatGPT traces, comprising 500 prompts that originally triggered Web search and 500 that did not. Since the experiments use only the first user message from each conversation, none of the prompts contains preceding conversational history. 
Before sampling, all conversations underwent the same PII-sanitization protocol introduced by~\cite{karnam2026bowling}. Briefly, an automated PII-identification pipeline excluded conversations containing personally identifiable information. After selecting the 1,000 prompts, several authors independently reviewed every prompt to verify that no personally identifiable or other user-specific information remained. Consequently, only sanitized user prompts were transmitted during the \invitro\ experiments. The filtering prompts used by the automated pipeline are reproduced in Appendix~\ref{app:pii_id_prompt}.

\subsection{System Prompts and Web-search Instructions}
\label{app:diffs_in_system_prompts}

We use publicly available snapshots of the system prompts for GPT-5.3-chat, GPT-4.1-mini, and GPT-o4-mini~\cite{johnson2026systempromptsleaks}. Although all three prompts provide guidance for Web-search usage, they differ substantially in their structure and triggering criteria. 
GPT-5.3-chat contains the most detailed and well-organized instructions, covering a broader range of search-triggering conditions and providing more fine-grained guidance on when Web search should be invoked. In contrast, GPT-4.1-mini provides a shorter and more concise set of search triggers, while GPT-o4-mini contains a similar set of triggers but presents them in a less structured manner. The extracted Web-search instructions used in our experiments are shown in~\ref{sec:system_prompt_gpt_models}.

\subsection{Extracted Web-search Instructions}
\label{app:replay_extracted_dev}

To investigate whether differences in Web-search behavior arise from the underlying model or from the Web-search instructions it receives, we extracted only the Web-search-related portions of the publicly available system prompts and reformatted them into a common structure while preserving their original meaning. These extracted instructions were then supplied as developer prompts during replay experiments, independent of the replay model's native system prompt. For example, we provided the Web-search instructions extracted from GPT-5.3-chat to GPT-o4-mini to evaluate whether the more conservative search guidance reduces GPT-o4-mini's tendency to invoke Web search. We repeated this procedure for all combinations of replay models and extracted Web-search instructions. Table~\ref{tab:dev_prompt_vs_model} in the main paper summarizes the resulting Web-search calling frequencies.

\subsection{Evaluation Criteria}
\label{app:replays_eval_criteria}

We evaluate the quality of replayed responses along three complementary dimensions: factuality, completeness, and relevance. Factuality measures whether the information presented in the response is factually correct. Completeness measures whether the response sufficiently addresses all important aspects of the user's request. Relevance measures how well the response remains focused on and addresses the user's request. Each dimension is evaluated independently on a 5-point Likert scale \cite{li2024llmsasjudgescomprehensivesurveyllmbased} by \texttt{gpt-5.6-luna}, where higher scores indicate better response quality. The evaluation prompts used by the judge are provided in Appendix~\ref{sec:prompts_metrics}. To compare the Auto and No-Web settings, we performed paired bootstrap significance tests over the per-prompt evaluation scores. Statistical significance was assessed using 100,000 paired bootstrap resamples, and two-sided p-values are reported in Table~\ref{tab:replay_pvalues}. 

\begin{table}[t]
\centering
\small
\begin{adjustbox}{max width=\columnwidth}
\begin{tabular}{lccc}
\toprule
\textbf{Model} & \textbf{Factuality} & \textbf{Completeness} & \textbf{Relevance} \\
\midrule
GPT-5.3-chat      & $<10^{-4}$ & $<10^{-4}$ & $<10^{-4}$ \\
Claude Sonnet 4.6 & $0.0038$   & $<10^{-4}$ & $0.0015$   \\
Grok-4.3          & $<10^{-4}$ & $<10^{-4}$ & $0.1470$   \\
DeepSeek-v4-flash & $<10^{-4}$ & $<10^{-4}$ & $<10^{-4}$ \\
\bottomrule
\end{tabular}
\end{adjustbox}
\caption{P-values from paired bootstrap significance tests comparing replay responses generated with and without Web search for prompts that autonomously invoked Web search.}
\label{tab:replay_pvalues}
\end{table}






\subsection{Human Validation of Judge Evaluations}
\label{sec:replay_human_judge}

We further conducted a human evaluation study using three independent annotators over 30 samples. The annotators achieved a pairwise within-1 inter-annotator agreement of $87\%, 83\%$, and $83\%$, while within-1 agreement between human annotations median and the judge model reached $67\%, 77\%$, and $73\%$ for factuality, relevance, and completeness, respectively. These results provide additional evidence that the judge-based evaluation aligns with human preferences and factuality assessments.

%% file: longitudinal.tex
\section{Additional Analyses of Querying Strategies}
\label{app:query_strategy}

\subsection{Distribution of Fan-out and Iterative Queries}

\begin{figure*}[ht]
    \centering
    \begin{subfigure}[b]{0.48\linewidth}
        \centering
        \includegraphics[width=1\linewidth]{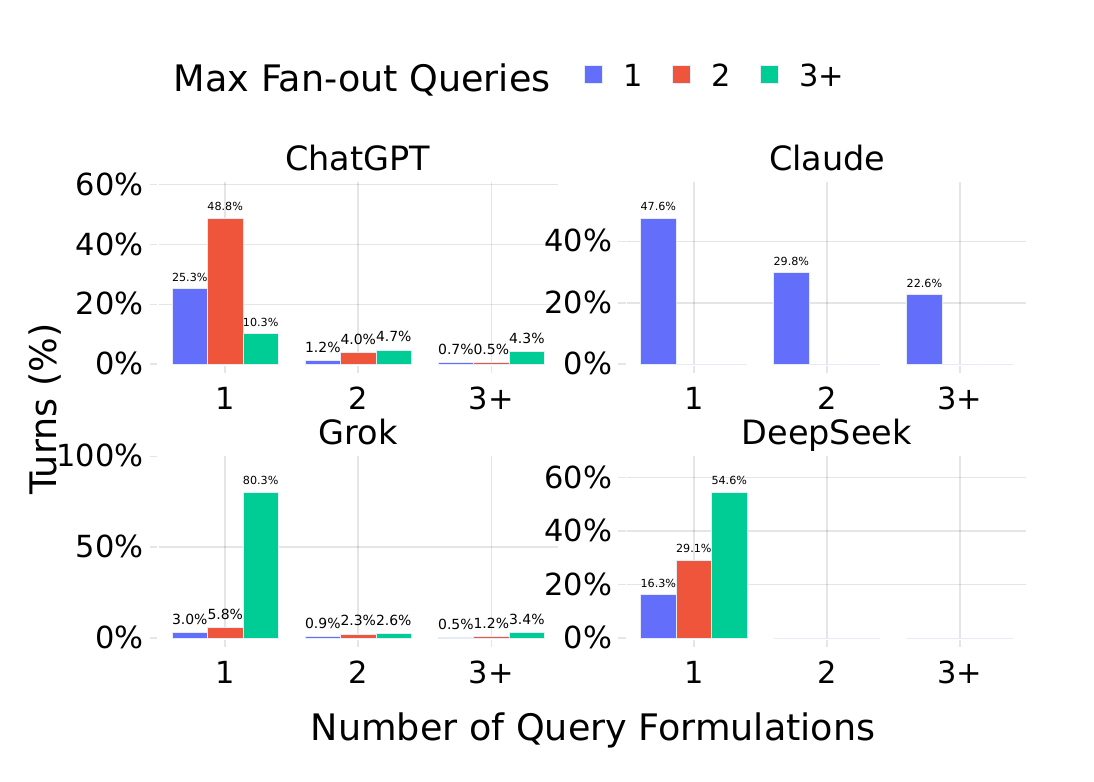}
        \caption{\invivo}
        \label{fig:number_of_query_reformulations_and_parallel_queries_over_time_invivo}
    \end{subfigure} 
    \begin{subfigure}[b]{0.48\linewidth}
        \centering
        \includegraphics[width=1\linewidth]{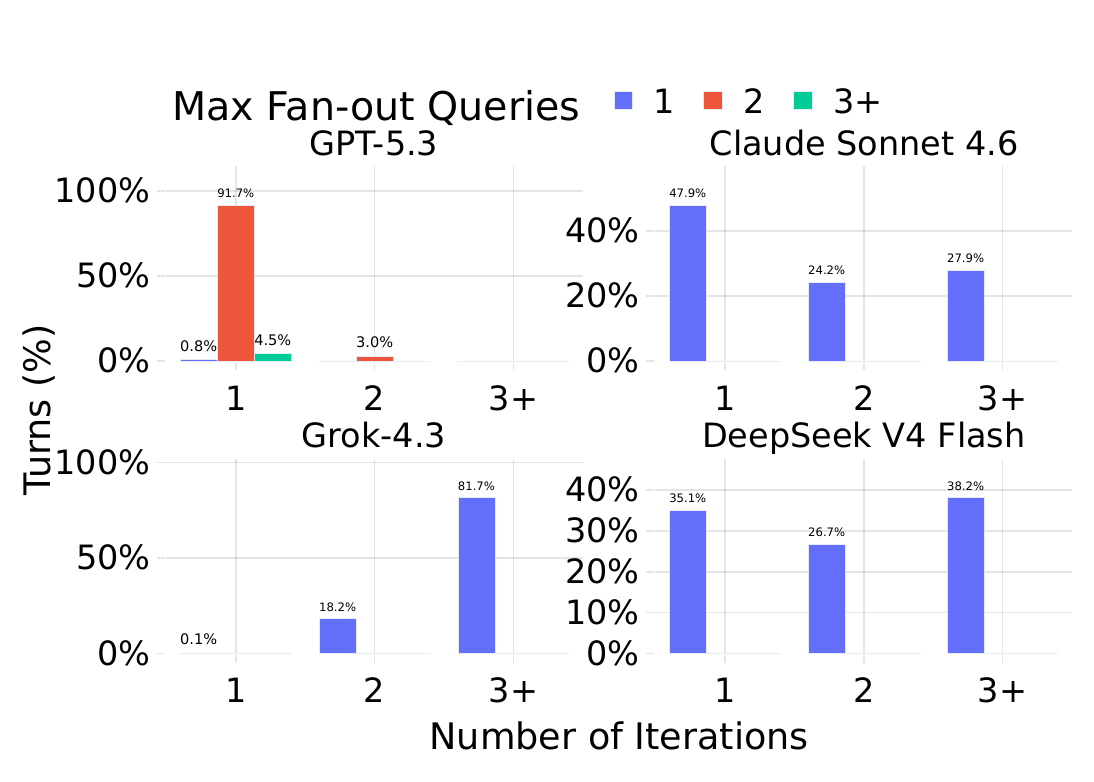}
    \caption{\invitro}
    \label{fig:number_of_query_reformulations_and_parallel_queries_over_time_insitu}
    \end{subfigure}
    \caption{Distribution of fan-out (parallel) queries and iterative query reformulations across platforms. ChatGPT and Grok employ a mixture of fan-out and iterative search, whereas Claude relies primarily on iterative refinement.}
    \label{fig:number_of_query_reformulations_and_parallel_queries_over_time}
\end{figure*}

Figure~\ref{fig:number_of_query_reformulations_and_parallel_queries_over_time} shows the distribution of Fan-out and Iterative queries for each platform.

\subsection{Longitudinal Evolution of Querying Strategies}
\label{app:query_strategy_longitudinal}

\begin{figure*}
    \centering
    \begin{subfigure}[b]{0.48\linewidth}
        \centering
        \includegraphics[width=1\linewidth]{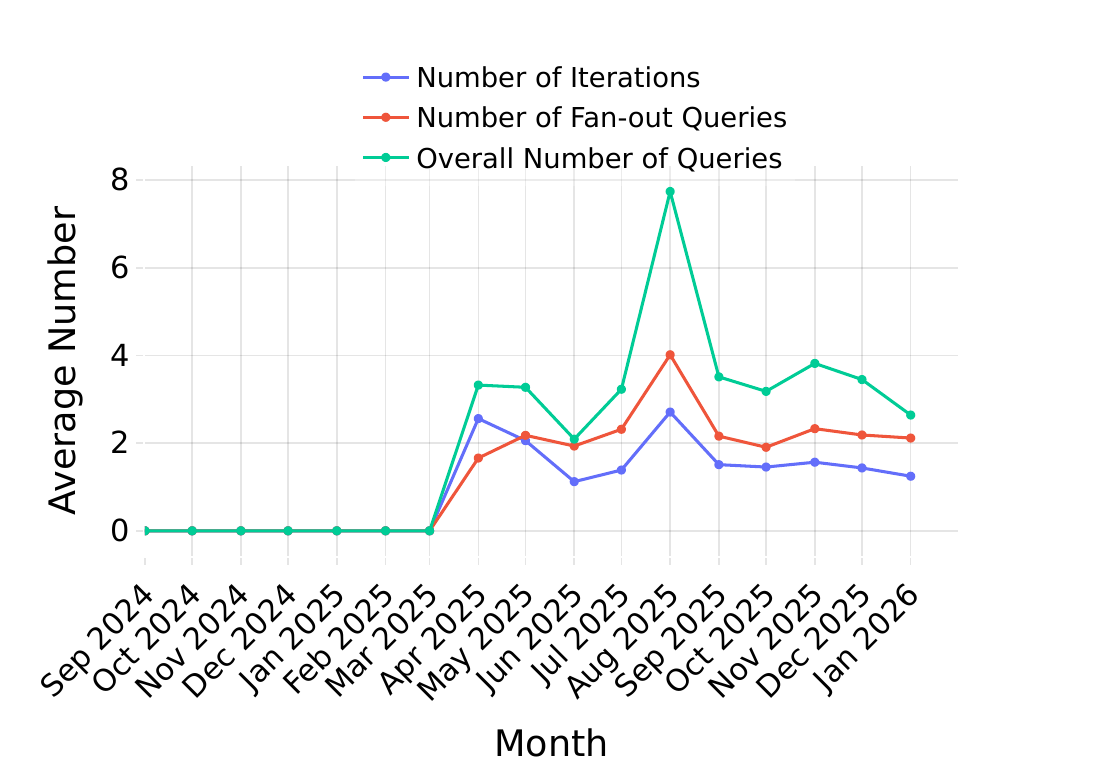}
        \caption{ChatGPT}
        \label{fig:query_comp_longitudinal_chatgpt}
    \end{subfigure}
    \begin{subfigure}[b]{0.48\linewidth}
        \centering
        \includegraphics[width=1\linewidth]{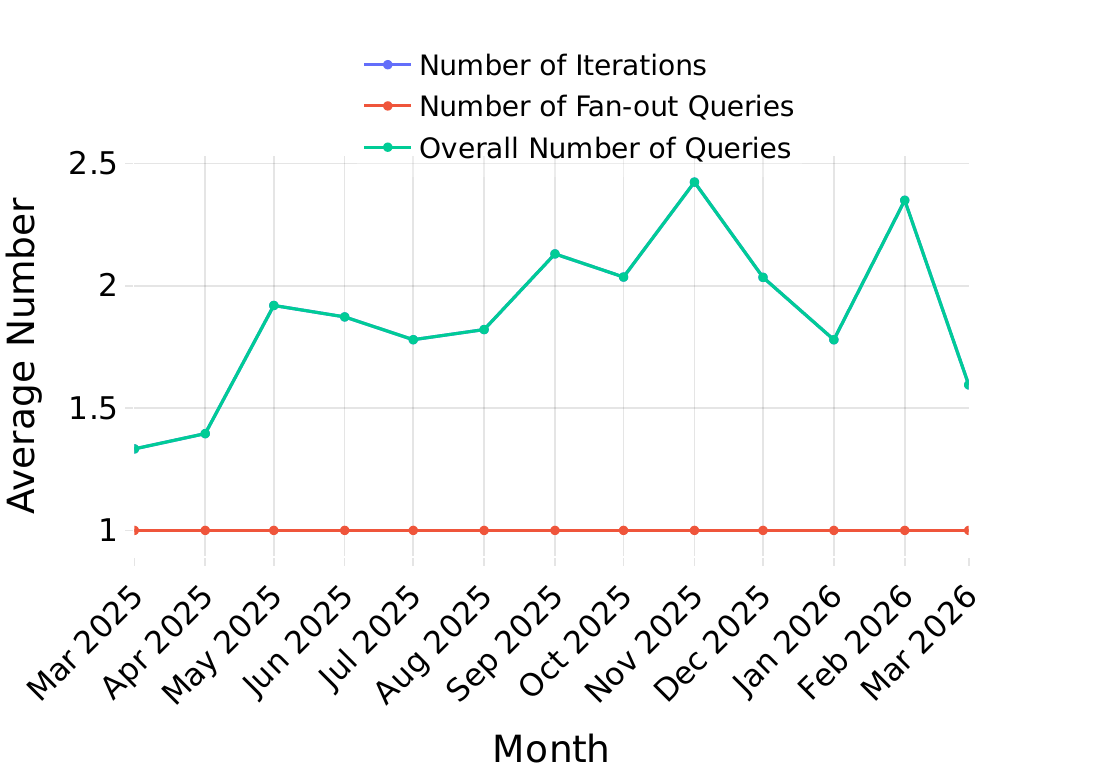}
        \caption{Claude}
        \label{fig:query_comp_longitudinal_claude}
    \end{subfigure}
    \begin{subfigure}[b]{0.48\linewidth}
        \centering
        \includegraphics[width=1\linewidth]{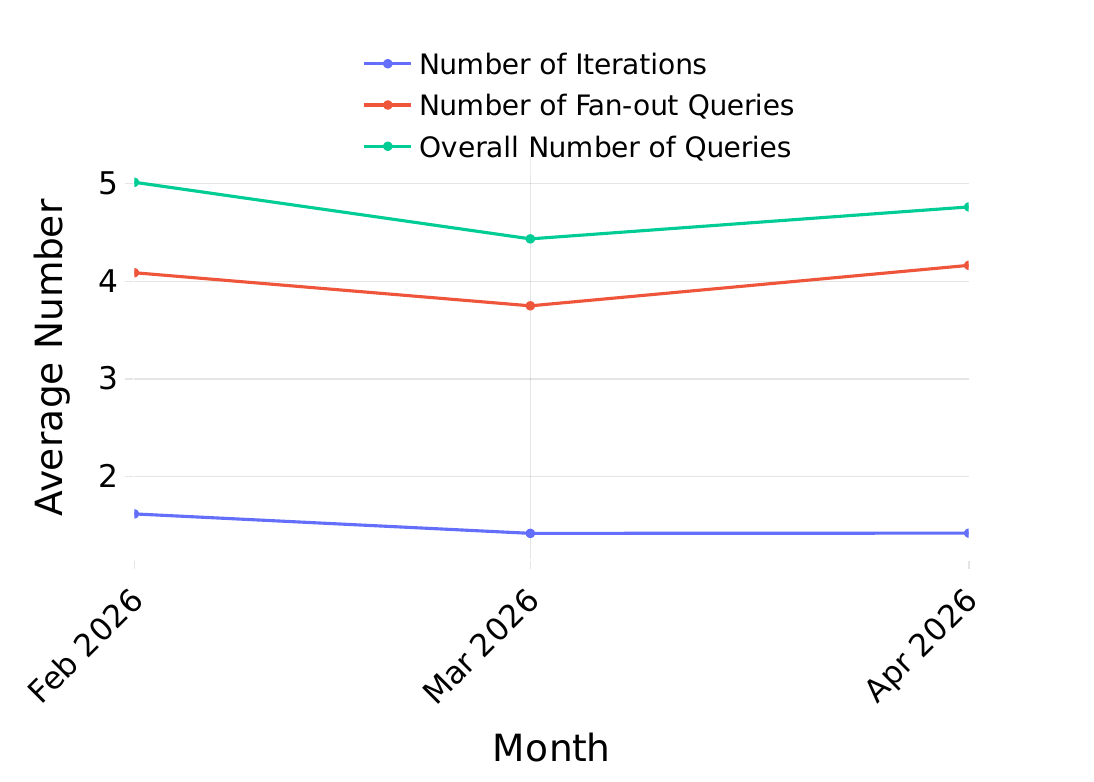}
        \caption{Grok}
        \label{fig:query_comp_longitudinal_grok}
    \end{subfigure}
    \begin{subfigure}[b]{0.48\linewidth}
        \centering
        \includegraphics[width=1\linewidth]{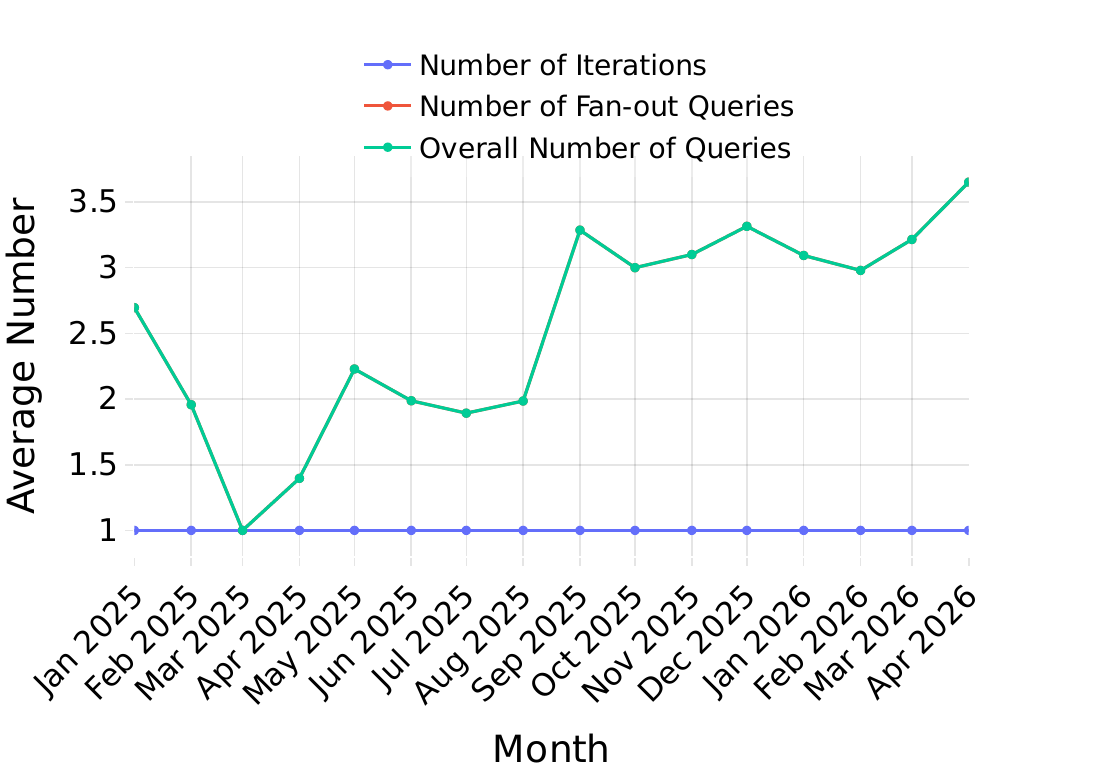}
        \caption{DeepSeek}
        \label{fig:query_comp_longitudinal_deepseek}
    \end{subfigure}
    \caption{Longitudinal evolution of querying strategies across platforms. The figure reports the average number of fan-out queries, iterative refinements, and total Web queries per response over time.}
    \label{fig:query_comp_longitudinal}
\end{figure*}

Figure~\ref{fig:query_comp_longitudinal} illustrates how querying strategies evolve over time across conversational platforms. We separately report the average number of fan-out queries, iterative refinements, and total Web queries per response. While the absolute number of Web queries changes across platform deployments, the overall querying behavior remains relatively stable. ChatGPT consistently issues more fan-out queries, whereas Claude relies primarily on iterative refinement.

\section{Citation Selection and Search Results}
\label{app:search_res_and_citations}

\subsection{Citation and Grounding Rates by Model}
\label{app:retrieval_citation_per_model}

In table~\ref{tab:grounding-rate-all-rev} we observed different citation rates for different platforms \invivo\ and \invitro. Table~\ref{tab:grounding-rate-all-rev_app} extends this analysis by additionally considering the samples that web call was issued for them across models \invitro{} (denoted as common). 

Table~\ref{tab:model-grounding-rates} extends the aggregate statistics reported in table~\ref{tab:grounding-rate-all-rev} by breaking citation down, along with \textit{Grounding Rate}=$\frac{N_{\text{Cited\&Search results}}}{N_{\text{Cited results}}}$), by individual deployed models within each platform \invivo.  Variation exists even among models belonging to the same provider,  indicating that model architecture and deployment policy substantially influence citation behavior.

\input{table}

\begin{table}[t]
\centering
\begin{adjustbox}{max width=\columnwidth}
\begin{tabular}{llrr}
\toprule
\textbf{Platform} &
\textbf{Model} &
\makecell{\textbf{Citation} \\ \textbf{Rate}} &
\makecell{\textbf{Grounding} \\ \textbf{Rate}} \\
\midrule
\multirow{13}{*}{ChatGPT}
& gpt-4-1                     & 18.6\% & 93.8\% \\
& gpt-4o                      & 17.5\% & 92.7\% \\
& gpt-5                       & 14.7\% & 98.5\% \\
& gpt-5-mini                  & 11.3\% & 91.6\% \\
& gpt-5-2                     & 7.8\%  & 99.9\% \\
& o3                          & 9.5\%  & 95.0\% \\
& text-davinci-002-render-sha & 20.0\% & 87.5\% \\
& gpt-4-1-mini                & 12.5\% & 90.3\% \\
& gpt-4o-mini                 & 14.2\% & 88.8\% \\
& gpt-5-instant               & 10.2\% & 100.0\% \\
& gpt-5-thinking              & 9.4\%  & 99.0\% \\
& gpt-5-2-thinking            & 6.0\%  & 97.5\% \\
\midrule
\multirow{1}{*}{Claude}
& --- & --- & --- \\
\midrule
\multirow{1}{*}{Grok}
& grok-3 & 1.3\% & 99.49\% \\
& grok-4 & 4.1\% & 99.87\% \\
& grok-420 & 0.6\% & 100.0\% \\
\midrule
\multirow{1}{*}{DeepSeek}
& deepseek-chat & 46.5\% & 100.0\% \\
& deepseek-reasoner & 27.4\% & 100.0\% \\
\bottomrule
\end{tabular}
\end{adjustbox}
\caption{Citation and grounding rates across deployed models within each platform in the \invivo{} dataset. Citation rate denotes the fraction of search result URLs that are eventually cited, while grounding rate denotes the fraction of citations originating from search results.}
\label{tab:model-grounding-rates}
\end{table}

\subsection{Longitudinal Search Results and Citation Trends}
\label{app:retrieval_citation_longitudinal}
\begin{figure*}
    \centering
    \begin{subfigure}[b]{0.48\linewidth}
        \centering
        \includegraphics[width=1\linewidth]{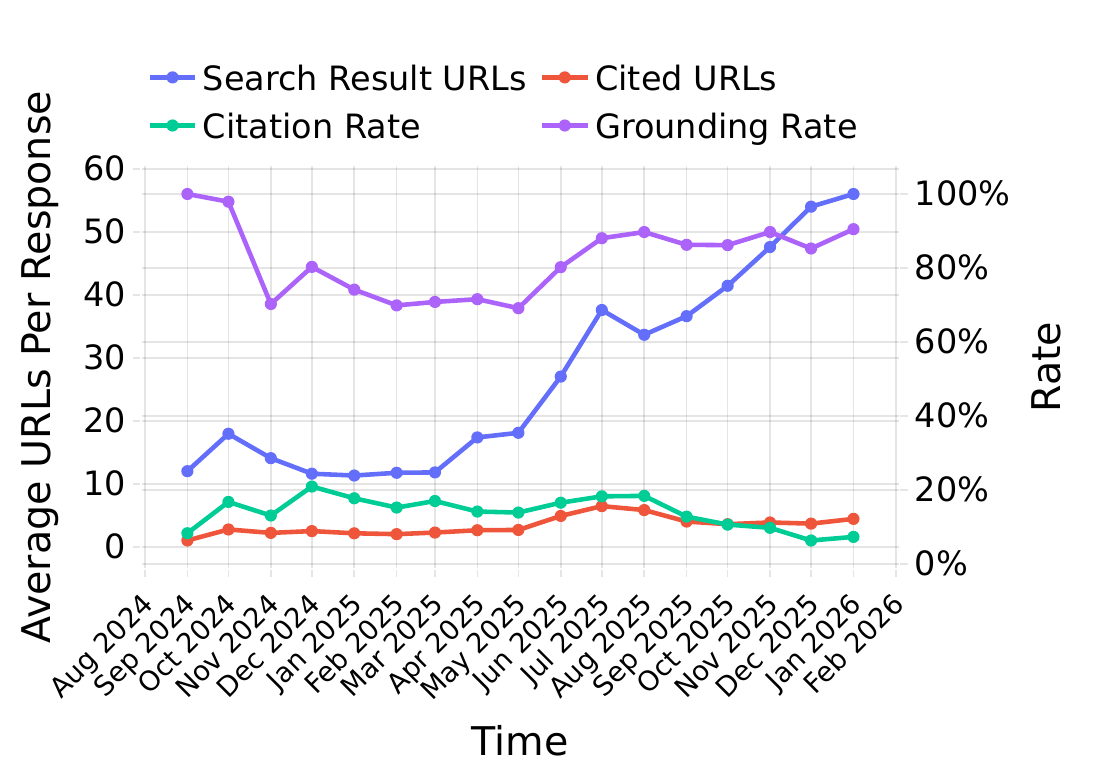}
        \caption{ChatGPT}
        \label{fig:citation_comp_longitudinal_openai}
    \end{subfigure}
    \begin{subfigure}[b]{0.48\linewidth}
        \centering
        \includegraphics[width=1\linewidth]{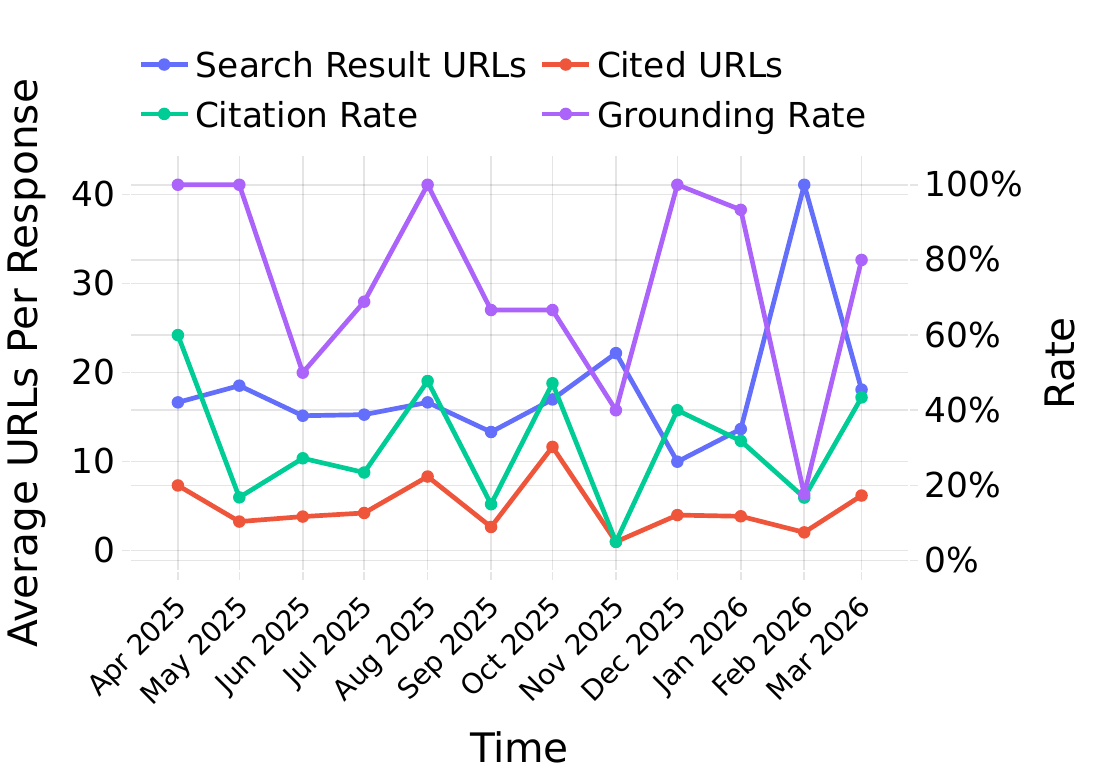}
        \caption{Claude}
        \label{fig:citation_comp_longitudinal_claude}
    \end{subfigure}
    \begin{subfigure}[b]{0.48\linewidth}
        \centering
        \includegraphics[width=1\linewidth]{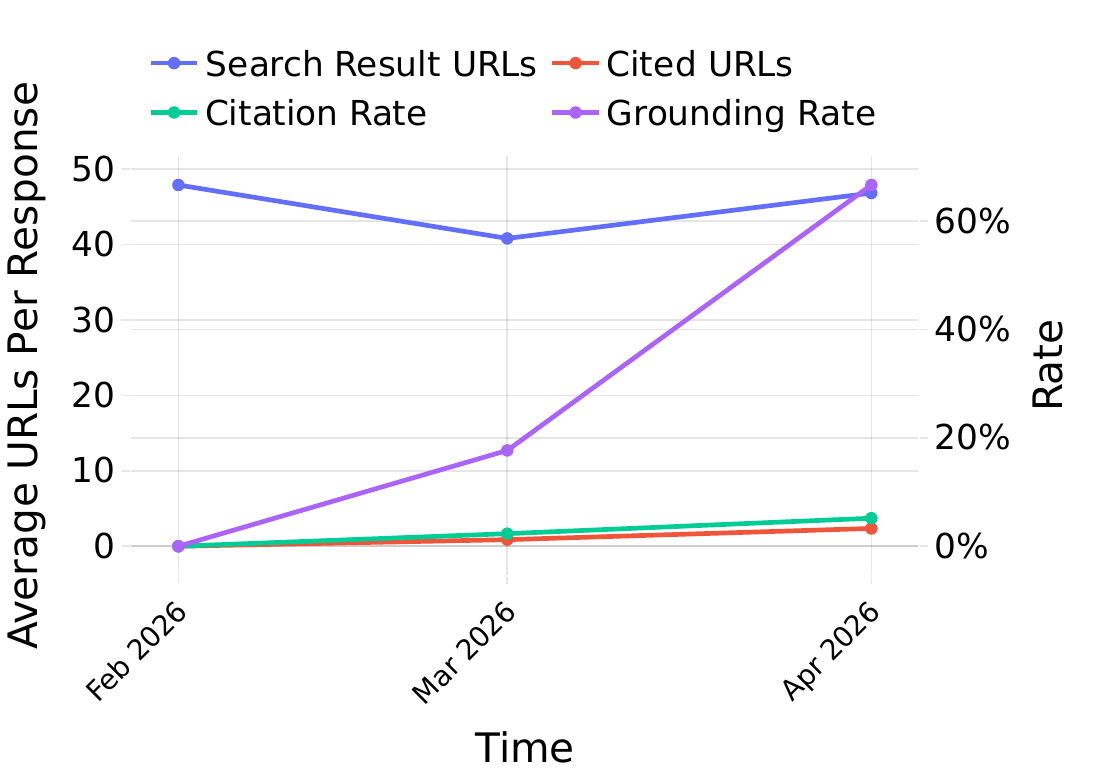}
        \caption{Grok}
        \label{fig:citation_comp_longitudinal_grok}
    \end{subfigure}
    \begin{subfigure}[b]{0.48\linewidth}
        \centering
        \includegraphics[width=1\linewidth]{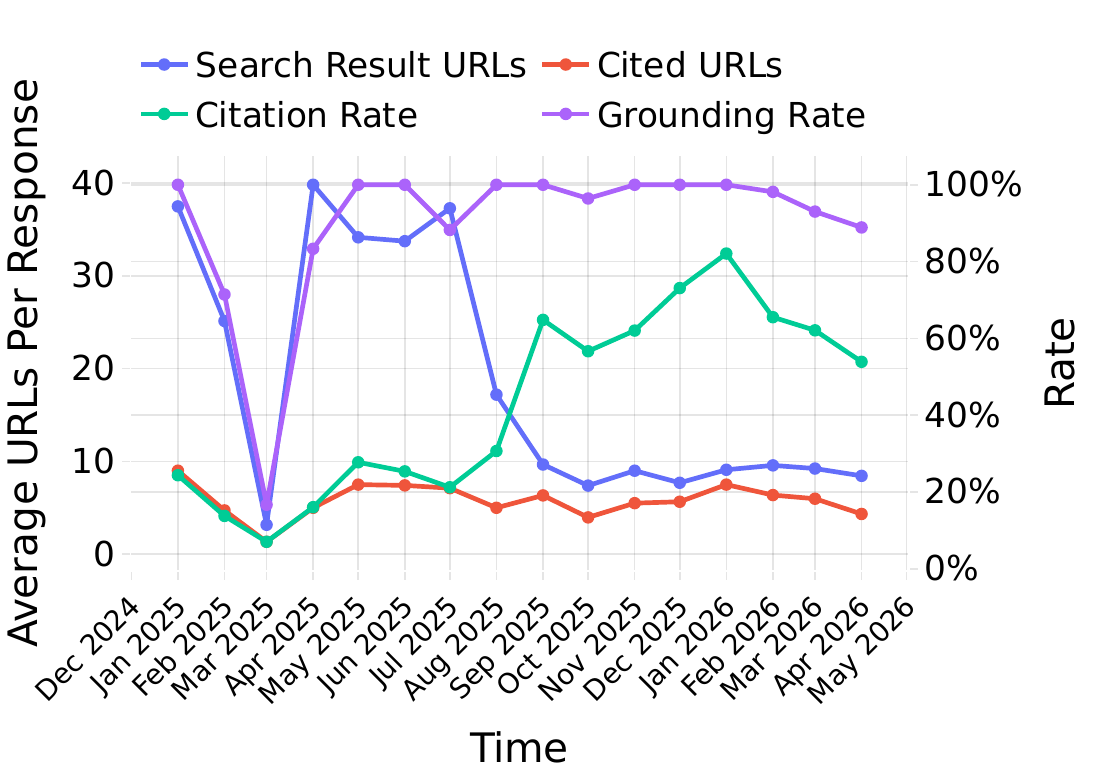}
        \caption{DeepSeek}
        \label{fig:citation_comp_longitudinal_deepseek}
    \end{subfigure}
    \caption{Longitudinal evolution of search and citation behavior. The figure reports search result URLs, cited URLs, citation rates, and grounding rates per response across time.}
    \label{fig:citation_comp_longitudinal}
\end{figure*}

Figure~\ref{fig:citation_comp_longitudinal} presents the temporal evolution of search results and citation behavior across platforms. We report the average number of search result URLs, cited URLs, citation rates, and grounding rates over time. While the volume of search result varies substantially as platforms evolve, citations remain comparatively stable, suggesting that citation selection policies change more slowly than retrieval strategies.

%% file: table.tex
\begin{table}[t]
\centering
\Large
\renewcommand{\arraystretch}{1.15}
\begin{adjustbox}{max width=\columnwidth}
\begin{tabular}{p{2cm}lrrrrr}
\toprule
\textbf{Category} &
\makecell{\textbf{Platform}\\\textbf{/ Model}} &
\makecell{\textbf{\#Search Results}\\\textbf{URLs}} &
\makecell{\textbf{\#Cited}\\\textbf{URLs}} &
\makecell{\textbf{\#Cited Search}\\\textbf{Results}} &
\makecell{\textbf{Citation}\\\textbf{Rate (95\% CI)}}\\
\midrule

\multirow{4}{*}{\invivo}
& ChatGPT
& \makecell[c]{\textbf{1,196,238}\\{\normalsize(28.4)}}
& \makecell[c]{\textbf{168,510}\\{\normalsize(4.0)}}
& \textbf{159,580}
& \makecell[c]{\textbf{13.3\%}\\{\normalsize(13.1,13.6)}} \\

& Claude
& \makecell[c]{\textbf{27,162}\\{\normalsize(16.6)}}
& \makecell[c]{\textbf{5,668}\\{\normalsize(3.5)}}
& \textbf{5,409}
& \makecell[c]{\textbf{19.9\%}\\{\normalsize(17.2,22.9)}} \\

& Grok
& \makecell[c]{\textbf{110,684}\\{\normalsize(40.7)}}
& \makecell[c]{\textbf{1,950}\\{\normalsize(0.7)}}
& \textbf{1,943}
& \makecell[c]{\textbf{1.8\%}\\{\normalsize(1.4,2.2)}} \\

& DeepSeek
& \makecell[c]{\textbf{22,338}\\{\normalsize(14.4)}}
& \makecell[c]{\textbf{7,610}\\{\normalsize(4.9)}}
& \textbf{7,610}
& \makecell[c]{\textbf{34.1\%}\\{\normalsize(31.6,36.7)}} \\

\midrule

\multirow{4}{*}{\makecell{\invitro\\(all)}}
& GPT-5.3-chat
& \makecell[c]{\textbf{4,940}\\{\normalsize(35.3)}}
& \makecell[c]{\textbf{534}\\{\normalsize(3.8)}}
& \textbf{526}
& \makecell[c]{\textbf{10.6\%}\\{\normalsize(9.3,12.0)}} \\

& Claude Sonnet 4.6
& \makecell[c]{\textbf{17,086}\\{\normalsize(20.7)}}
& \makecell[c]{\textbf{6,790}\\{\normalsize(8.2)}}
& \textbf{6,790}
& \makecell[c]{\textbf{39.7\%}\\{\normalsize(37.1,42.6)}} \\

& Grok-4.3
& \makecell[c]{\textbf{17,090}\\{\normalsize(22.3)}}
& \makecell[c]{\textbf{4,406}\\{\normalsize(5.7)}}
& \textbf{4,364}
& \makecell[c]{\textbf{25.5\%}\\{\normalsize(24.2,26.9)}} \\

& DeepSeek-v4-flash
& \makecell[c]{\textbf{10,582}\\{\normalsize(18.1)}}
& --
& --
& -- \\

\midrule

\multirow{4}{*}{\makecell{\invitro\\(common)}}
& GPT-5.3-chat
& \makecell[c]{\textbf{4,702}\\{\normalsize(36.4)}}
& \makecell[c]{\textbf{509}\\{\normalsize(3.9)}}
& \textbf{501}
& \makecell[c]{\textbf{10.7\%}\\{\normalsize(9.3,12.2)}} \\

& Claude Sonnet 4.6
& \makecell[c]{\textbf{3,177}\\{\normalsize(24.6)}}
& \makecell[c]{\textbf{1,009}\\{\normalsize(7.8)}}
& \textbf{1,009}
& \makecell[c]{\textbf{31.8\%}\\{\normalsize(26.1,39.3)}} \\

& Grok-4.3
& \makecell[c]{\textbf{3,156}\\{\normalsize(24.4)}}
& \makecell[c]{\textbf{795}\\{\normalsize(6.2)}}
& \textbf{771}
& \makecell[c]{\textbf{24.4\%}\\{\normalsize(21.4,27.8)}} \\

& DeepSeek-v4-flash
& \makecell[c]{\textbf{2,271}\\{\normalsize(17.6)}}
& --
& --
& -- \\

\bottomrule
\end{tabular}
\end{adjustbox}

\caption{Citation rates (reported with 95\% bootstrap confidence intervals) across platforms in \invivo\ and models in \invitro{} shows high variability. Numbers in parentheses denote the average number of search results or cited URLs per response.}
\label{tab:grounding-rate-all-rev_app}
\end{table}

%% file: hallucinated_urls.tex
\section{Hallucinated URL Detection}
\label{sec:hallucinated_url_detection}

We detect hallucinated URLs using a four-way classification framework inspired by prior work on URL persistence and web citation reliability~\cite{rao2026detecting, onweller2026cited}. Our pipeline first evaluates whether a cited URL is currently reachable and then distinguishes between naturally decayed links and fabricated URLs using archival evidence from the Wayback Machine\footnote{\url{https://web.archive.org/}}.

First, we assess URL reachability over HTTP using the \texttt{curl\_cffi} library\footnote{\url{https://pypi.org/project/curl-cffi/}}. We send HTTP requests with redirects enabled and classify URLs that return successful responses (2xx status codes) as \textit{Valid}.
For URLs that return explicit non-existence signals, specifically HTTP 404 or 410, we perform an additional archival verification step using the Wayback Machine. If an archived snapshot exists, we classify the URL as \textit{Dead}, indicating that the URL previously existed but later became unavailable. In contrast, if no archived snapshot is found, we classify the URL as \textit{Hallucinated}, a conservative lowerbound, suggesting that the URL was likely fabricated by the model and may never have existed.
As no archival service provides complete coverage of the web, rare edge cases may exist where the historical existence of a URL cannot be conclusively established from publicly available archival evidence alone. Therefore, URLs without archival evidence should be interpreted as likely, rather than definitively, hallucinated.

Finally, URLs that cannot be conclusively classified are assigned to an \textit{Unknown} category. This category includes network failures, timeouts, excessive redirects, bot-blocking mechanisms, and transient 5xx server errors.  We intentionally exclude these cases from hallucination judgments because they do not provide reliable evidence of non-existence.

%% file: nli.tex
\section{Grounding and Entailment Analysis}
\label{sec:app-nli}

This section complements Section~\ref{sec:assembly_model_responses} by providing additional entailment analysis and validation for response claims and their associated citations.




\subsection{Claim Factuality Evaluation}
\label{sec:nli_factuality_evaluation}
To evaluate the factuality of individual response claims, we use \texttt{GPT-4o-mini} with Web search enabled to assess each claim on a 5-point Likert scale. The evaluation prompts are provided in Section~\ref{sec:prompts_claim_nli}. 
We group claims according to the source of evidence supporting them---\emph{cited parametric evidence}, \emph{cited search result evidence}, \emph{search result evidence}, and \emph{parametric knowledge}---and compute the average factuality score for each category. To quantify uncertainty, we estimate 95\% bootstrap confidence intervals by resampling responses with replacement. The resulting factuality scores and confidence intervals are reported in Table~\ref{tab:claim_factuality}.

\begin{table}[t]
\centering
\resizebox{\columnwidth}{!}{
\begin{tabular}{llr}
\toprule
\textbf{Platform} &
\makecell{\textbf{Supporting} \\ \textbf{Evidence}} &
\makecell{\textbf{Avg. Factuality} \\ \textbf{Score (95\% CI)}} \\
\midrule

\multirow{4}{*}{ChatGPT}
& Associated Citations
& 3.47 {\color{black}\scriptsize(3.36--3.59)} \\

& Other Citations
& 3.35 {\color{black}\scriptsize(3.24--3.47)} \\

& Search Results
& 3.33 {\color{black}\scriptsize(3.19--3.48)} \\

& Parametric Knowledge
& 2.84 {\color{black}\scriptsize(2.71--2.97)} \\

\midrule

\multirow{4}{*}{Claude}
& Associated Citations
& 3.92 {\color{black}\scriptsize(3.56--4.30)} \\

& Other Citations
& 3.53 {\color{black}\scriptsize(3.07--4.18)} \\

& Search Results
& 3.64 {\color{black}\scriptsize(3.21--4.06)} \\

& Parametric Knowledge
& 3.31 {\color{black}\scriptsize(2.51--4.20)} \\

\midrule

\multirow{4}{*}{Grok}
& Associated Citations
& 4.40 {\color{black}\scriptsize(4.21--4.58)} \\

& Other Citations
& 4.35 {\color{black}\scriptsize(4.18--4.49)} \\

& Search Results
& 4.26 {\color{black}\scriptsize(4.18--4.33)} \\

& Parametric Knowledge
& 3.40 {\color{black}\scriptsize(3.25--3.52)} \\

\midrule

\multirow{4}{*}{DeepSeek}
& Associated Citations
& 4.17 {\color{black}\scriptsize(4.08--4.25)} \\

& Other Citations
& 4.31 {\color{black}\scriptsize(4.21--4.40)} \\

& Search Results
& 4.16 {\color{black}\scriptsize(3.95--4.37)} \\

& Parametric Knowledge
& 3.92 {\color{black}\scriptsize(3.80--4.03)} \\

\bottomrule
\end{tabular}
}
\caption{Average factuality scores of response claims grouped by the source of supporting evidence. Values are reported as mean with 95\% bootstrap confidence intervals obtained by resampling responses.}
\label{tab:claim_factuality}
\end{table}

\begin{table}[t]
\centering
\resizebox{\columnwidth}{!}{
\begin{tabular}{llr}
\toprule
\textbf{Model} &
\makecell{\textbf{Supporting} \\ \textbf{Evidence}} &
\makecell{\textbf{Avg. Factuality} \\ \textbf{Score (95\% CI)}} \\
\midrule

\multirow{4}{*}{GPT-5.3}
& Associated Citations
& 3.47 {\color{black}\scriptsize(3.28--3.65)} \\

& Other Citations
& 3.07 {\color{black}\scriptsize(2.90--3.26)} \\

& Search Results
& 2.88 {\color{black}\scriptsize(2.74--3.03)} \\

& Parametric Knowledge
& 2.60 {\color{black}\scriptsize(2.39--2.79)} \\

\midrule

\multirow{4}{*}{Claude Sonnet 4.6}
& Associated Citations
& 3.00 {\color{black}\scriptsize(2.84--3.16)} \\

& Other Citations
& 3.08 {\color{black}\scriptsize(2.90--3.26)} \\

& Search Results
& 2.94 {\color{black}\scriptsize(2.73--3.16)} \\

& Parametric Knowledge
& 2.66 {\color{black}\scriptsize(2.46--2.88)} \\

\midrule

\multirow{4}{*}{Grok-4.3}
& Associated Citations
& 3.41 {\color{black}\scriptsize(3.21--3.60)} \\

& Other Citations
& 3.23 {\color{black}\scriptsize(3.07--3.38)} \\

& Search Results
& 3.11 {\color{black}\scriptsize(2.85--3.37)} \\

& Parametric Knowledge
& 2.86 {\color{black}\scriptsize(2.68--3.06)} \\

\bottomrule
\end{tabular}
}
\caption{Average factuality scores of response claims grouped by the supporting evidence source in the \invitro\ replay experiments. Values are reported as mean with 95\% bootstrap confidence intervals obtained by resampling responses.}
\label{tab:claim_factuality_insitu}
\end{table}

\subsection{Entailment Validation}
\label{sec:nli_human_validtion}

To validate the entailment evaluation pipeline, we additionally conduct a human annotation study over 30 randomly sampled examples. Three annotators independently labeled whether response claims were supported by their associated cited sources. The annotations achieved an inter-annotator agreement of $78.2\%$, while agreement between human annotations and the automated judge reached $93.1\%$, indicating moderate-to-high consistency between automated and human entailment assessments.

%% file: other_platform_res.tex
\section{Additional \invivo{} Results}
\label{sec:app-other-platforms}

In this section, we show the results of our analysis on all platforms.

\subsection{Web Search Call Trends of AI Platforms}

\begin{figure*}[t]
\centering
    \begin{subfigure}{0.45\linewidth}
    \centering
    \includegraphics[width=1\linewidth]{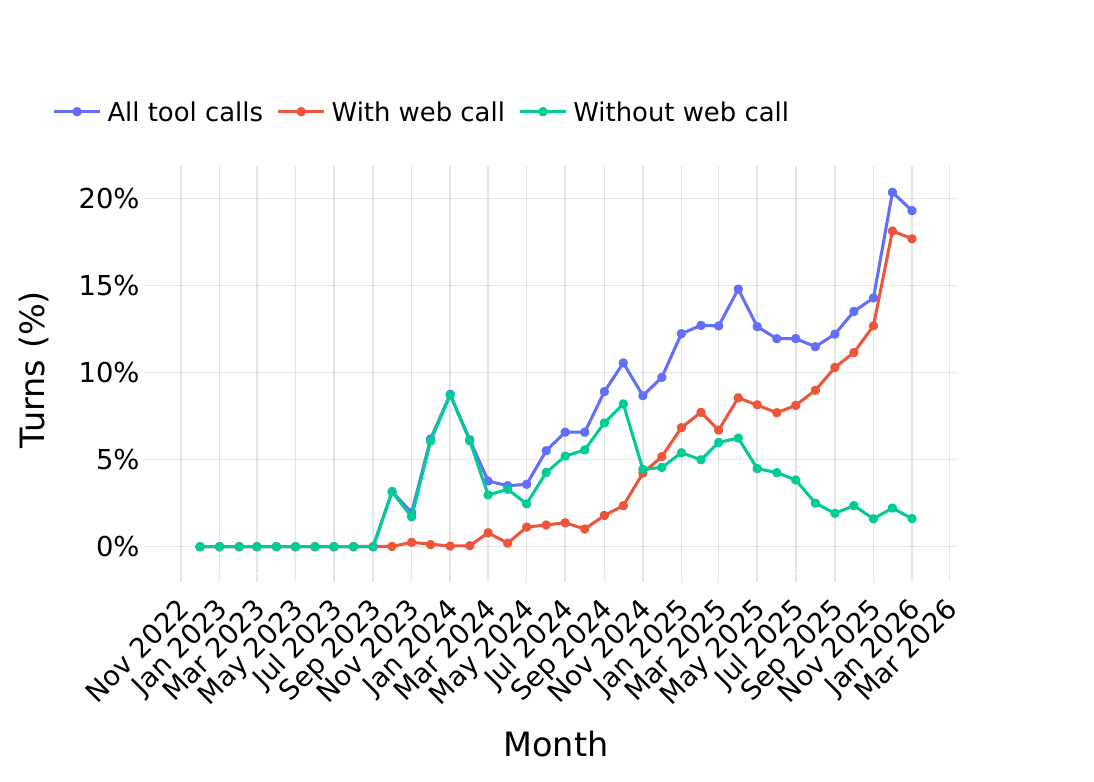}
    \caption{ChatGPT}
    \label{fig:tooly_turns_openai}
    \end{subfigure}
    \centering
    \begin{subfigure}{0.45\linewidth}
    \centering
    \includegraphics[width=1\linewidth]{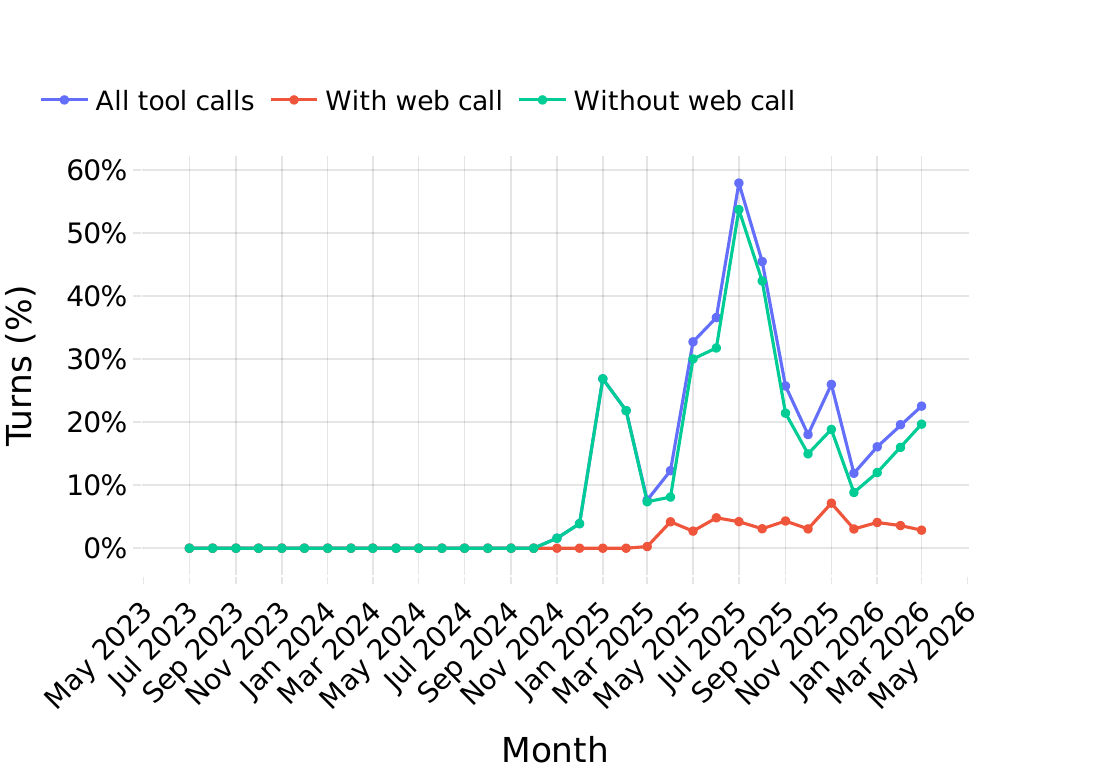}
    \caption{Claude}
    \label{fig:tooly_turns_claude}
    \end{subfigure}
    
    \centering
    \begin{subfigure}{0.45\linewidth}
    \centering
    \includegraphics[width=1\linewidth]{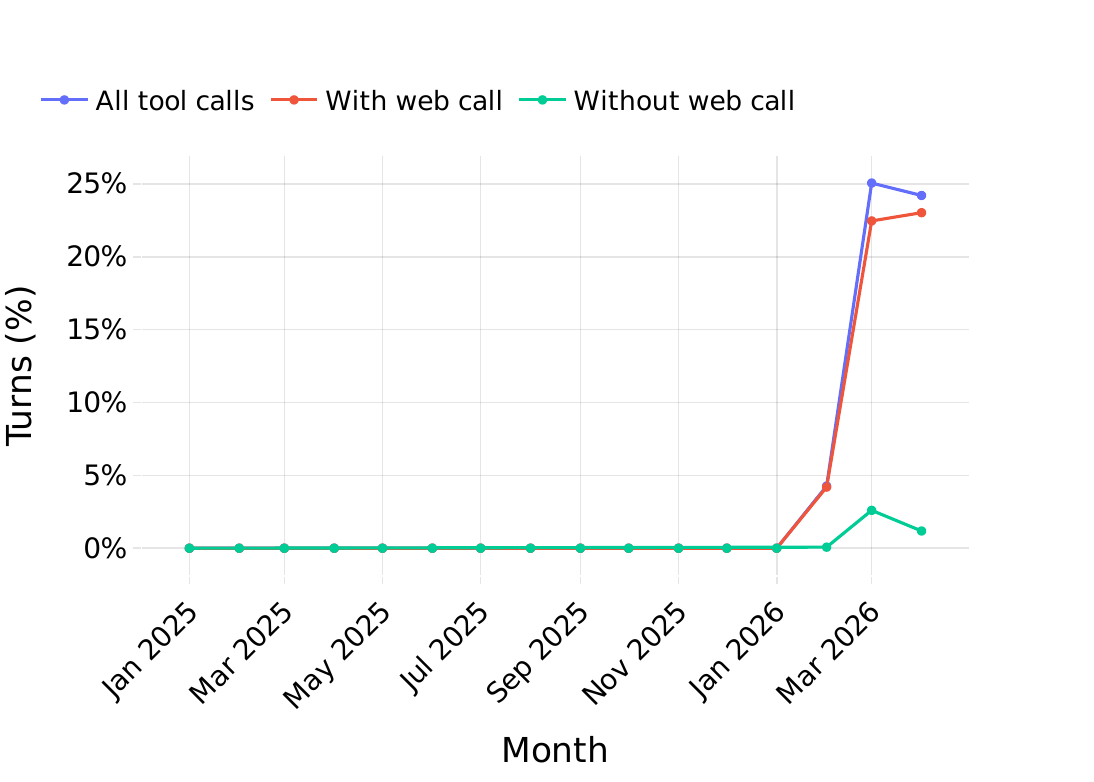}
    \caption{Grok}
    \label{fig:tooly_turns_grok}
    \end{subfigure}
    \centering
    \begin{subfigure}{0.45\linewidth}
    \centering
    \includegraphics[width=1\linewidth]{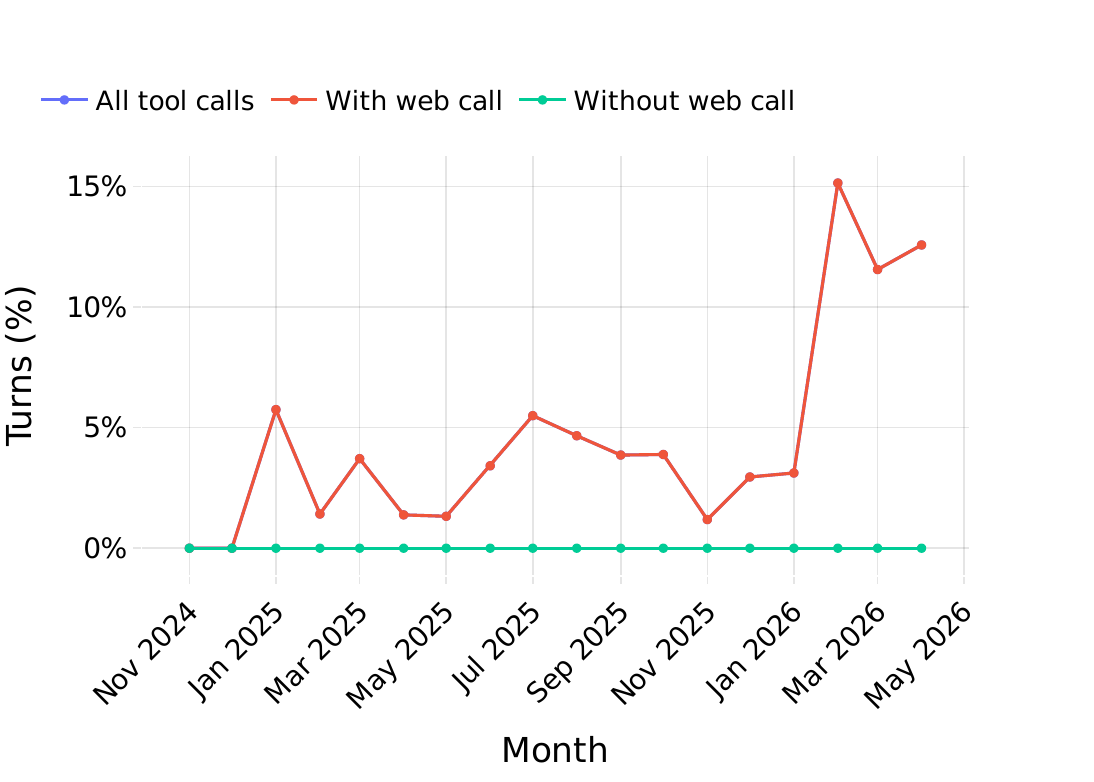}
    \caption{DeepSeek}
    \label{fig:tooly_turns_deepseek}
    \end{subfigure}
    
\caption{Fraction of turns invoking Web search and other tools over time across conversational platforms. Web-search calls increasingly dominate overall tool usage.}
\label{fig:tooly_turns_other_platforms}
\end{figure*}

\begin{figure*}[ht]
    \centering
    \begin{subfigure}{0.32\linewidth}
    \centering
    \includegraphics[width=1\linewidth]{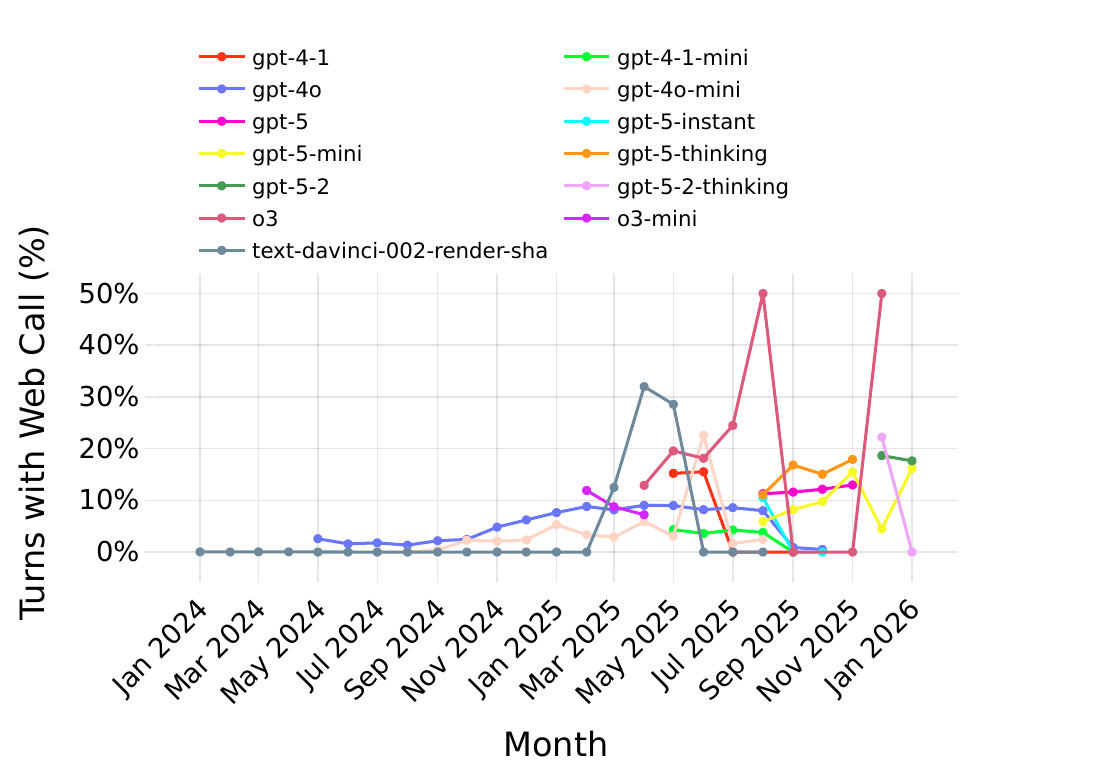}
    \caption{ChatGPT}
    \label{fig:web_call_trend_by_model_openai}
    \end{subfigure}
    \centering
    \begin{subfigure}{0.32\linewidth}
    \centering
    \includegraphics[width=1\linewidth]{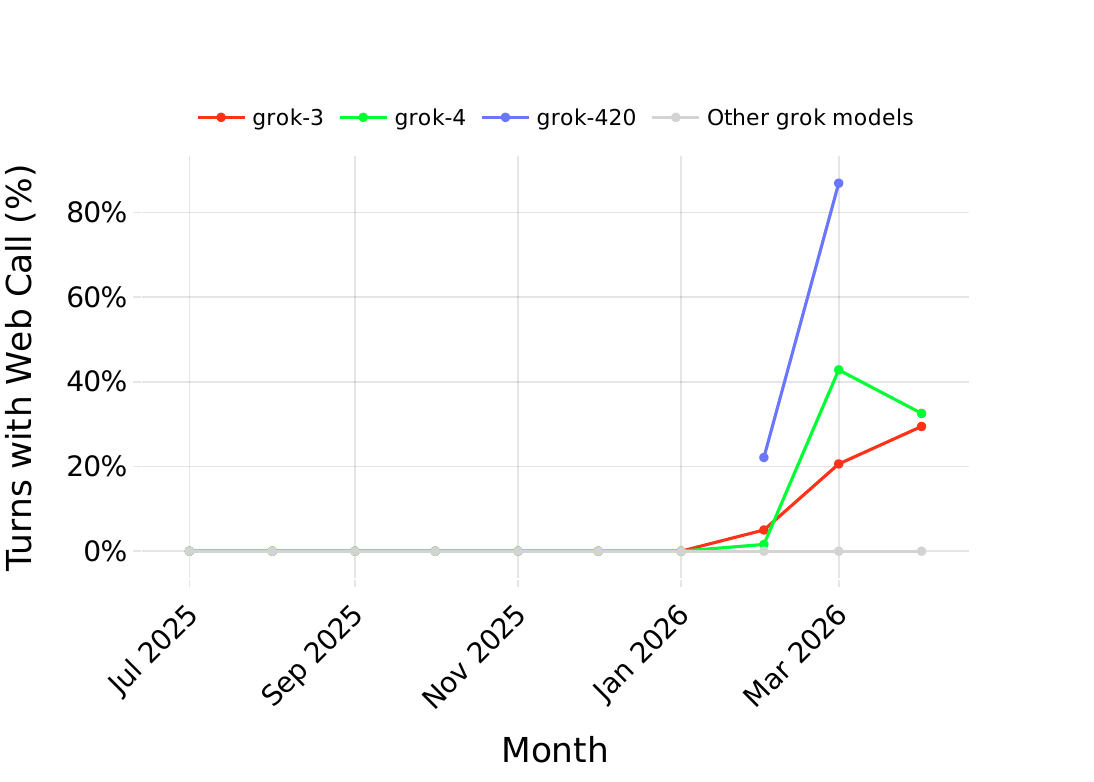}
    \caption{Grok}
    \label{fig:web_call_trend_by_model_grok}
    \end{subfigure}
    \centering
    \begin{subfigure}{0.32\linewidth}
    \centering
    \includegraphics[width=1\linewidth]{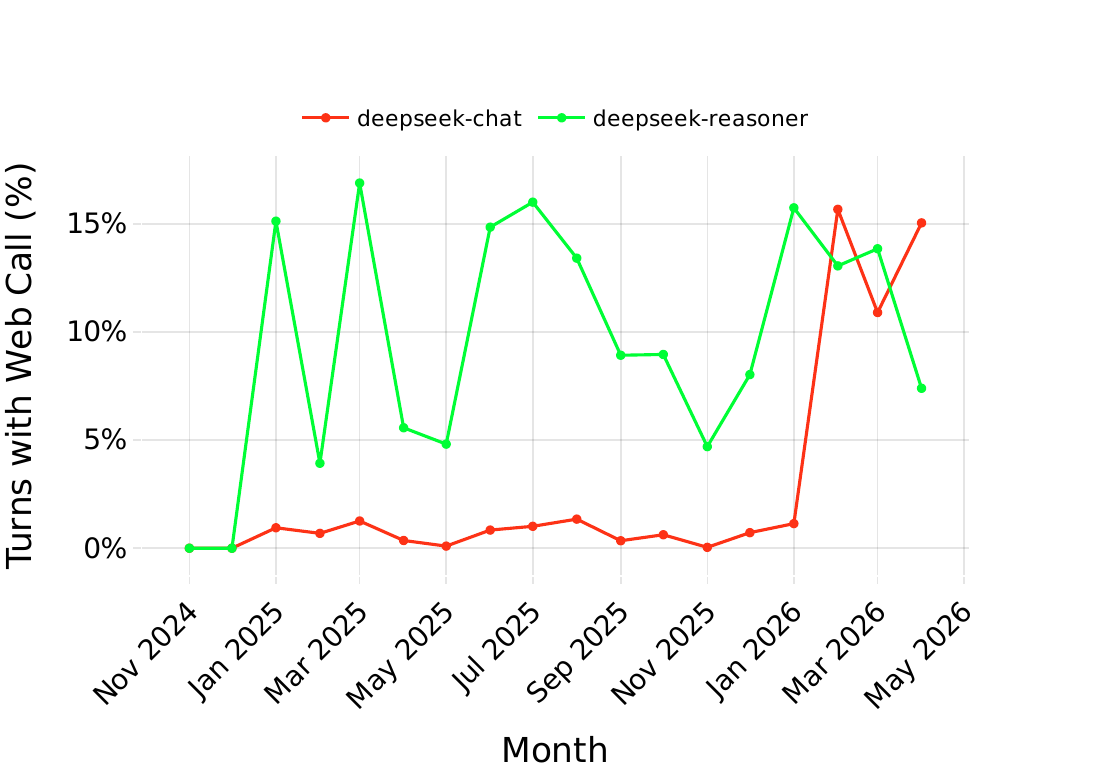}
    \caption{DeepSeek}
    \label{fig:web_call_trend_by_model_deepseek}
    \end{subfigure}
    
\caption{Fraction of turns invoking Web search across deployed models within each platform. Reasoning-oriented models consistently invoke Web search more frequently than non-reasoning models.}
\label{fig:web_call_trend_by_model_other_platforms}
\end{figure*}

Figure~\ref{fig:tooly_turns_other_platforms} shows the monthly share of turns involving tool calls across four platforms, broken down into web-search calls and non-web calls.
A common trend across all four platforms is the steady rise of web-calling turns over time, reflecting growing demand for grounding model outputs in search result web content.
ChatGPT adopted tool calling as early as 2023, and while web-search calls have continued to increase steadily, non-web tool calls have plateaued through 2026.
In contrast, non-web tool use on Claude grew sharply beginning in mid-2025, suggesting that user demand for tool functionality is not uniform across platforms.
Grok introduced web calls in early 2026, with adoption rising rapidly thereafter, while DeepSeek has supported web calling since 2024 and exhibits a sharp rise beginning in early 2026.
By early 2026, the gap between web-search calls and all tool calls nearly disappears for ChatGPT, Grok, and DeepSeek, whereas Claude continues to maintain a noticeable share of non-web tool calls.

Figure~\ref{fig:web_call_trend_by_model_other_platforms} further breaks down web-calling turns by model within each platform.
Across all three platforms, reasoning- or thinking-oriented models (e.g., o3, gpt-5-thinking, grok-4, deepseek-reasoner) consistently exhibit higher rates of web calling than non-reasoning models, suggesting that reasoning-capable models more readily leverage web search as part of their response generation.

\subsection{Topics Inducing Web search}


\begin{figure*}[ht]
\centering
    \begin{subfigure}{0.45\linewidth}
    \centering
    \includegraphics[width=1\linewidth]{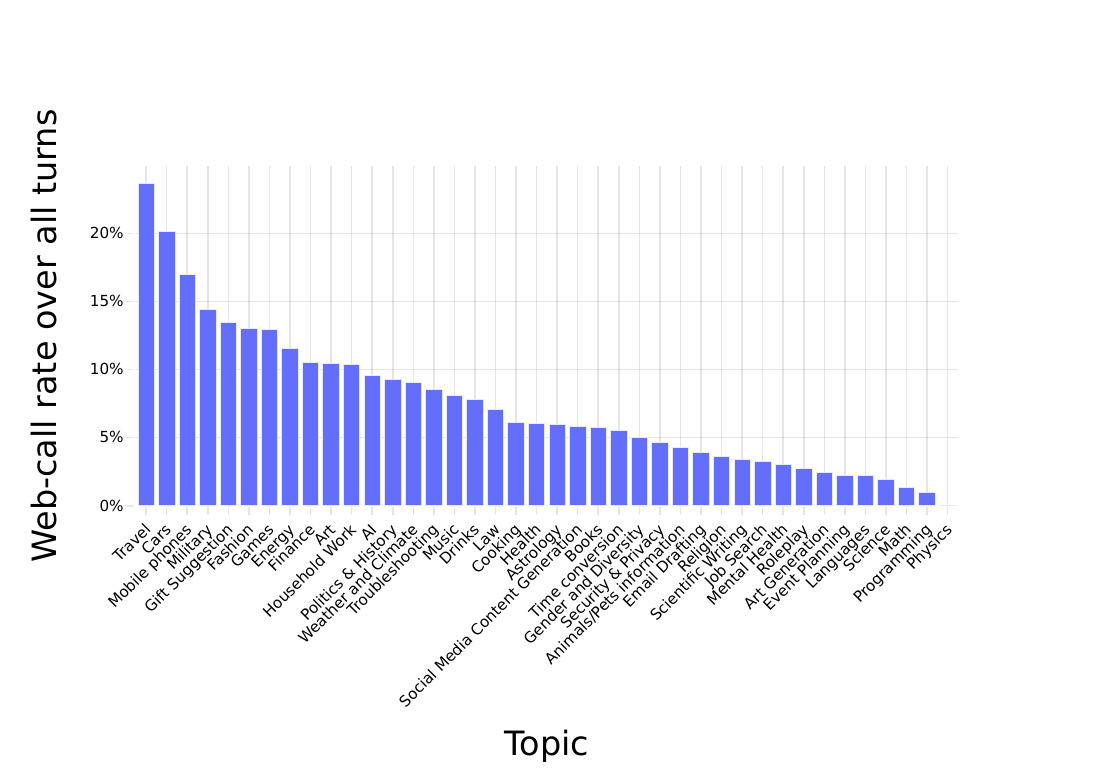}
    \caption{ChatGPT}
    \label{fig:topic_whole_openai}
    \end{subfigure}
    \centering
    \begin{subfigure}{0.45\linewidth}
    \centering
    \includegraphics[width=1\linewidth]{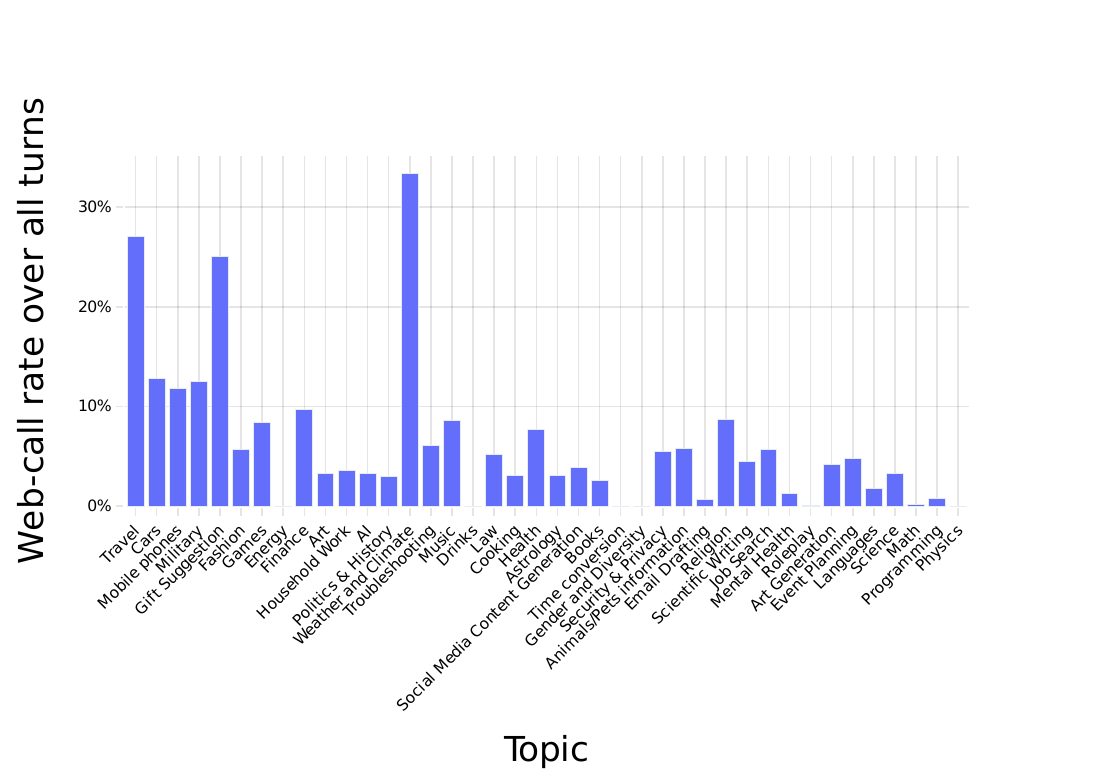}
    \caption{Claude}
    \label{fig:topic_whole_claude}
    \end{subfigure}
    
    \centering
    \begin{subfigure}{0.45\linewidth}
    \centering
    \includegraphics[width=1\linewidth]{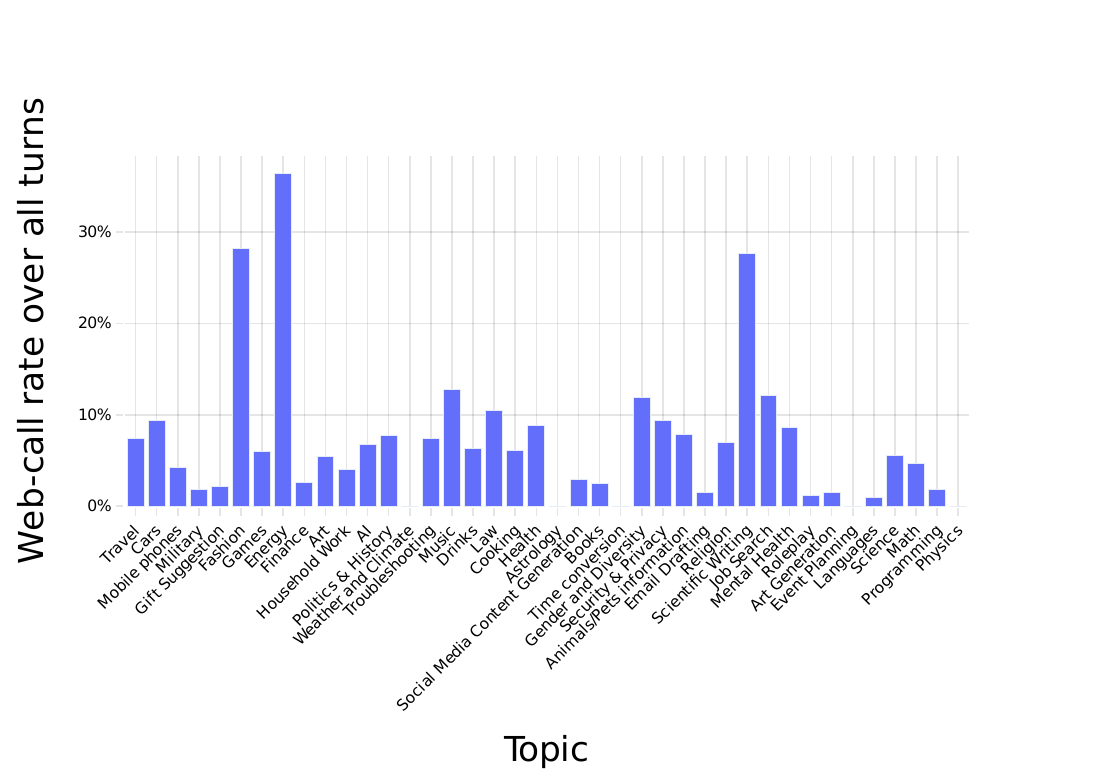}
    \caption{Grok}
    \label{fig:topic_whole_grok}
    \end{subfigure}
    \centering
    \begin{subfigure}{0.45\linewidth}
    \centering
    \includegraphics[width=1\linewidth]{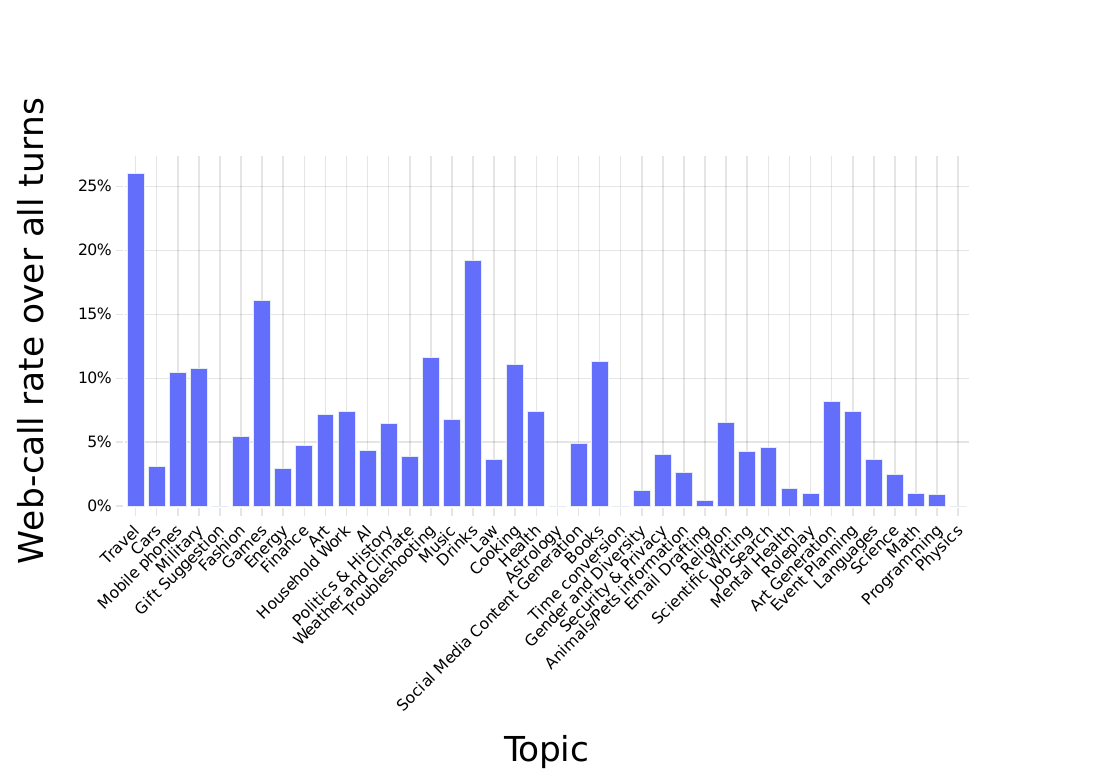}
    \caption{Deepseek}
    \label{fig:topic_whole_deepseek}
    \end{subfigure}
    
\caption{Fraction of conversational turns invoking Web search across topics. Topics requiring up-to-date or location-specific information consistently exhibit higher Web-search rates.}
\label{fig:topic_whole_other_platforms}
\end{figure*}

Figure~\ref{fig:topic_whole_other_platforms} shows the web-calling rate across topics on each platform.
A common pattern across all four platforms is that topics requiring up-to-date or location-specific information, such as Travel, Weather and Climate, Shopping, and Events, consistently exhibit the highest web-calling rates.
In contrast, topics that rely primarily on parametric knowledge or reasoning, such as Programming, Math, and Language, sit at the lower end of the distribution.
This indicates that web search is invoked selectively based on the informational needs of the topic rather than uniformly across all queries.
At the same time, the topic rankings differ noticeably across platforms.
On ChatGPT, Travel and Shopping-related topics dominate, whereas Claude shows the highest web-calling rates for Weather and Climate, Gift Suggestions, and Shopping.
Grok exhibits a distinctive profile in which Scientific Writing and Animals/Pets-related queries appear among the top, while DeepSeek concentrates web calls on Travel, Drinks, and Shopping.
These differences likely reflect both the user populations of each platform and the kinds of queries for which users perceive web grounding to be most beneficial.
%


\begin{figure*}
    \centering
    \begin{subfigure}[b]{0.48\linewidth}
        \centering
        \includegraphics[width=1\linewidth]{figures/web_tool_invocation/topic_web_call_rate_over_time.pdf}
        \caption{ChatGPT}
        \label{fig:topic_web_call_rate_over_time_chatgpt}
    \end{subfigure}
    \begin{subfigure}[b]{0.48\linewidth}
        \centering
        \includegraphics[width=1\linewidth]{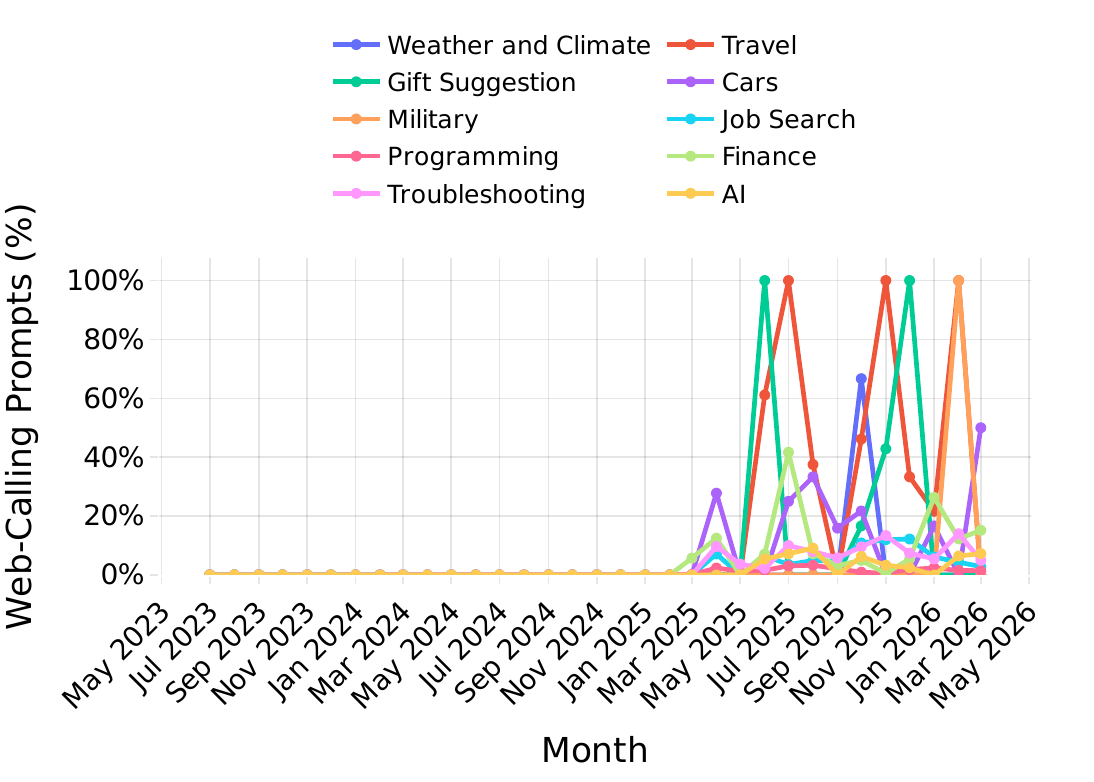}
        \caption{Claude}
        \label{fig:topic_web_call_rate_over_time_claude}
    \end{subfigure}
    \begin{subfigure}[b]{0.48\linewidth}
        \centering
        \includegraphics[width=1\linewidth]{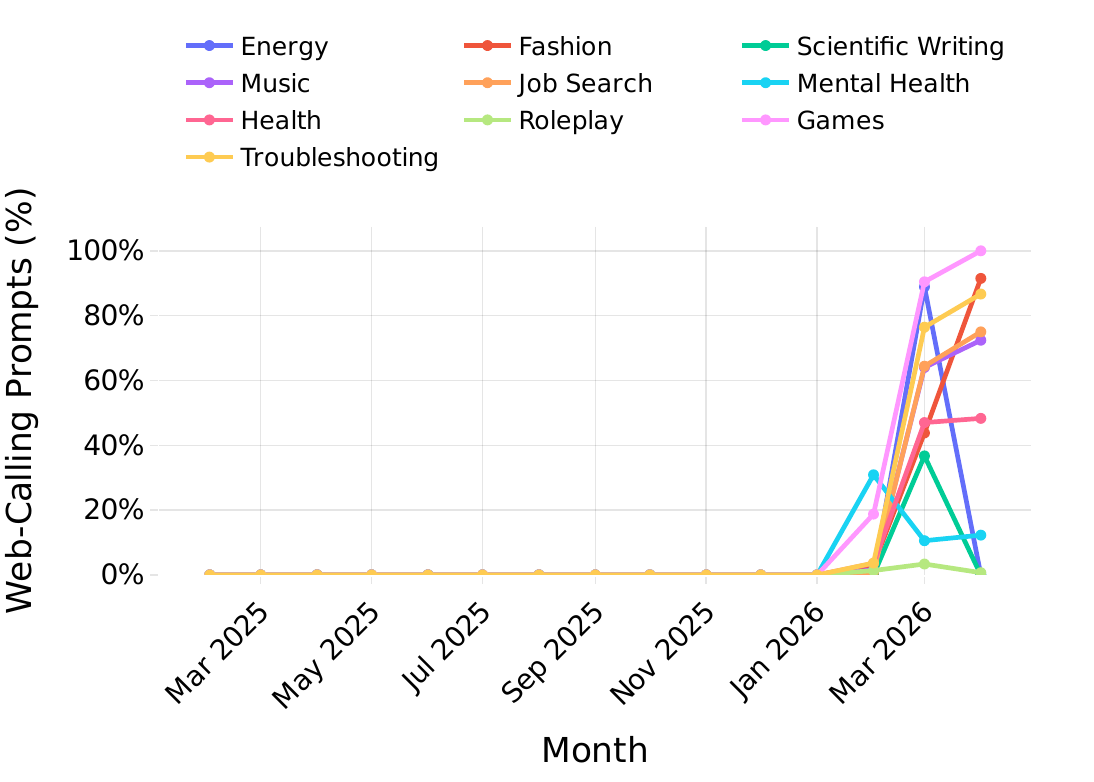}
        \caption{Grok}
        \label{fig:topic_web_call_rate_over_time_grok}
    \end{subfigure}
    \begin{subfigure}[b]{0.48\linewidth}
        \centering
        \includegraphics[width=1\linewidth]{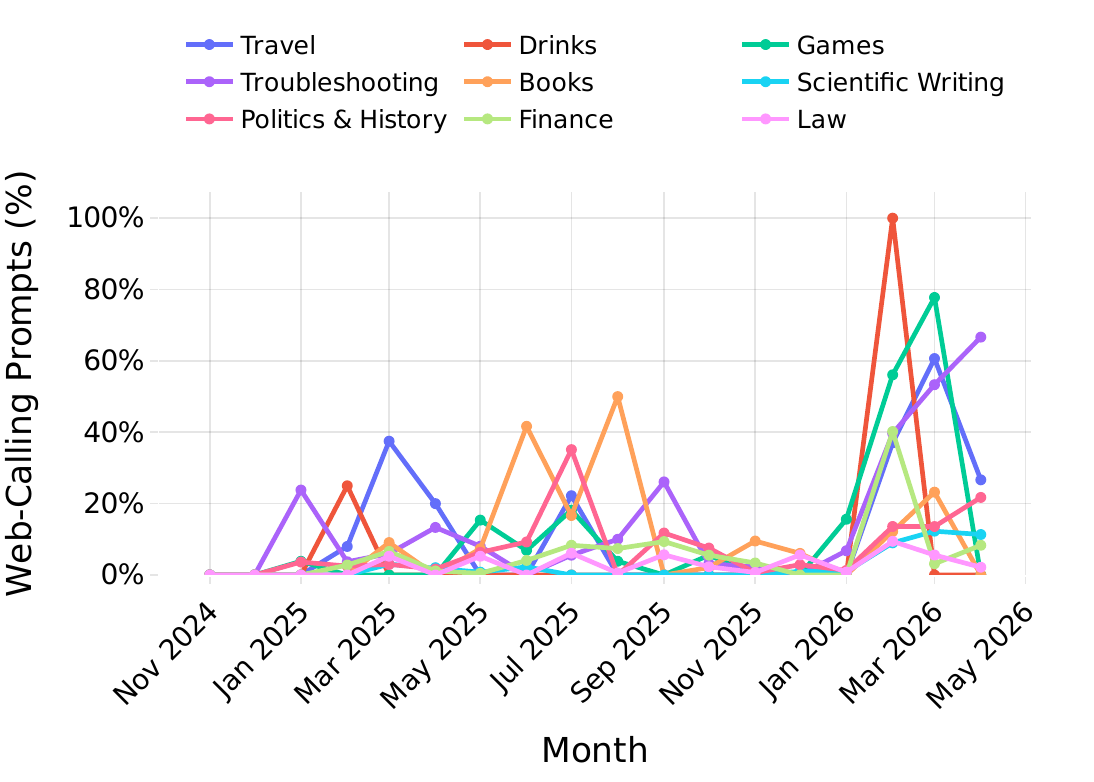}
        \caption{DeepSeek}
        \label{fig:topic_web_call_rate_over_time_deepseek}
    \end{subfigure}
    \caption{Longitudinal Web-search rates across topics. Although Web-search usage generally increases over time, substantial topic-dependent variation also persists across platforms.}
    \label{fig:topic_web_call_rate_over_time_all_platforms}
\end{figure*}

Figure~\ref{fig:topic_web_call_rate_over_time_all_platforms} shows that the rising trend in Web search usage generally holds across different conversation topics. However, the fraction of user prompts triggering Web search varies considerably by topic -- for example, Travel-related prompts invoke Web search much more frequently than Health-related prompts on ChatGPT. This variation is also platform-dependent: ChatGPT exhibits a relatively gradual and broad-based increase across topics, whereas Claude, Grok, and DeepSeek show sharper, topic-specific fluctuations, with some categories briefly approaching 100\%.

\subsection{Provenance of Web Query Reformulation} 

\begin{figure*}[ht]
\centering
\begin{subfigure}{0.31\linewidth}
    \centering
    \includegraphics[width=1\linewidth]{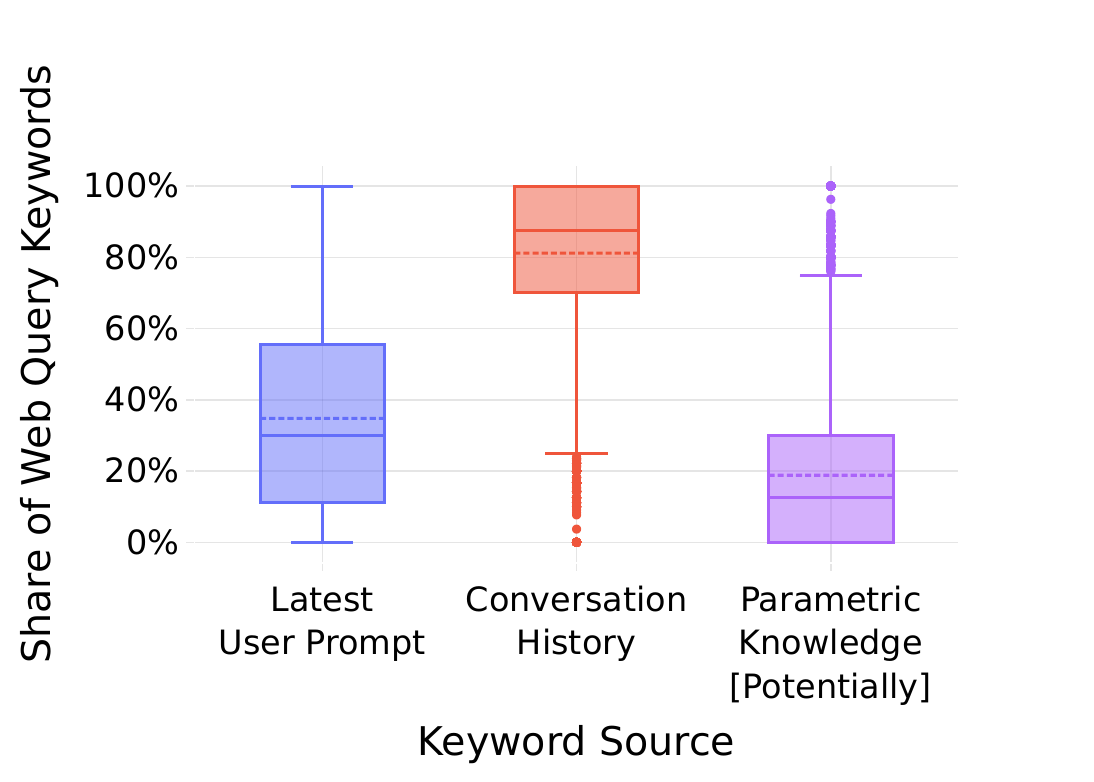}
    \caption{ChatGPT}
    \label{fig:query_source_1_openai}
    \end{subfigure}
    \centering
    \begin{subfigure}{0.31\linewidth}
    \centering
    \includegraphics[width=1\linewidth]{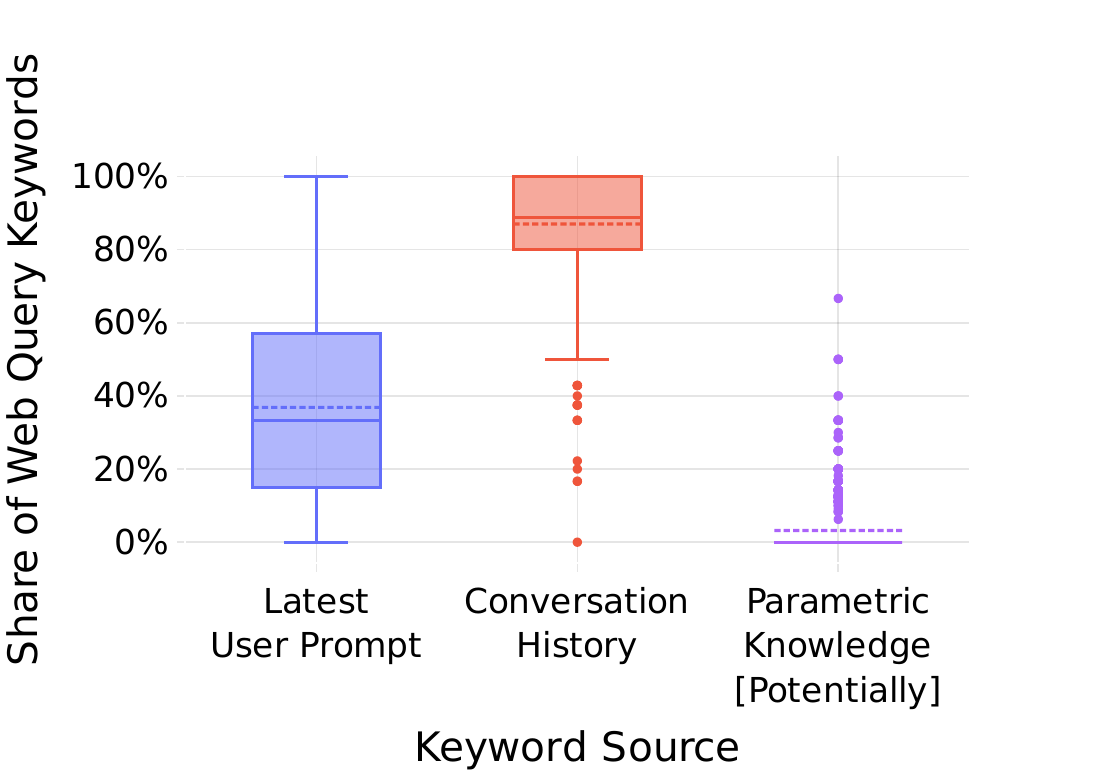}
    \caption{Claude}
    \label{fig:query_source_1_claude}
    \end{subfigure}
    \centering
    \begin{subfigure}{0.31\linewidth}
    \centering
    \includegraphics[width=1\linewidth]{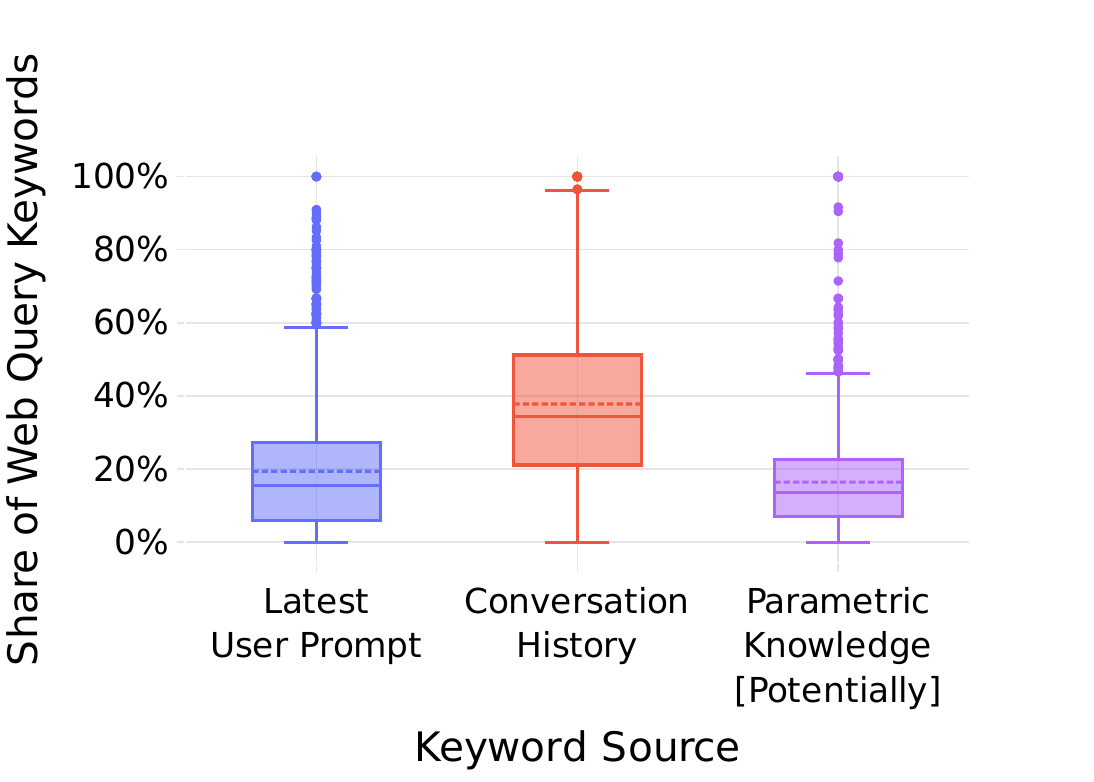}
    \caption{Grok}
    \label{fig:query_source_1_grok}
    \end{subfigure}
    
\caption{Source of new keywords in web queries in the first iteration across platforms.}
\label{fig:query_source_1_other_platforms}
\end{figure*}

\begin{figure*}[ht]
\centering
\begin{subfigure}{0.31\linewidth}
    \centering
    \includegraphics[width=1\linewidth]{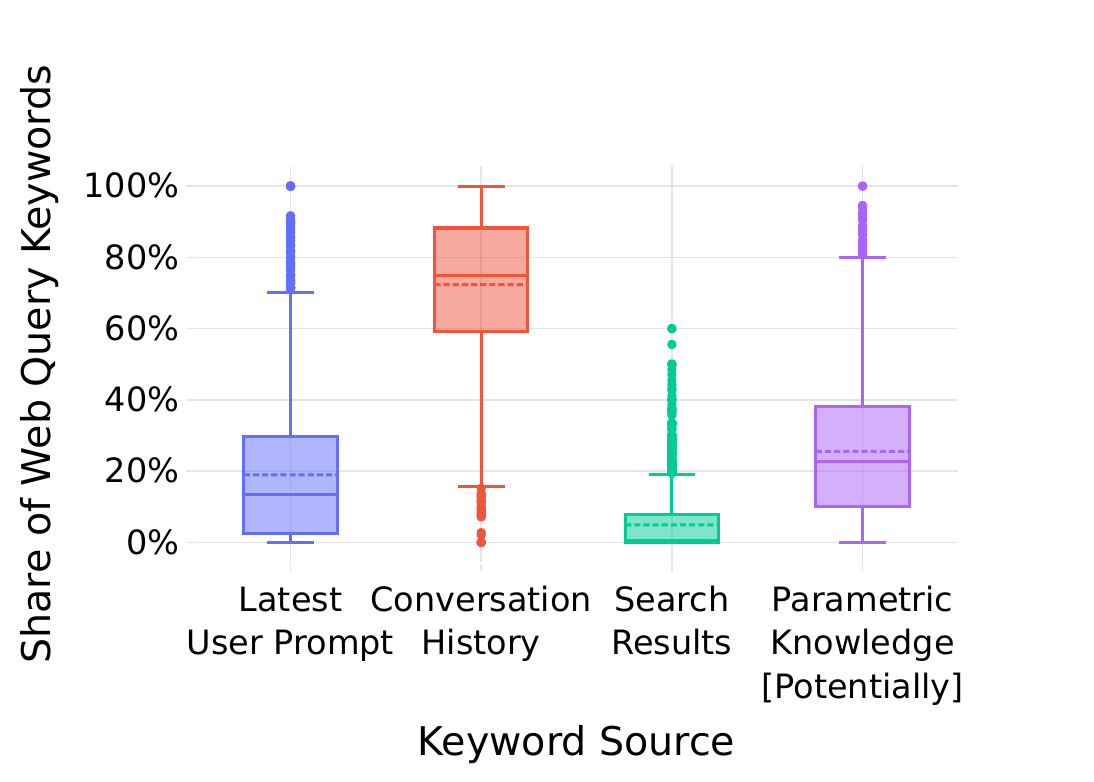}
    \caption{ChatGPT}
    \label{fig:query_source_multi_openai}
    \end{subfigure}
    \centering
    \begin{subfigure}{0.31\linewidth}
    \centering
    \includegraphics[width=1\linewidth]{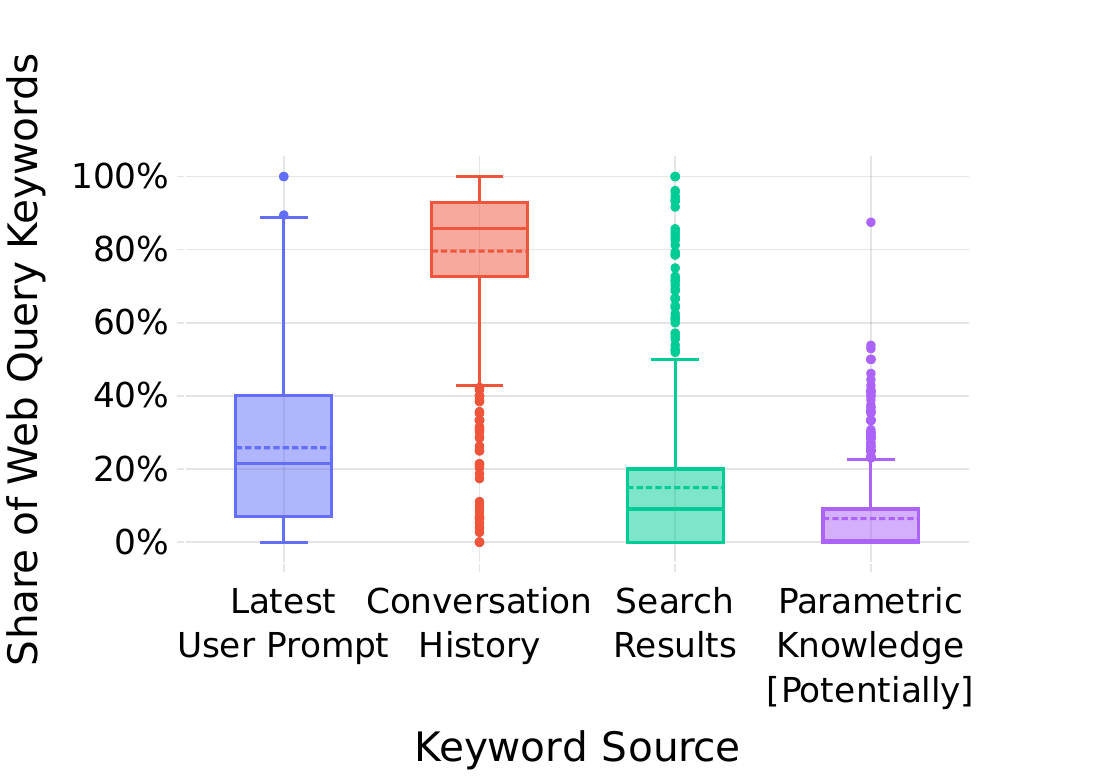}
    \caption{Claude}
    \label{fig:query_source_multi_claude}
    \end{subfigure}
    \centering
    \begin{subfigure}{0.31\linewidth}
    \centering
    \includegraphics[width=1\linewidth]{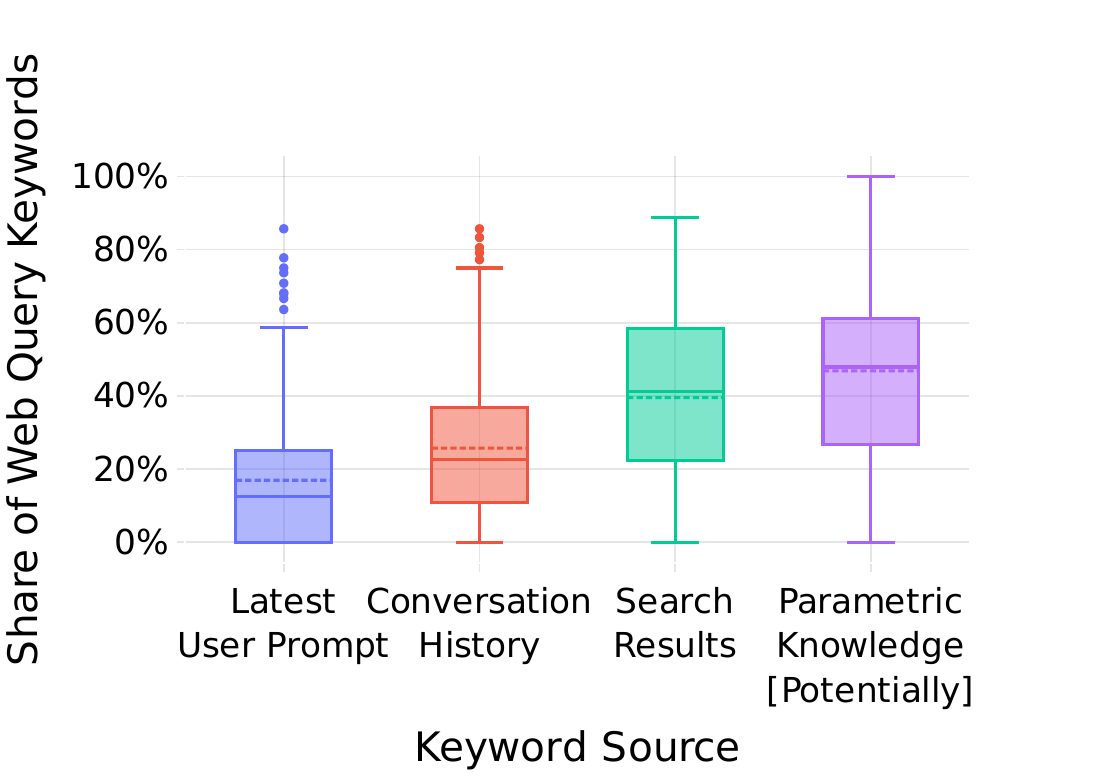}
    \caption{Grok}
    \label{fig:query_source_multi_grok}
    \end{subfigure}
    
\caption{Sources of newly introduced keywords during iterative query reformulation.  Search results increasingly contribute to later query formulations, particularly on Grok.}
\label{fig:query_source_multi_other_platforms}
\end{figure*}

Figures~\ref{fig:query_source_1_other_platforms} and~\ref{fig:query_source_multi_other_platforms} show the distribution of keyword sources in web queries at the first reformulation iteration and at subsequent reformulation iterations, respectively, across ChatGPT, Claude, and Grok.
DeepSeek is excluded from this analysis as its traces do not expose the actual queries issued to web search.
In the first iteration, both ChatGPT and Claude rely predominantly on \textit{Conversation History} as the source of new keywords, while \textit{Latest User Prompt} and \textit{Parametric Knowledge} contribute substantially smaller portions.
Grok shows a markedly different profile, with all three sources contributing comparable shares.
In subsequent reformulation iterations, \textit{Search Results} additionally appear as a possible keyword source, reflecting whether earlier retrievals can inform later queries.
On ChatGPT and Claude, \textit{Conversation History} continues to dominate even after reformulation, while \textit{Search Results} contribute only marginally, indicating that query reformulation on these platforms is primarily driven by the existing conversational context rather than by adapting to what has just been retrieved.
On Grok, in contrast, \textit{Search Results} emerge as a meaningful source alongside \textit{Conversation History} and \textit{Parametric Knowledge}, suggesting that Grok more actively incorporates search result content when reformulating its queries.

\subsection{Query Specificity}
\label{app:query_specificity}

\begin{figure*}[ht]
    \centering
    \begin{subfigure}[b]{0.32\linewidth}
    \includegraphics[width=1\linewidth]{figures/query_reformulation/query_specificity_distribution_by_iteration_dimension_specificity_direction__openai.pdf}
    \caption{ChatGPT}
    \end{subfigure}
    \centering
    \begin{subfigure}[b]{0.32\linewidth}
    \includegraphics[width=1\linewidth]{figures/other_platforms/query_reformulation/query_specificity_distribution_by_iteration_dimension_specificity_direction__claude.pdf}
    \caption{Claude}
    \end{subfigure}
    \centering
    \begin{subfigure}[b]{0.32\linewidth}
    \includegraphics[width=1\linewidth]{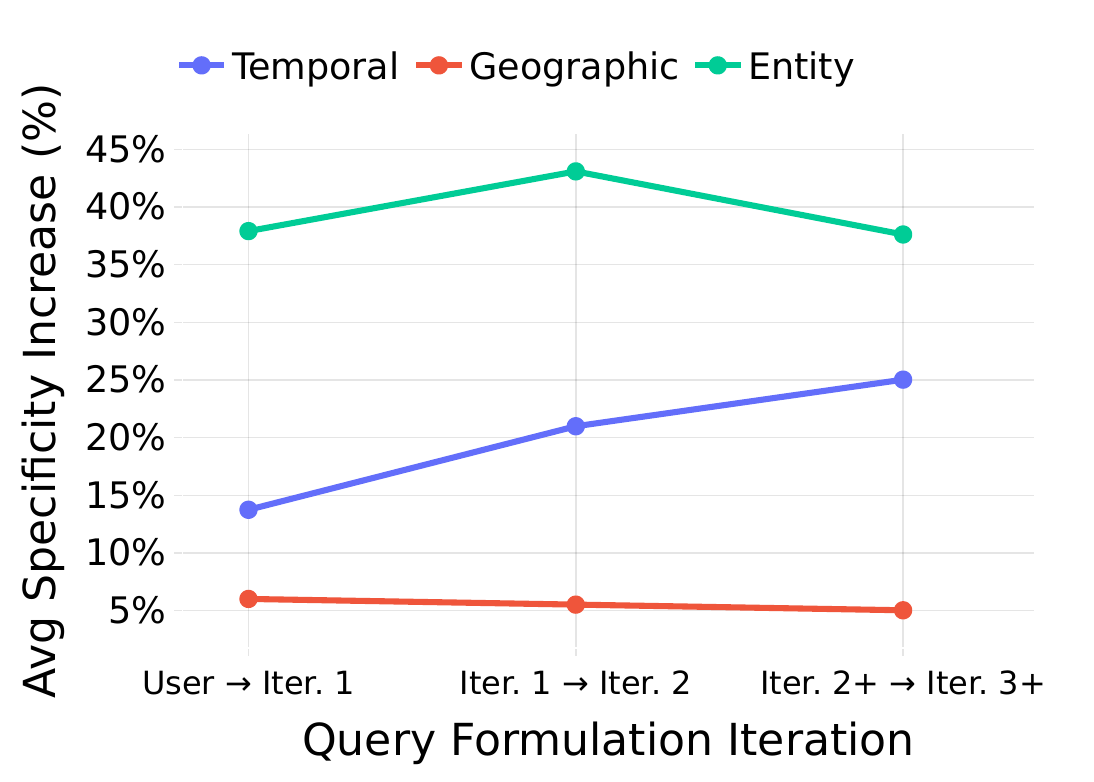}
    \caption{Grok}
    \end{subfigure}
    \caption{Average increase in temporal, geographic, and entity specificity across successive query reformulations in the \invivo{} setting. Higher values indicate that reformulated queries become more constrained along the corresponding dimension..}  \label{fig:query_specificity_distribution_by_iteration_grok}
\end{figure*}

\begin{figure*}[ht]
    \centering
    \begin{subfigure}[b]{0.48\linewidth}
    \includegraphics[width=1\linewidth]{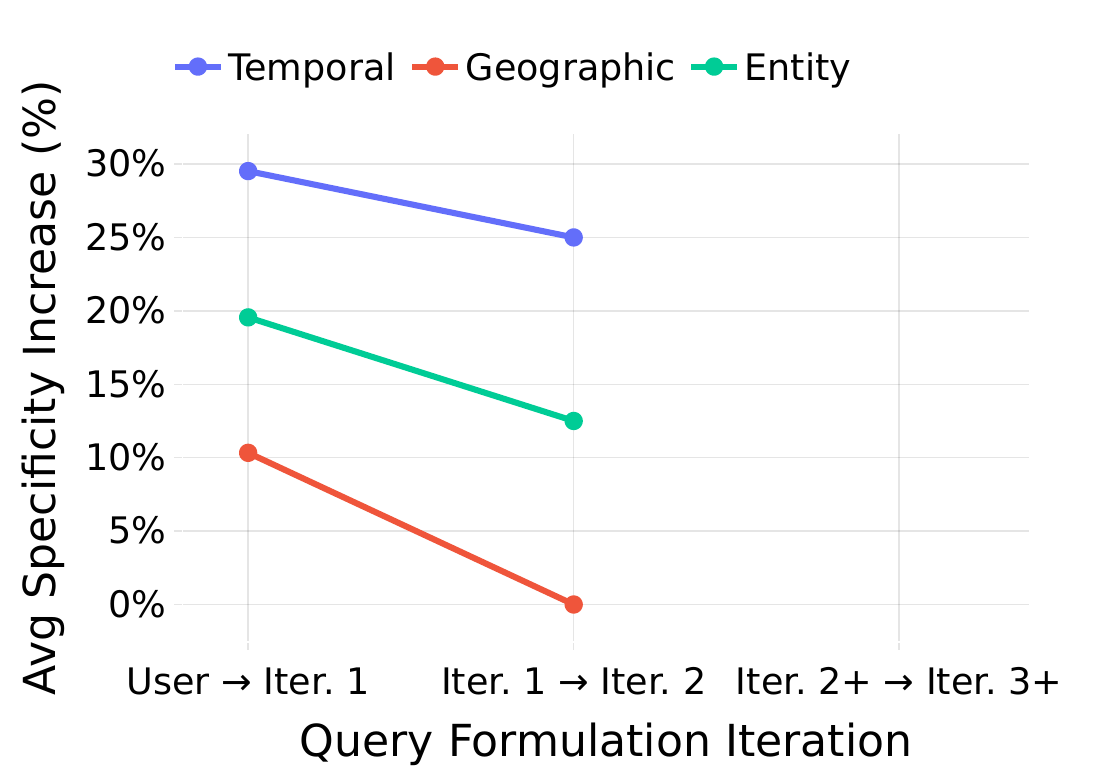}
    \caption{ChatGPT}
    \end{subfigure}
    \centering
    \begin{subfigure}[b]{0.48\linewidth}
    \includegraphics[width=1\linewidth]{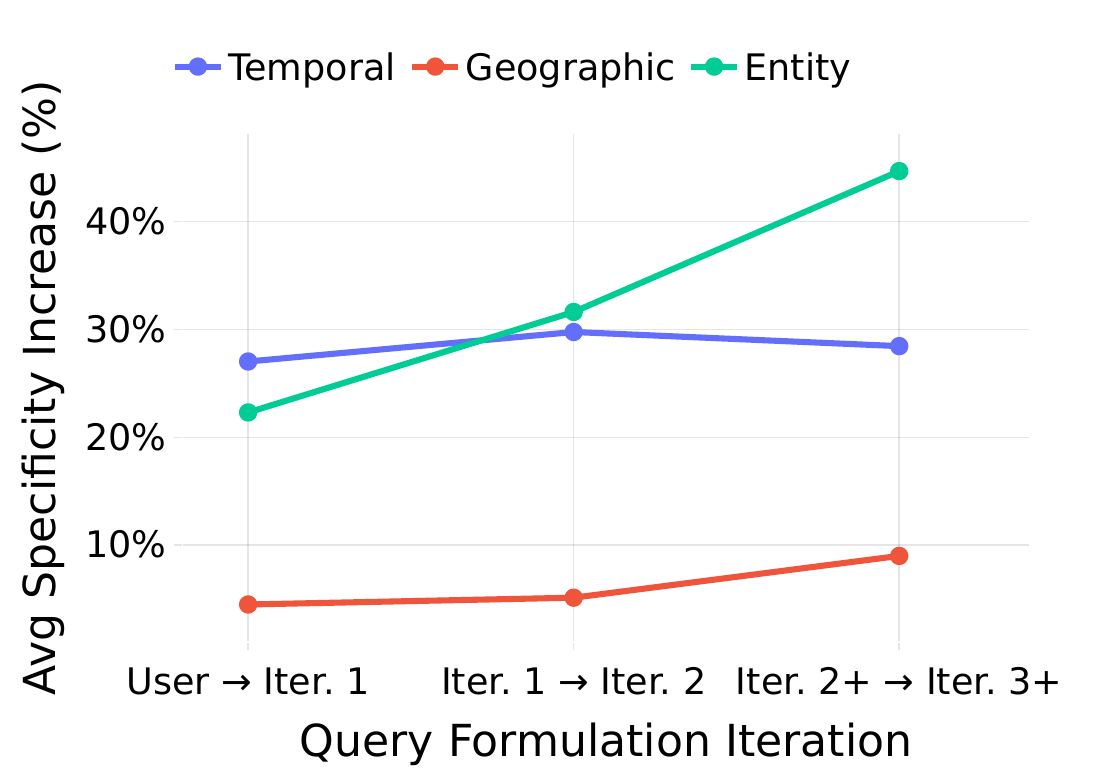}
    \caption{Claude}
    \end{subfigure}

    \centering
    \begin{subfigure}[b]{0.48\linewidth}
    \includegraphics[width=1\linewidth]{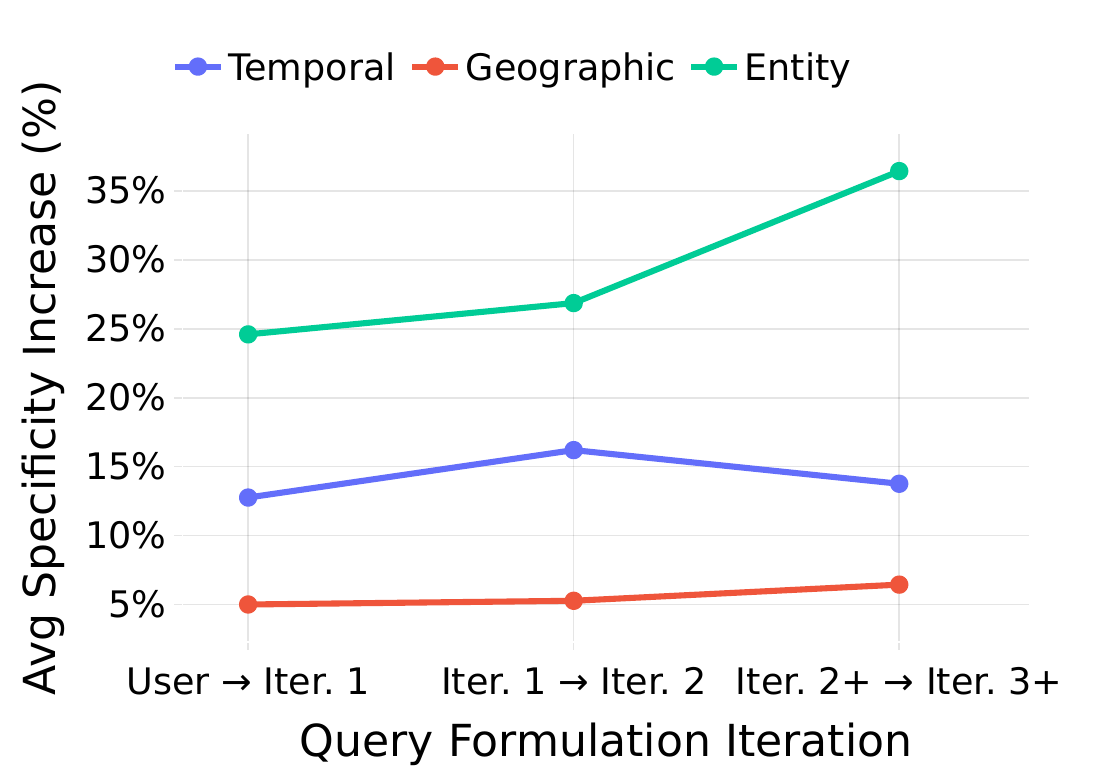}
    \caption{Grok}
    \end{subfigure}
    \centering
    \begin{subfigure}[b]{0.48\linewidth}
    \includegraphics[width=1\linewidth]{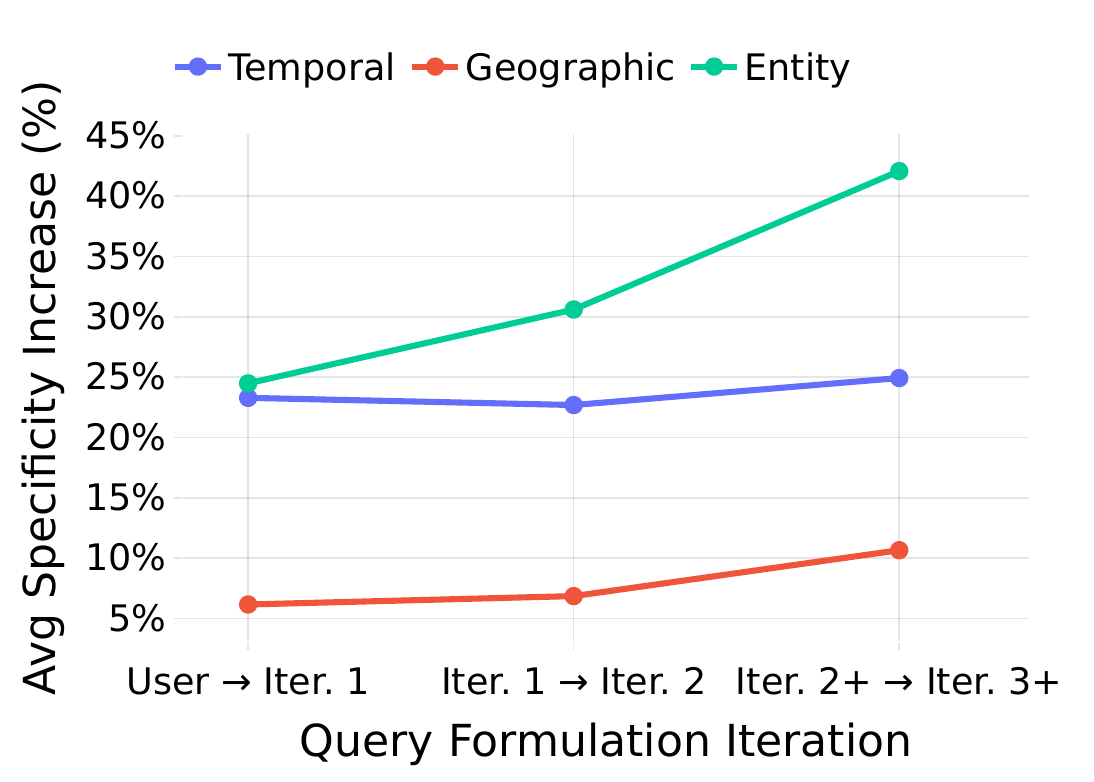}
    \caption{DeepSeek}
    \end{subfigure}
    \caption{Query Specificity analysis across dimensions \invitro.}  \label{fig:query_specificity_distribution_by_iteration_insitu}
\end{figure*}

To quantify query specificity, we compare consecutive query formulations along three dimensions: temporal, geographic, and entity specificity. For each dimension independently, the judge model scores each query on a 5-point Likert scale and an increase in these scores shows that the later query is more specific than the earlier one. If the reformulated query introduces additional constraints that make the search more precise (e.g., adding an exact date, a more specific location, or a named entity), we assign a score of +1 for that dimension. If specificity remains unchanged or decreases, we assign a score of 0. We then average these binary scores across all query pairs to obtain the average specificity increase reported in Figures~\ref{fig:query_specificity_distribution_by_iteration_grok} and \ref{fig:query_specificity_distribution_by_iteration_insitu}. Thus, a value of 40\% indicates that specificity increased in that dimension for approximately 40\% of query reformulations.

\subsection{Validity and Hallucination of Cited URLs}

\input{valid_url_table_invivo}

\input{valid_url_table_insitu}

Table~\ref{tab:url_status_invivo_platform} reports the validity breakdown of cited URLs across the four platforms, separated into search result (R) and parametric (P) citations.
Consistent with the observation in the main text, hallucinated URLs account for only a small fraction of search result citations on every platform -- 2.00\% on ChatGPT, 1.11\% on Claude, 0.26\% on Grok, and 1.73\% on DeepSeek -- indicating that hallucinated URLs are rare regardless of platform.
A similar pattern holds for parametric citations on ChatGPT, where the hallucination rate remains low (0.97\%). Claude's parametric citations show a higher hallucination rate (3.09\%), though this estimate is based on a small sample (8 of 259 parametric citations) and should be interpreted with caution. Overall, these results suggest that the absence of retrieval grounding does not substantially increase the likelihood of URL hallucination for ChatGPT, while the pattern for Claude is less conclusive given the limited sample size.

\subsection{Source Preferences}

\begin{figure*}[ht]
    \centering
    \begin{subfigure}[b]{0.49\columnwidth}
    \includegraphics[width=1\linewidth]{figures/source_selection/source_rank_violinplot_split_cited_conversation_grounding.pdf}
    \caption{ChatGPT}
    \label{fig:source_rank_violinplot_tranco_openai}
    \end{subfigure}
    \centering
    \begin{subfigure}[b]{0.49\columnwidth}
    \includegraphics[width=1\linewidth]{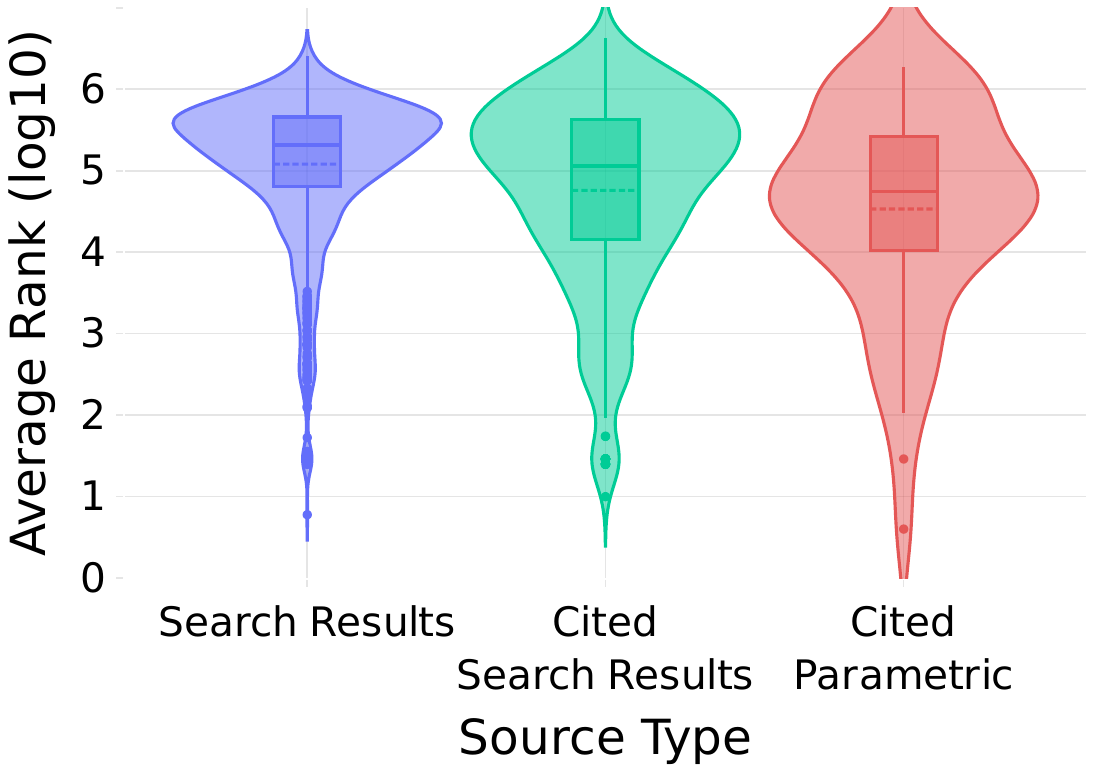}
    \caption{Claude}
    \end{subfigure}
    \centering
    \begin{subfigure}[b]{0.49\columnwidth}
    \includegraphics[width=1\linewidth]{figures/other_platforms/source_selection/source_rank_violinplot_split_cited_conversation_grounding_grok.pdf}
    \caption{Grok}
    \end{subfigure}
    \centering
    \begin{subfigure}[b]{0.49\columnwidth}
    \includegraphics[width=1\linewidth]{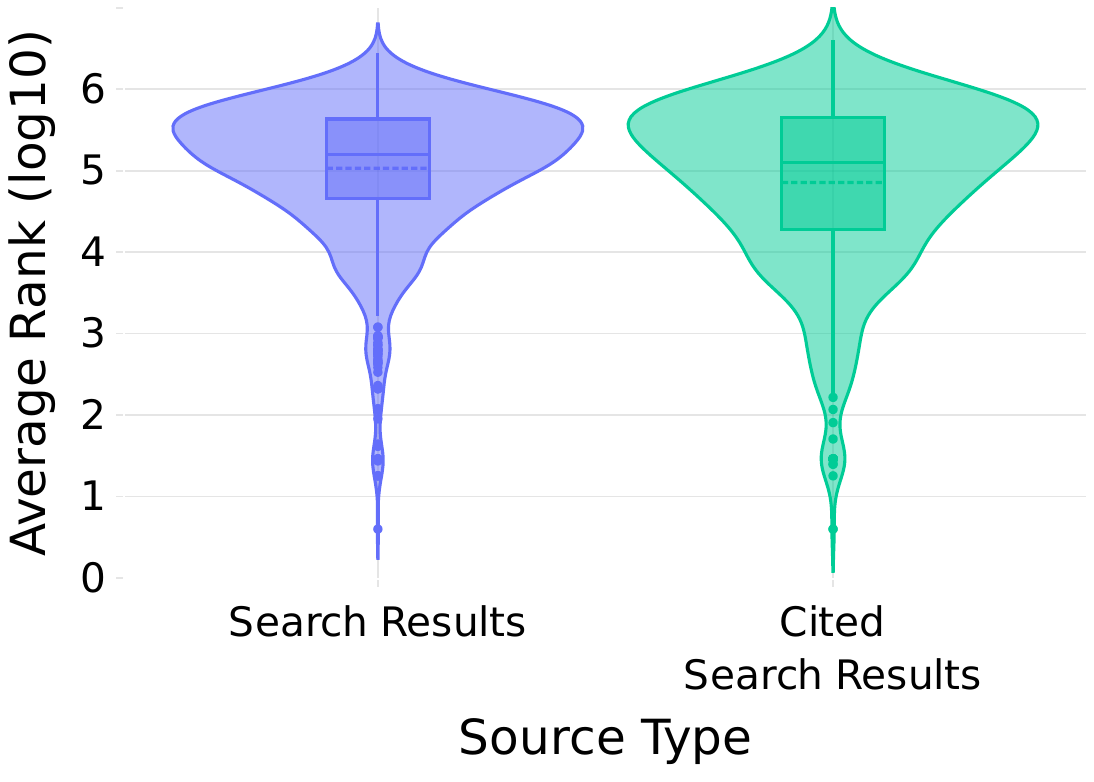}
    \caption{DeepSeek}
    \end{subfigure}
    \caption{Tranco ranks of domains of search results and cited URLs across platforms.}  \label{fig:source_rank_violinplot_tranco_other_platforms}
\end{figure*}

\begin{figure*}[ht]
\centering
    \begin{subfigure}{0.48\linewidth}
    \centering
    \includegraphics[width=1\linewidth]{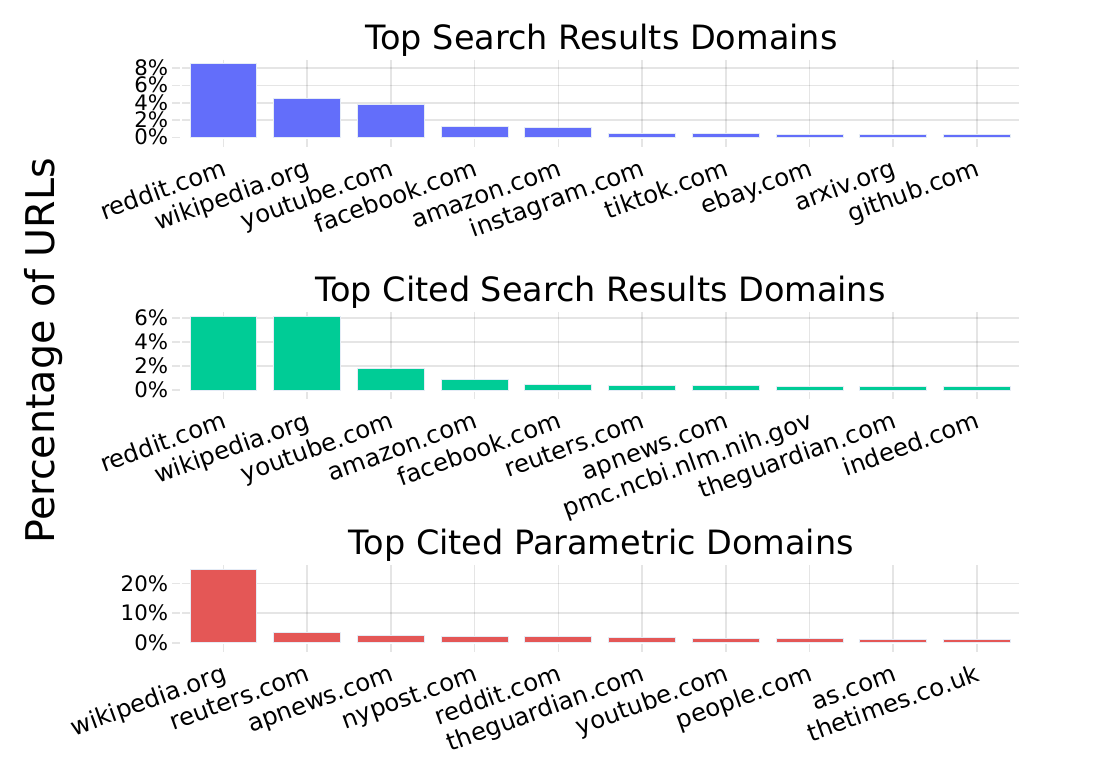}
    \caption{ChatGPT}
    \label{fig:top_domains_openai}
    \end{subfigure}
    \begin{subfigure}{0.48\linewidth}
    \centering
    \includegraphics[width=1\linewidth]{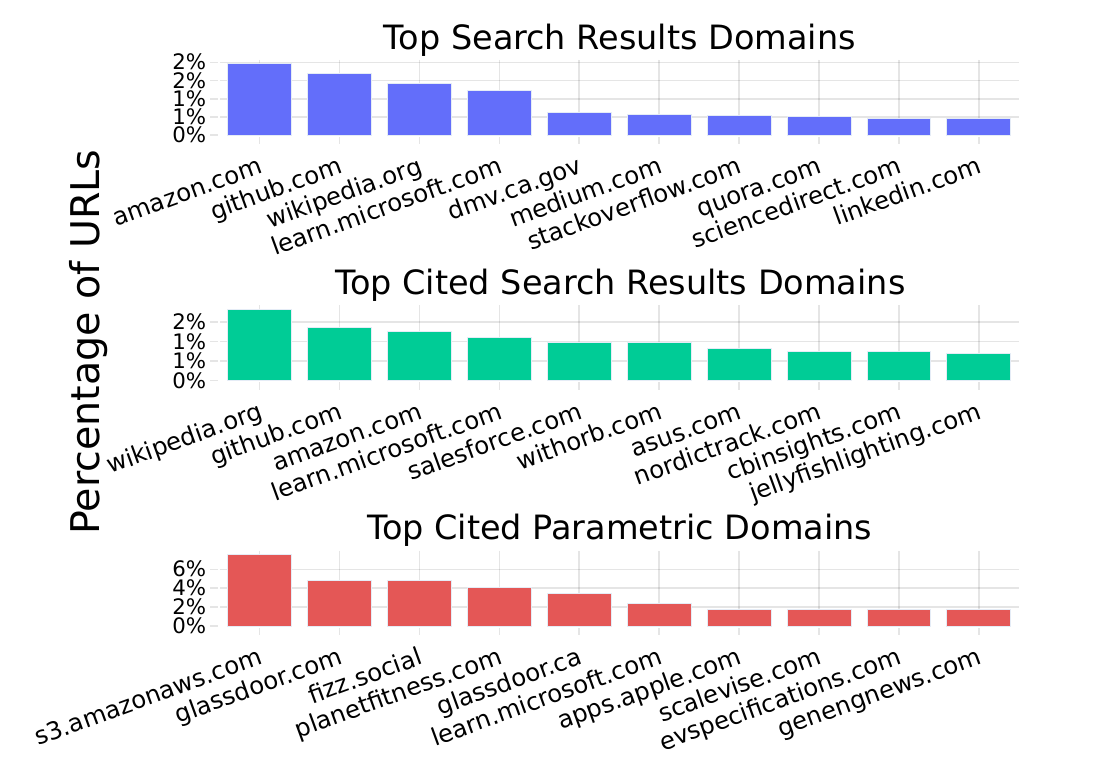}
    \caption{Claude}
    \label{fig:top_domains_claude}
    \end{subfigure}
    
    \centering
    \begin{subfigure}{0.48\linewidth}
    \centering
    \includegraphics[width=1\linewidth]{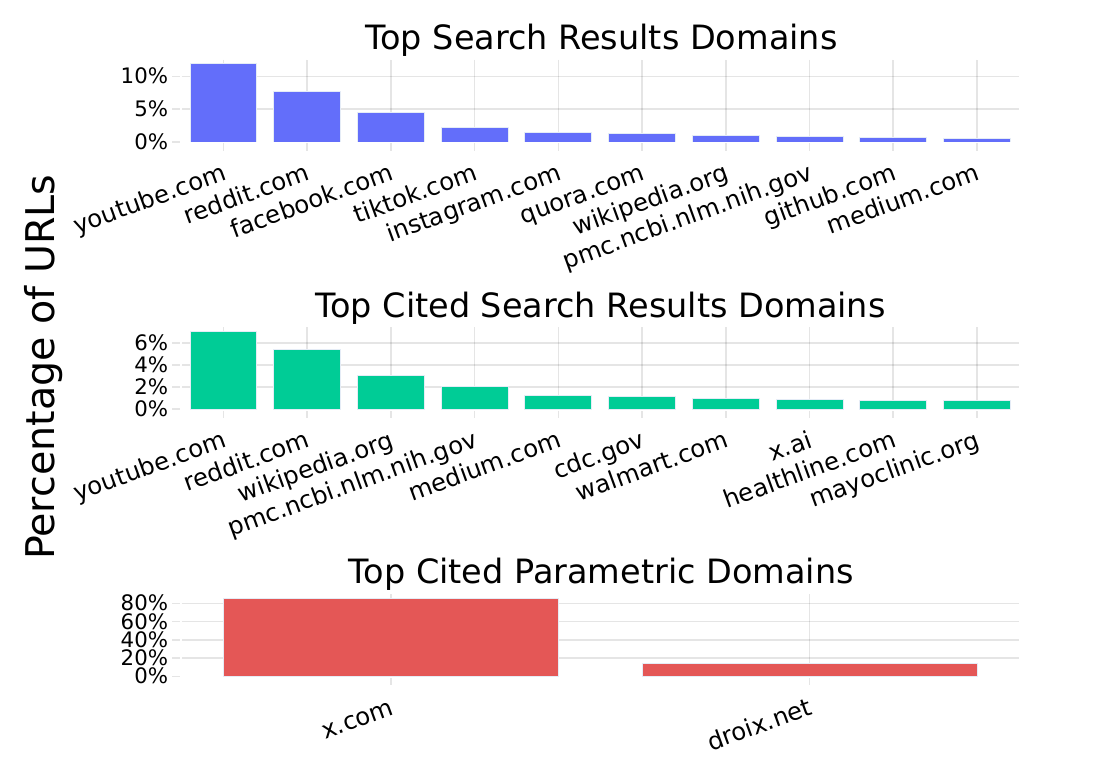}
    \caption{Grok}
    \label{fig:top_domains_grok}
    \end{subfigure}
    \centering
    \begin{subfigure}{0.48\linewidth}
    \centering
    \includegraphics[width=1\linewidth]{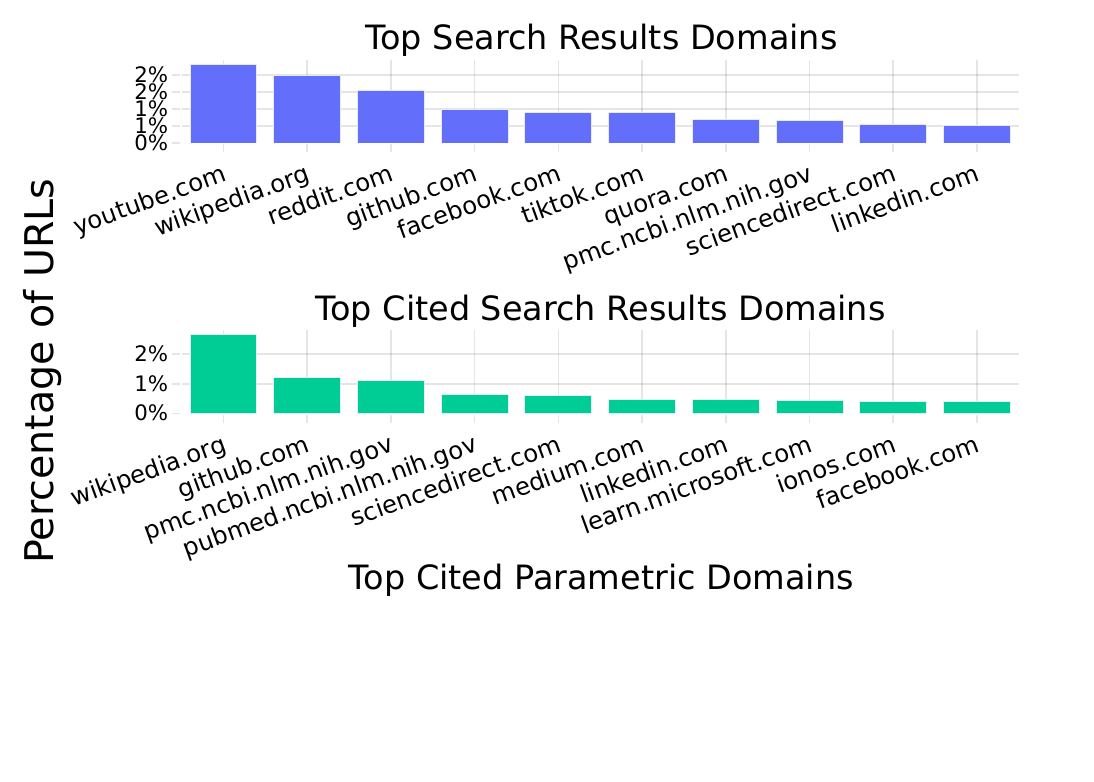}
    \caption{DeepSeek}
    \label{fig:top_domains_deepseek}
    \end{subfigure}
    
\caption{Top 10 domains across platforms.}
\label{fig:top_domains_other_platforms}
\end{figure*}

Figure~\ref{fig:source_rank_violinplot_tranco_other_platforms} shows the Tranco rank distributions of search results, cited search results, and cited parametric URLs across platforms.
Consistent with the pattern observed on ChatGPT, Grok's cited parametric URLs tend to have higher-ranked domains (lower Tranco rank values) than search result citations, while Claude's cited parametric URLs show a rank distribution similar to its search result citations.
Figure~\ref{fig:top_domains_other_platforms} shows the top 20 domains on each platform across three categories: \textit{Top Search Result Domains}, \textit{Top Cited Search Result Domains}, and \textit{Top Cited Parametric Domains}.
Across all four platforms, the search result domain distributions are dominated by a small number of high-traffic reference sites such as wikipedia.org, reddit.com, and youtube.com.
Beyond this commonality, however, the composition of top search result domains reveals distinct platform characteristics: ChatGPT, Grok, and Deepseek are dominated by general consumer sites such as reddit.com, youtube.com, and facebook.com, whereas Claude shows a different profile in which coding- and programming-related sites, such as github.com, learn.microsoft.com, and stackoverflow.com, occupy the top positions.
The \textit{Top Cited Search Result Domains} largely mirror the search result distributions on each platform, indicating that citations are drawn primarily from the search result URLs.
The \textit{Top Cited Parametric Domains}, which capture domains referenced from parametric knowledge without retrieval, exhibit the most distinct cross-platform differences.
ChatGPT is overwhelmingly dominated by wikipedia.org, while Claude shows a comparatively flat distribution spanning diverse domains such as salesforce.com, withorb.com, and asus.com.
Grok produces only 14 parametric citations out of 3{,}916 in total, all of which are concentrated on x.com and droix.net.
For DeepSeek, no parametric citations are observed, as its trace format structurally attaches citations only to search result URLs, rendering parametric citations unobservable by construction. 
Overall, although the search result domains differ across platforms, conversational agents consistently prioritize a relatively small set of authoritative domains when generating citations.

%% file: valid_url_table_invivo.tex
\begin{table}[ht]
\centering
\small
\resizebox{\columnwidth}{!}{
\begin{tabular}{ccccccc}
\toprule[1.3pt]
Platform & Type & Valid & Unk. & Hall. & Dead & Total \\
\midrule

\multirow{4}{*}{ChatGPT}
& \multirow{2}{*}{R}
& 137,217 & 16,607 & 3,190 & 2,566 & 159,580 \\
& & \scriptsize(85.99\%) & \scriptsize(10.41\%) & \scriptsize(2.00\%) & \scriptsize(1.61\%) & \\
\cmidrule(lr){3-7}
& \multirow{2}{*}{P}
& 7,930 & 845 & 87 & 68 & 8,930 \\
& & \scriptsize(88.80\%) & \scriptsize(9.46\%) & \scriptsize(0.97\%) & \scriptsize(0.76\%) & \\

\midrule

\multirow{4}{*}{Claude}
& \multirow{2}{*}{R}
& 4,781 & 491 & 60 & 77 & 5,409 \\
& & \scriptsize(88.39\%) & \scriptsize(9.08\%) & \scriptsize(1.11\%) & \scriptsize(1.42\%) & \\
\cmidrule(lr){3-7}
& \multirow{2}{*}{P}
& 205 & 29 & 8 & 17 & 259 \\
& & \scriptsize(79.15\%) & \scriptsize(11.20\%) & \scriptsize(3.09\%) & \scriptsize(6.56\%) & \\

\midrule

\multirow{4}{*}{Grok}
& \multirow{2}{*}{R}
& 1,784 & 149 & 5 & 5 & 1,943 \\
& & \scriptsize(91.82\%) & \scriptsize(7.67\%) & \scriptsize(0.26\%) & \scriptsize(0.26\%) & \\
\cmidrule(lr){3-7}
& \multirow{2}{*}{P}
& 4 & 3 & 0 & 0 & 7 \\
& & \scriptsize(57.14\%) & \scriptsize(42.86\%) & \scriptsize(0.00\%) & \scriptsize(0.00\%) & \\

\midrule

\multirow{4}{*}{DeepSeek}
& \multirow{2}{*}{R}
& 6,710 & 694 & 132 & 74 & 7,610 \\
& & \scriptsize(88.17\%) & \scriptsize(9.12\%) & \scriptsize(1.73\%) & \scriptsize(0.97\%) & \\
\cmidrule(lr){3-7}
& \multirow{1}{*}{P}
& 0 & 0 & 0 & 0 & 0 \\

\bottomrule[1.3pt]
\end{tabular}
}
\caption{\textbf{\invivo{} cited URL validity breakdown} for different platforms by citation type.
R denotes retrieved citations, while P denotes citations attributed to parametric knowledge.
Valid URLs are reachable, Unk. indicates unverifiable URLs, Hall. indicates hallucinated URLs, and Dead indicates stale URLs.}
\label{tab:url_status_invivo_platform}
\end{table}

%% file: valid_url_table_insitu.tex
\begin{table}[ht]
\centering
\small
\resizebox{\columnwidth}{!}{%
\begin{tabular}{ccccccc}
\toprule[1.3pt]
Platform & Type & Valid & Unk. & Hall. & Dead & Total \\
\midrule

\multirow{4}{*}{\shortstack{ChatGPT\\(All)}}
& \multirow{2}{*}{R}
& 460 & 62 & 1 & 3 & 526 \\
& 
& \scriptsize(87.45\%)
& \scriptsize(11.79\%)
& \scriptsize(0.19\%)
& \scriptsize(0.57\%)
& \\
\cmidrule(lr){3-7}
& \multirow{2}{*}{P}
& 8 & 0 & 0 & 0 & 8 \\
&
& \scriptsize(100.00\%)
& \scriptsize(0.00\%)
& \scriptsize(0.00\%)
& \scriptsize(0.00\%)
& \\

\midrule

\multirow{4}{*}{\shortstack{Claude\\(All)}}
& \multirow{2}{*}{R}
& 6,064 & 538 & 121 & 67 & 6,790 \\
&
& \scriptsize(89.31\%)
& \scriptsize(7.92\%)
& \scriptsize(1.78\%)
& \scriptsize(0.99\%)
& \\
\cmidrule(lr){3-7}
& \multirow{2}{*}{P}
& 0 & 0 & 0 & 0 & 0 \\
&
& \scriptsize(0.00\%)
& \scriptsize(0.00\%)
& \scriptsize(0.00\%)
& \scriptsize(0.00\%)
& \\

\midrule

\multirow{4}{*}{\shortstack{Grok\\(All)}}
& \multirow{2}{*}{R}
& 4,037 & 304 & 5 & 18 & 4,364 \\
&
& \scriptsize(92.51\%)
& \scriptsize(6.97\%)
& \scriptsize(0.11\%)
& \scriptsize(0.41\%)
& \\
\cmidrule(lr){3-7}
& \multirow{2}{*}{P}
& 37 & 3 & 1 & 1 & 42 \\
&
& \scriptsize(88.10\%)
& \scriptsize(7.14\%)
& \scriptsize(2.38\%)
& \scriptsize(2.38\%)
& \\



\midrule

\multirow{4}{*}{\shortstack{ChatGPT\\(Common)}}
& \multirow{2}{*}{R}
& 435 & 62 & 1 & 3 & 501 \\
&
& \scriptsize(86.83\%)
& \scriptsize(12.38\%)
& \scriptsize(0.20\%)
& \scriptsize(0.60\%)
& \\
\cmidrule(lr){3-7}
& \multirow{2}{*}{P}
& 8 & 0 & 0 & 0 & 8 \\
&
& \scriptsize(100.00\%)
& \scriptsize(0.00\%)
& \scriptsize(0.00\%)
& \scriptsize(0.00\%)
& \\

\midrule

\multirow{4}{*}{\shortstack{Claude\\(Common)}}
& \multirow{2}{*}{R}
& 870 & 108 & 22 & 9 & 1,009 \\
&
& \scriptsize(86.22\%)
& \scriptsize(10.70\%)
& \scriptsize(2.18\%)
& \scriptsize(0.89\%)
& \\
\cmidrule(lr){3-7}
& \multirow{2}{*}{P}
& 0 & 0 & 0 & 0 & 0 \\
&
& \scriptsize(0.00\%)
& \scriptsize(0.00\%)
& \scriptsize(0.00\%)
& \scriptsize(0.00\%)
& \\

\midrule

\multirow{4}{*}{\shortstack{Grok\\(Common)}}
& \multirow{2}{*}{R}
& 701 & 70 & 0 & 0 & 771 \\
&
& \scriptsize(90.92\%)
& \scriptsize(9.08\%)
& \scriptsize(0.00\%)
& \scriptsize(0.00\%)
& \\
\cmidrule(lr){3-7}
& \multirow{2}{*}{P}
& 21 & 2 & 0 & 1 & 24 \\
&
& \scriptsize(87.50\%)
& \scriptsize(8.33\%)
& \scriptsize(0.00\%)
& \scriptsize(4.17\%)
& \\


\bottomrule[1.3pt]
\end{tabular}%
}

\caption{
\textbf{\invitro{} cited URL validity breakdown} for different platforms
under the \textit{All} and \textit{Common} settings by citation type.
R denotes retrieved citations, while P denotes citations attributed to
parametric knowledge. Valid URLs are reachable, Unk. indicates
unverifiable URLs, Hall. indicates hallucinated URLs, and Dead indicates
stale or inaccessible URLs.
}
\label{tab:url_status_insitu_platform}
\end{table}

%% file: replay_platforms.tex
\section{Additional \invitro{} Results}
\label{sec:app-replay-platforms}

\subsection{Querying Strategies}
\label{app:query_strategy_invitro}

\begin{figure}[ht]
    \centering
    \begin{subfigure}[b]{0.48\columnwidth}
        \centering
        \includegraphics[width=\linewidth]{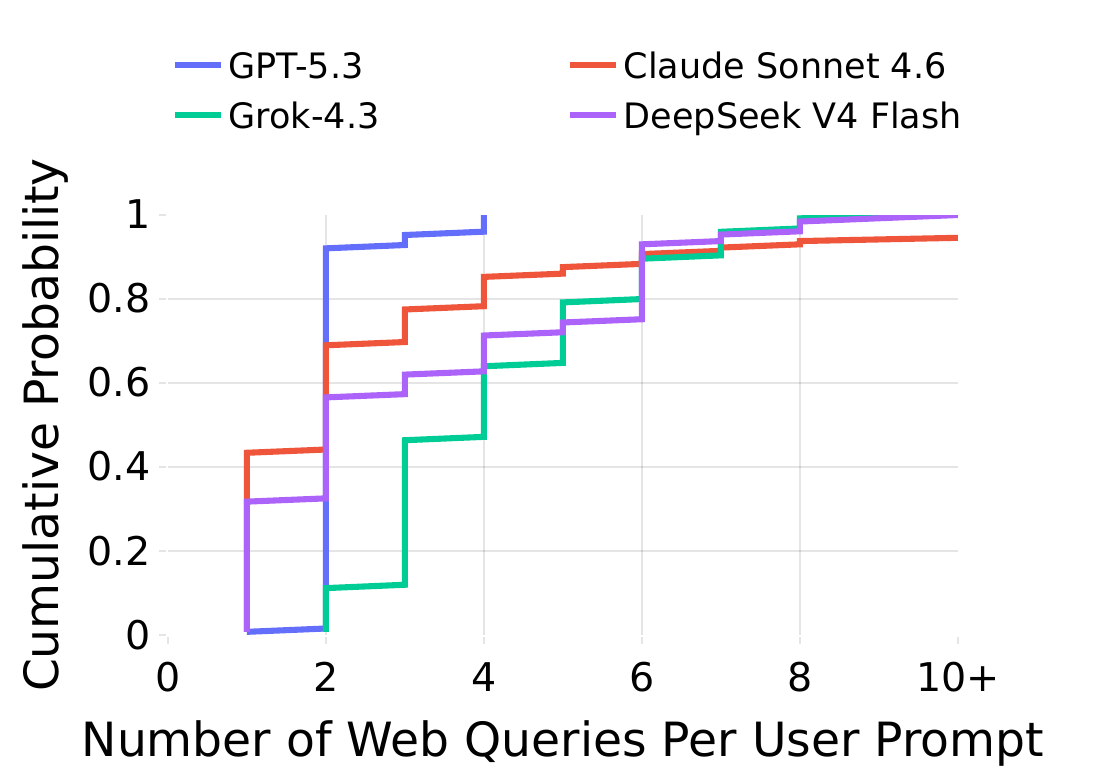}
        \caption{Web queries per prompt}
        \label{fig:total_Web_queries_insitu_rev}
    \end{subfigure} 
    \begin{subfigure}[b]{0.48\columnwidth}
        \centering
        \includegraphics[width=\linewidth]{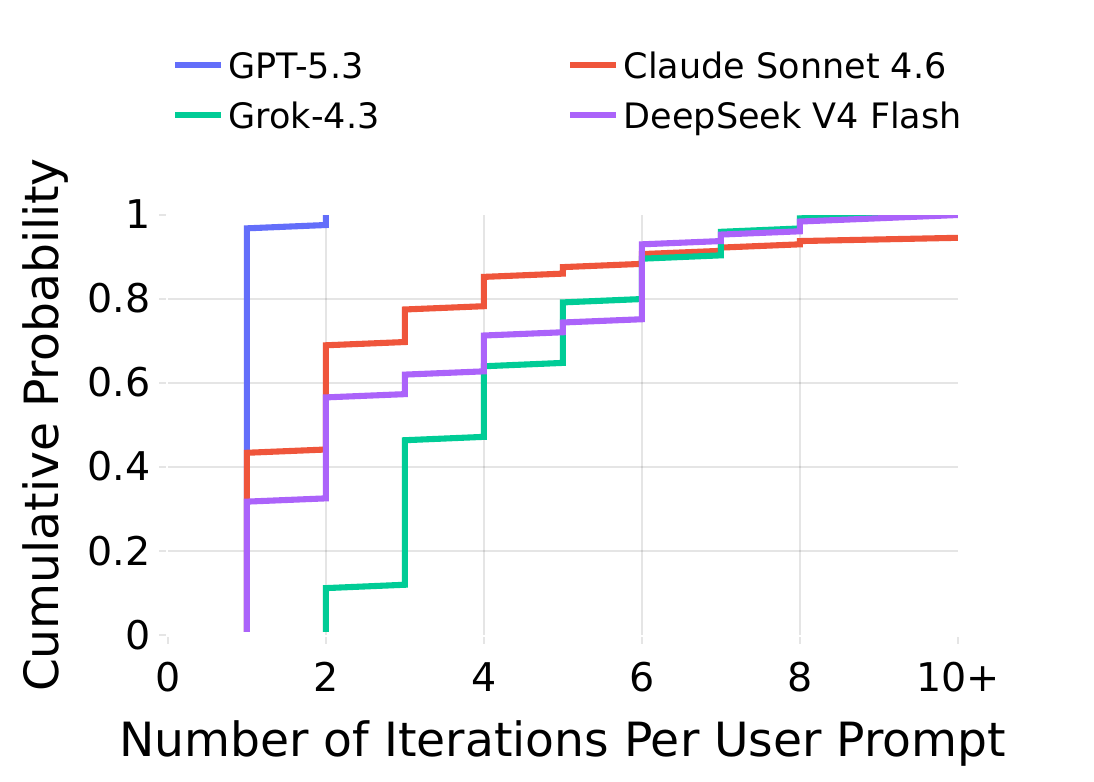}
        \caption{Iterations per prompt}
        \label{fig:iterations_per_prompt_insitu_rev}
    \end{subfigure}
    \caption{Query complexity increases in different models in \invitro{} over total number of web queries and iterations.}
    \label{fig:query_reformulation_cdfs_rev_invitro}
\end{figure}

Figure~\ref{fig:query_reformulation_cdfs_rev_invitro} shows that different models adopt different querying strategies.

\subsection{Search Result Domain Preferences}
\label{app:domain_preference_insitu}

\begin{figure*}[ht]
\centering
\begin{subfigure}{1\linewidth}
    \centering
    \includegraphics[width=0.5\linewidth]{figures/source_selection/top_retrieved_domains_overall.pdf}
    \caption{\invivo}
    \label{fig:top_retrieved_domains_invivo_app}
    \end{subfigure}
    \begin{subfigure}{1\linewidth}
    \centering
    \includegraphics[width=0.5\linewidth]{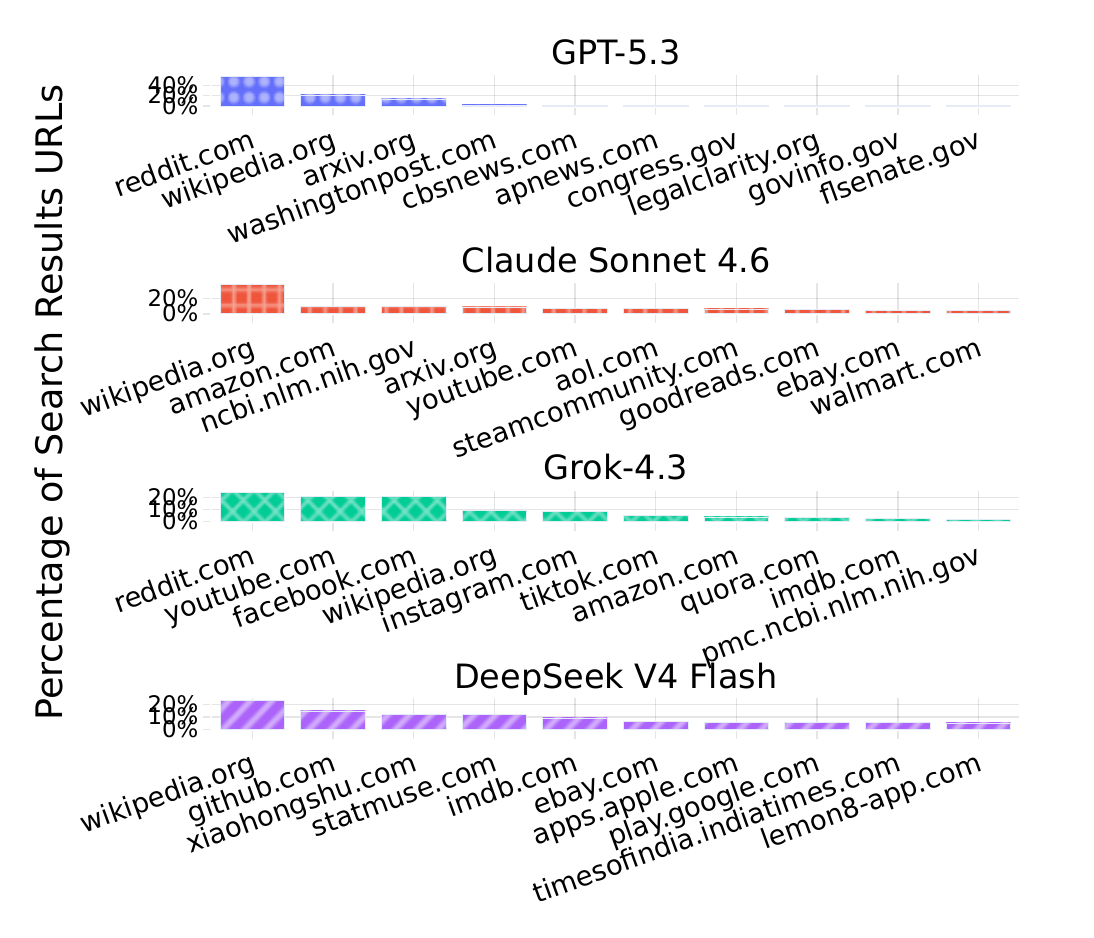}
    \caption{\invitro{} (all)}
    \label{fig:top_retrieved_domains_other_platforms_insitu_web}
    \end{subfigure}
    \begin{subfigure}{0.5\linewidth}
    \centering
    \includegraphics[width=1\linewidth]{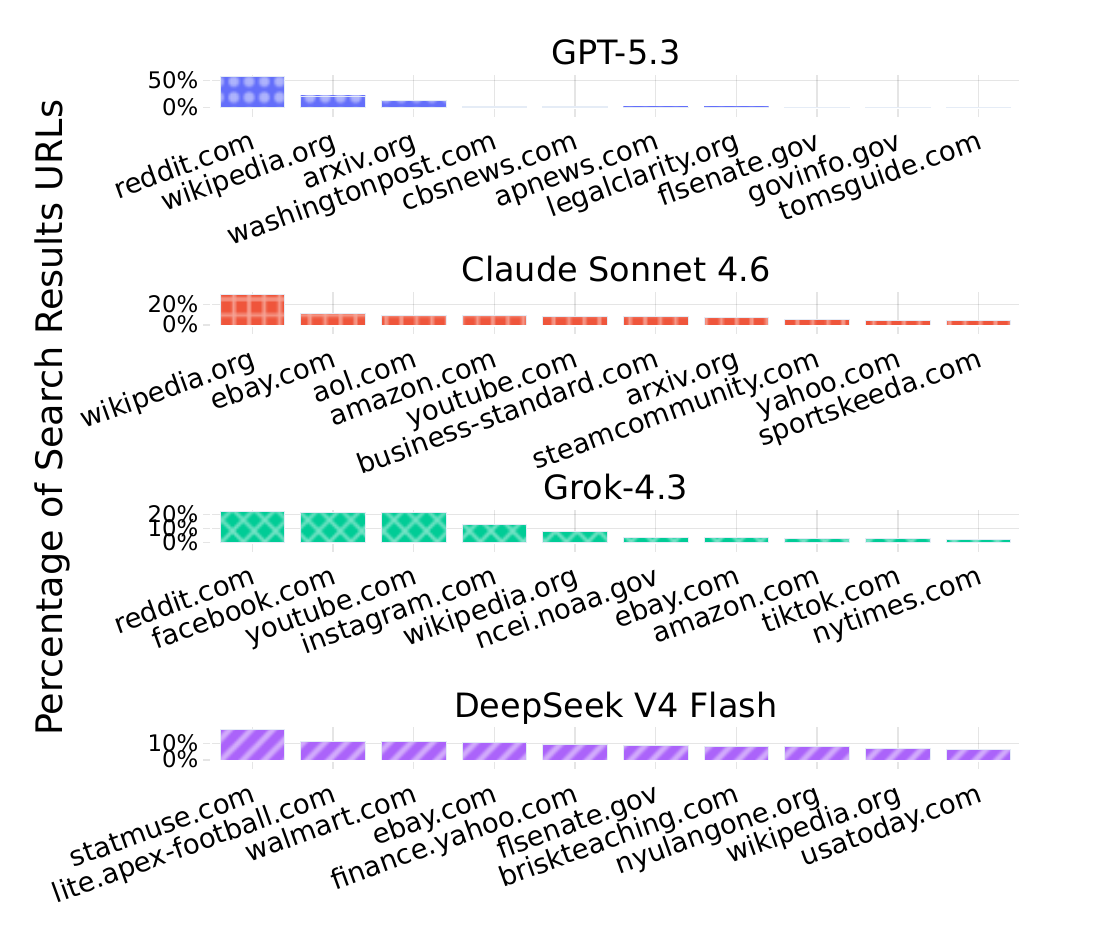}
    \caption{\invitro{} (common)}
    \label{fig:top_retrieved_domains_other_platforms_insitu_common}
    \end{subfigure}
    
\caption{Top 10 domains preferred by search engines across platforms, exhibiting substantial differences in domain preferences.}
\label{fig:top_retrieved_domains_other_platforms_invivo_insitu}
\end{figure*}

\begin{figure*}[ht]
\centering
    \begin{subfigure}{0.48\linewidth}
    \centering
    \includegraphics[width=1\linewidth]{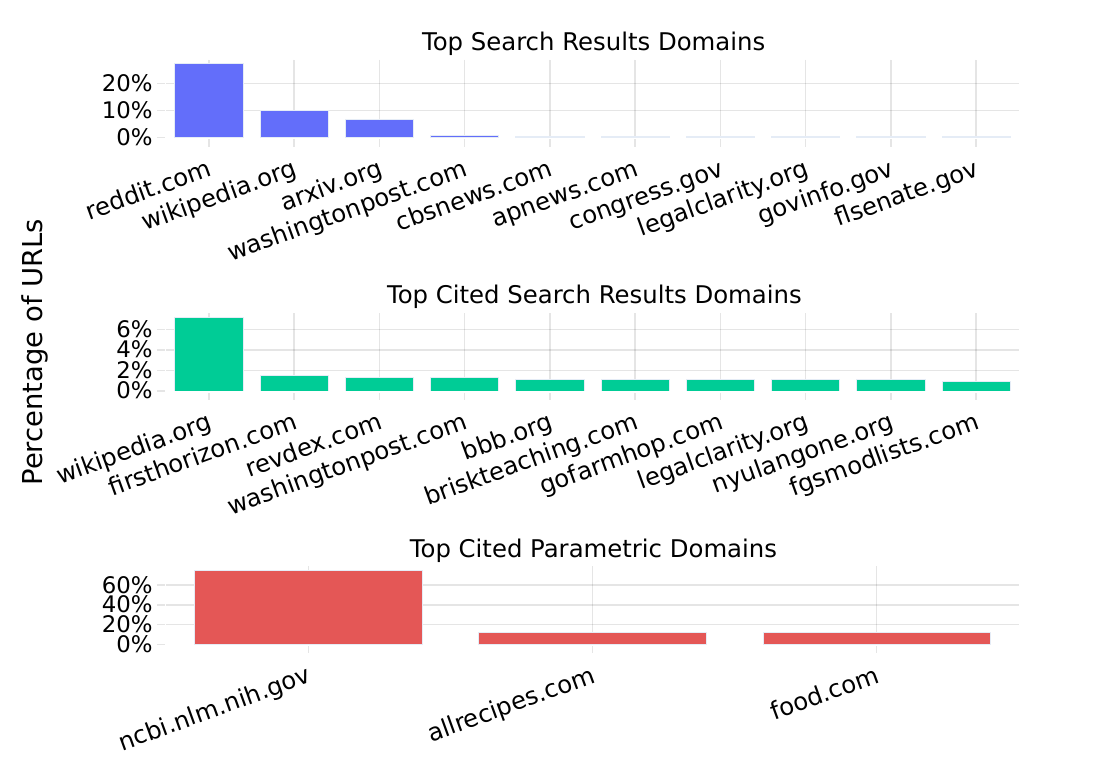}
    \caption{ChatGPT}
    \label{fig:top_domains_openai_insitu}
    \end{subfigure}
    \begin{subfigure}{0.48\linewidth}
    \centering
    \includegraphics[width=1\linewidth]{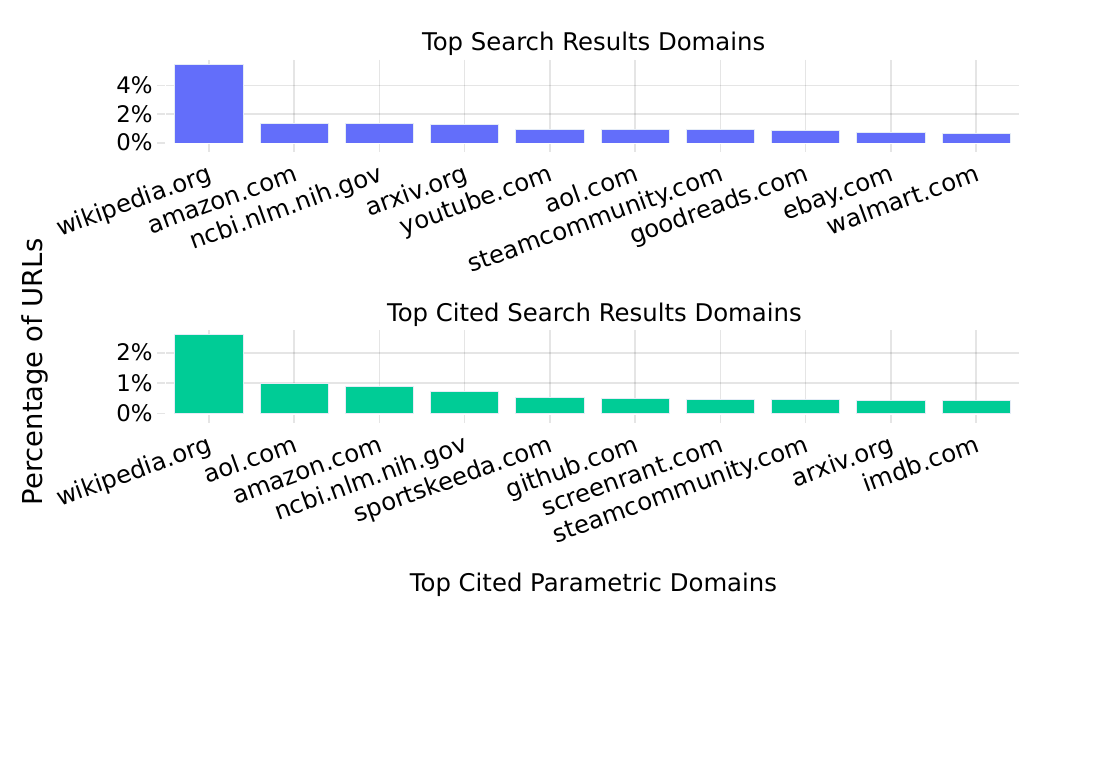}
    \caption{Claude}
    \label{fig:top_domains_claude_insitu}
    \end{subfigure}
    
    \centering
    \begin{subfigure}{0.48\linewidth}
    \centering
    \includegraphics[width=1\linewidth]{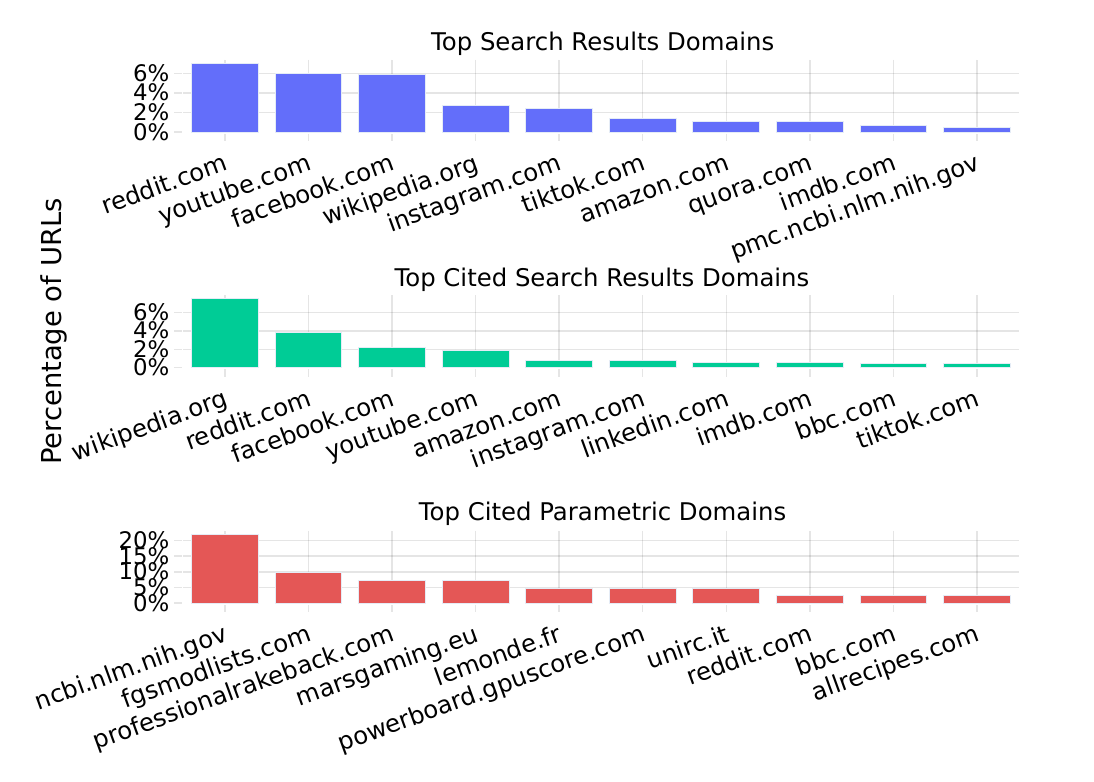}
    \caption{Grok}
    \label{fig:top_domains_grok_insitu}
    \end{subfigure}
    \centering
    \begin{subfigure}{0.48\linewidth}
    \centering
    \includegraphics[width=1\linewidth]{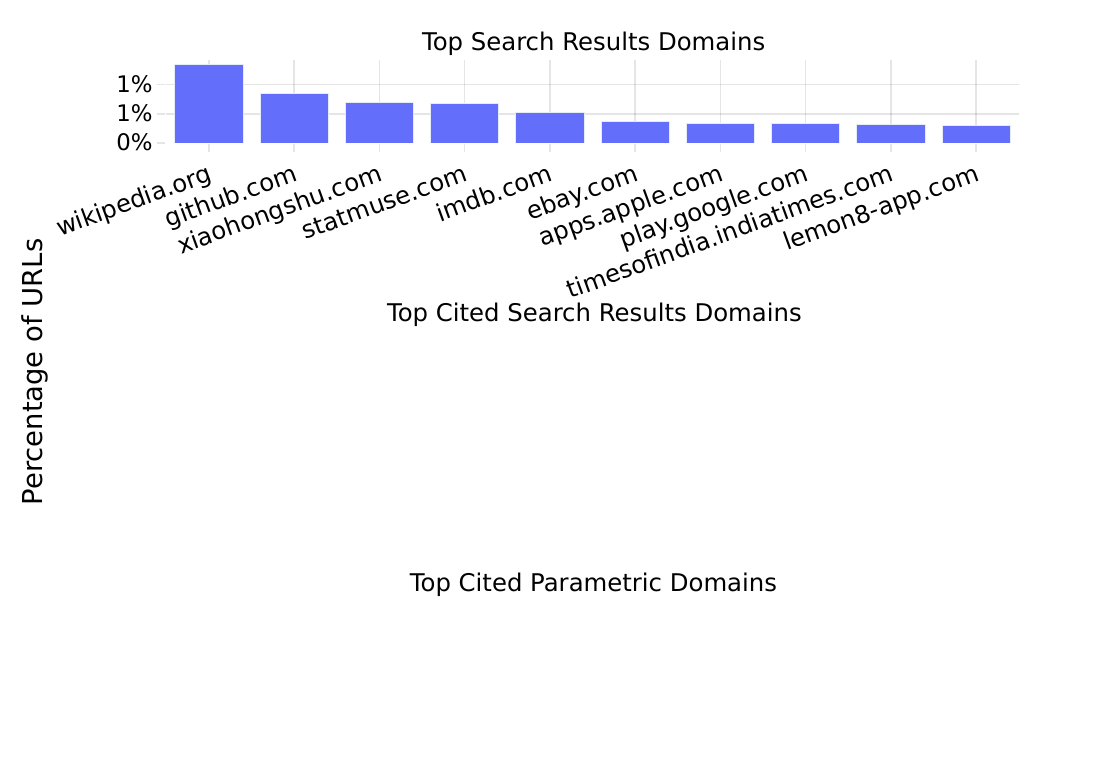}
    \caption{DeepSeek}
    \label{fig:top_domains_deepseek_insitu}
    \end{subfigure}
    
\caption{Top 10 domains across platforms \invitro.}
\label{fig:top_domains_other_platforms_insitu}
\end{figure*}

\begin{figure*}[ht]
\centering
    \begin{subfigure}{0.48\linewidth}
    \centering
    \includegraphics[width=1\linewidth]{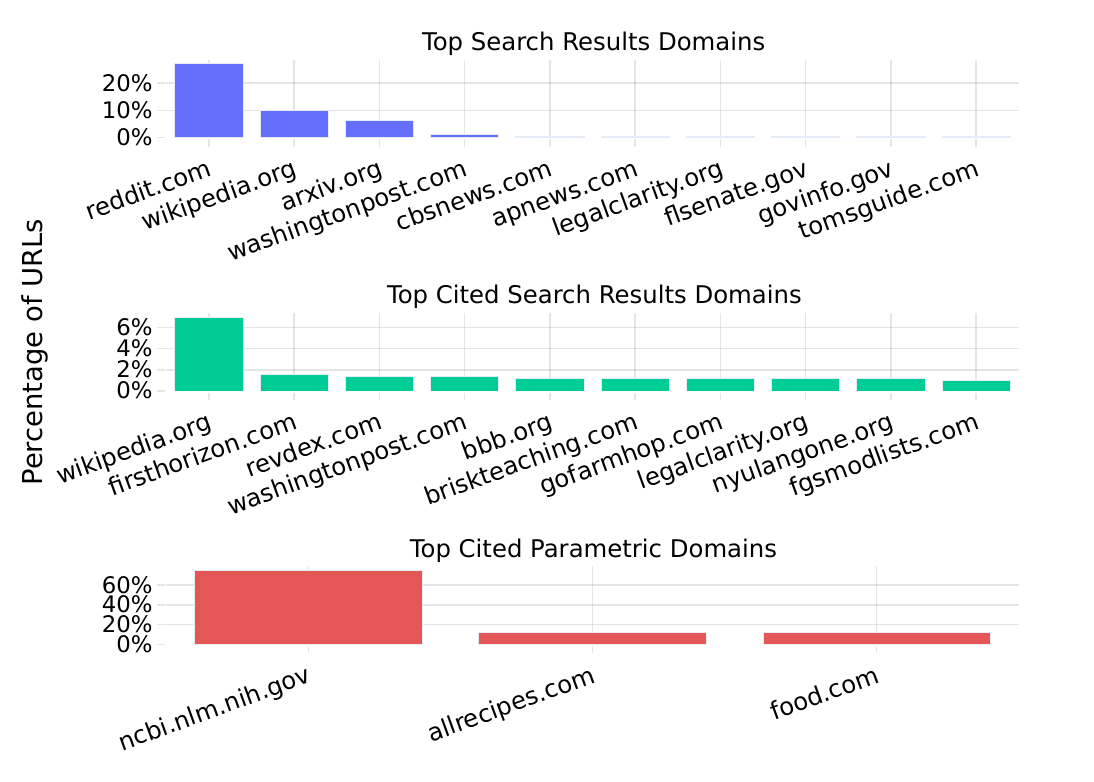}
    \caption{ChatGPT}
    \label{fig:top_domains_openai_insitu_common}
    \end{subfigure}
    \begin{subfigure}{0.48\linewidth}
    \centering
    \includegraphics[width=1\linewidth]{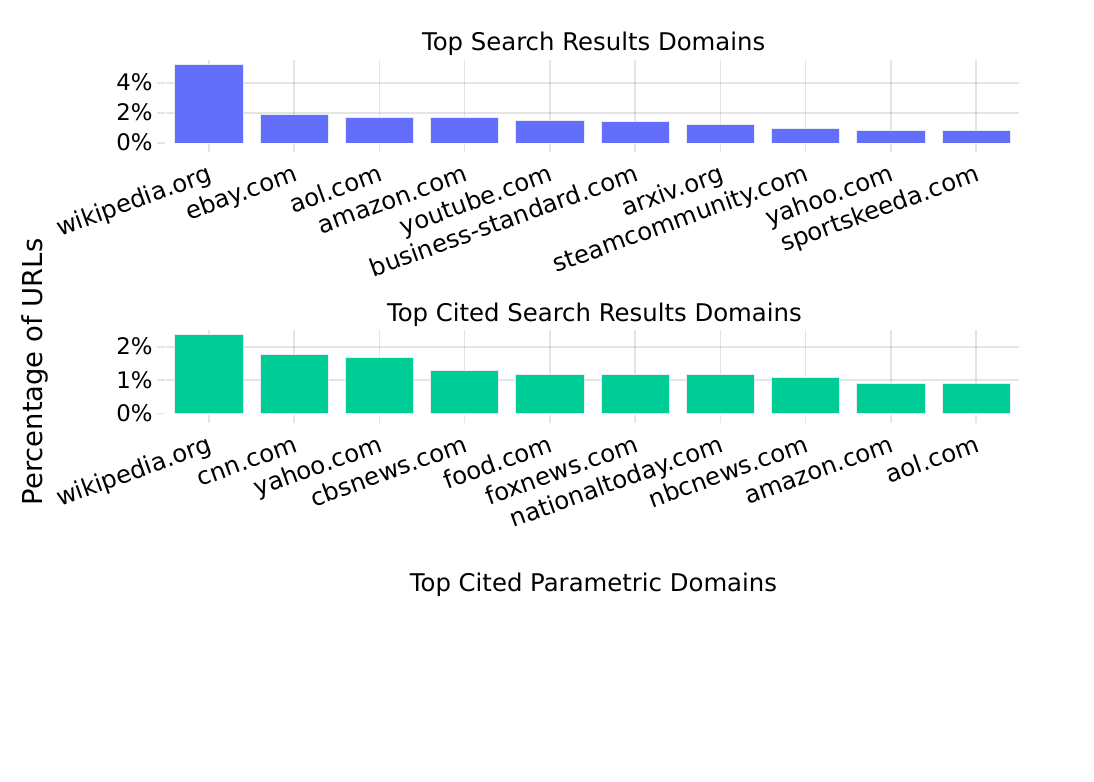}
    \caption{Claude}
    \label{fig:top_domains_claude_insitu_common}
    \end{subfigure}
    
    \centering
    \begin{subfigure}{0.48\linewidth}
    \centering
    \includegraphics[width=1\linewidth]{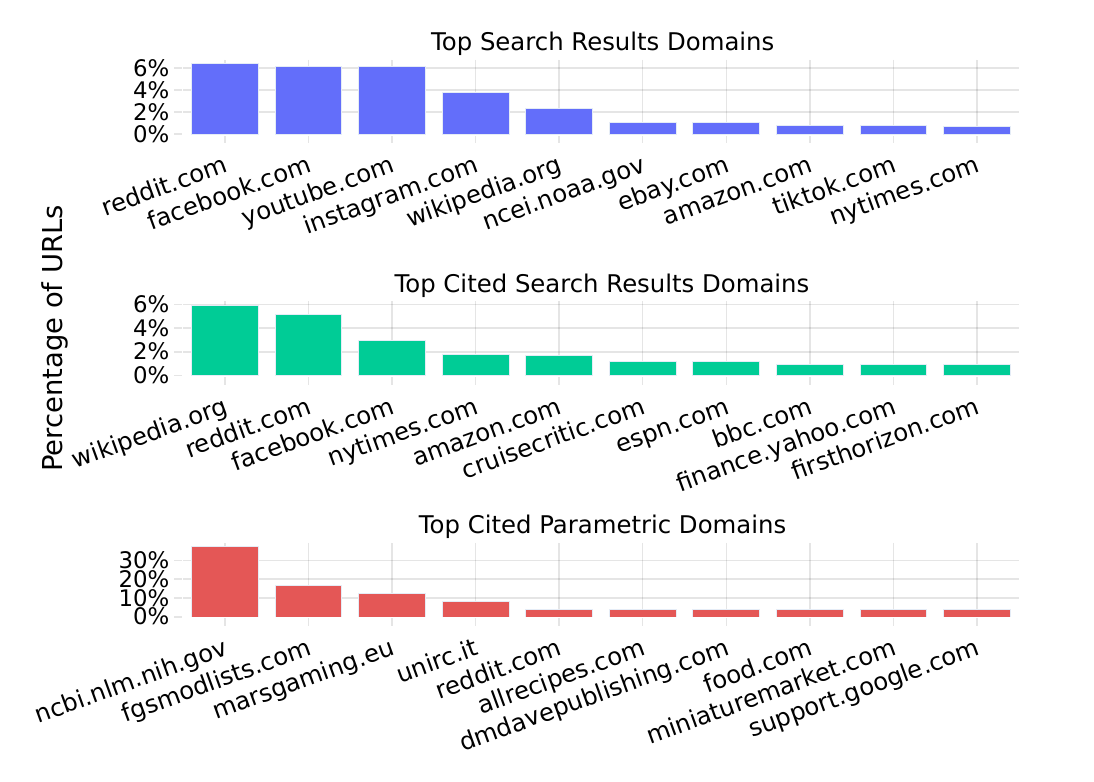}
    \caption{Grok}
    \label{fig:top_domains_grok_insitu_common}
    \end{subfigure}
    \centering
    \begin{subfigure}{0.48\linewidth}
    \centering
    \includegraphics[width=1\linewidth]{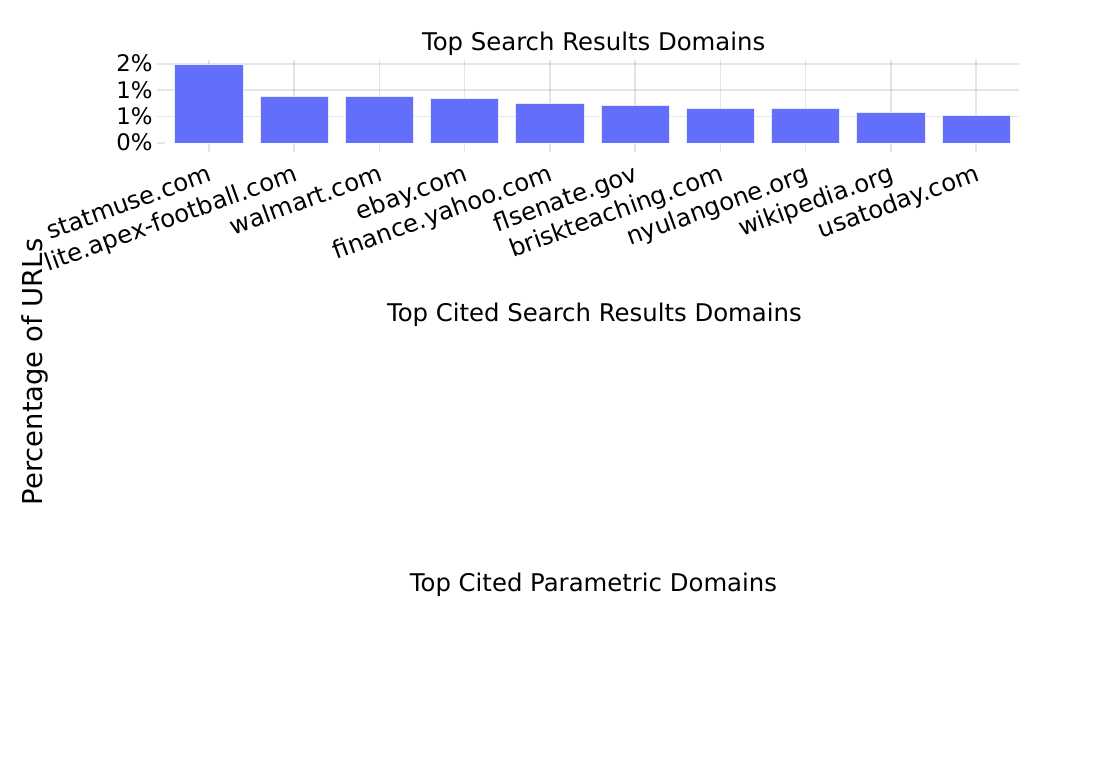}
    \caption{DeepSeek}
    \label{fig:top_domains_deepseek_insitu_common}
    \end{subfigure}
    
\caption{Top 10 domains across platforms \invitro\ over the intersection of all web calling samples.}
\label{fig:top_domains_other_platforms_insitu_common}
\end{figure*}

Figures~\ref{fig:top_domains_other_platforms_insitu} and~\ref{fig:top_domains_other_platforms_insitu_common} present the distribution of search result and cited domains for the \invitro\ experiments (over all web calling samples and common web calling samples across models, respectively). Because every model receives the same user prompts, the observed differences primarily reflect platform-specific search systems rather than differences in user behavior. Similar to the \invivo{} observations, differences exist in the domains emphasized by different search systems. More specifically, Figure~\ref{fig:top_retrieved_domains_other_platforms_invivo_insitu} shows top 10 domains preferred by the search engines.

\subsection{Search Results and Citation Ranks}
\label{app:tranco_ranks_insitu}

\begin{figure*}[ht]
    \centering
    \begin{subfigure}[b]{0.3\linewidth}
        \centering
        \includegraphics[width=1\columnwidth]{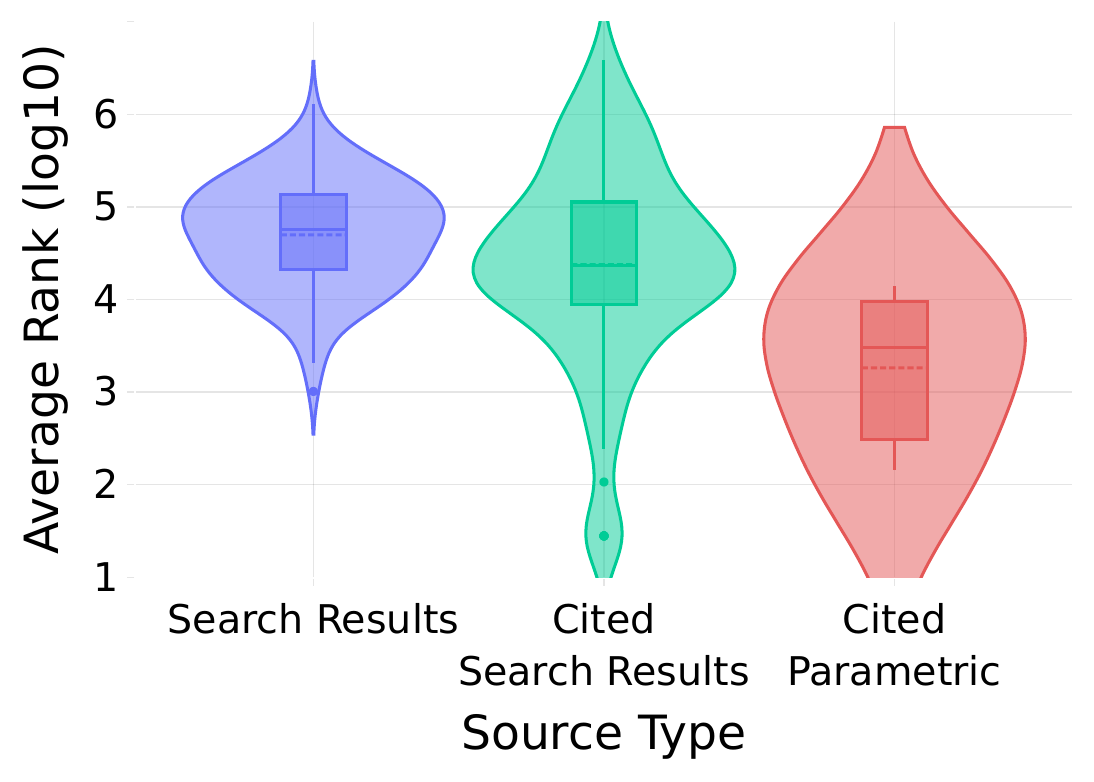}
        \caption{ChatGPT}
        \label{fig:source_rank_violinplot_split_cited_openai_insitu}
    \end{subfigure}
    \begin{subfigure}[b]{0.3\linewidth}
        \centering
        \includegraphics[width=1\columnwidth]{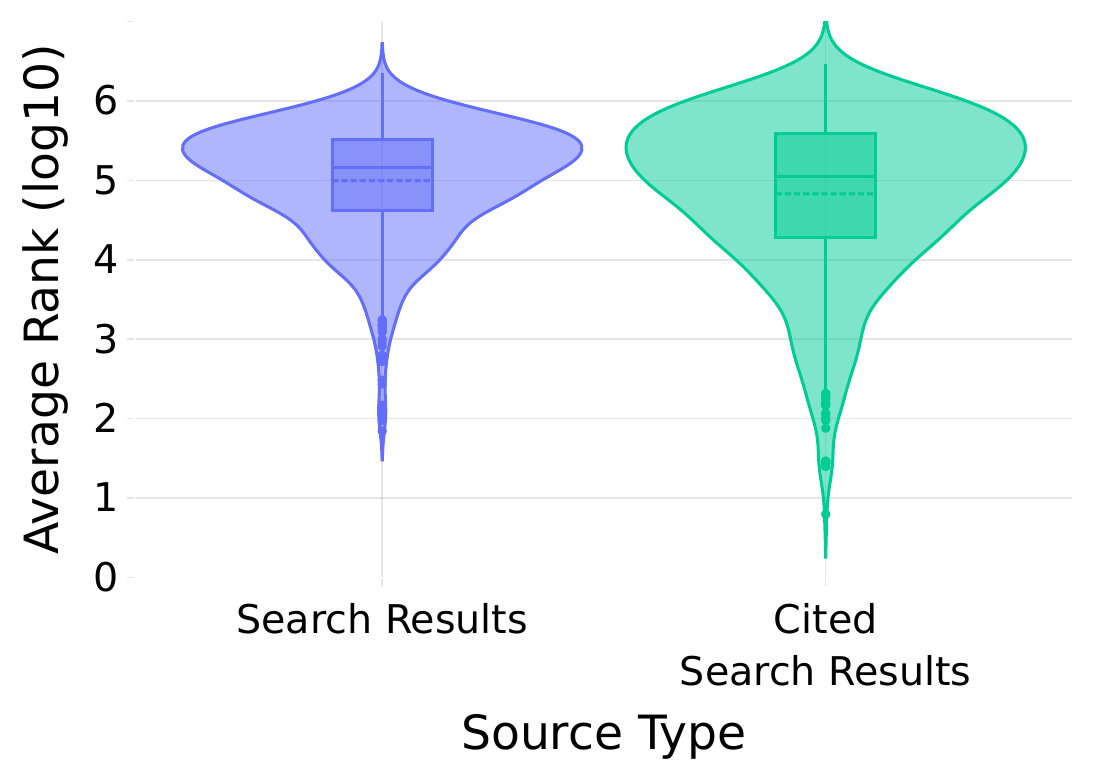}
        \caption{Claude}
        \label{fig:source_rank_violinplot_split_cited_claude_insitu}
    \end{subfigure}
    \begin{subfigure}[b]{0.3\linewidth}
        \centering
        \includegraphics[width=1\columnwidth]{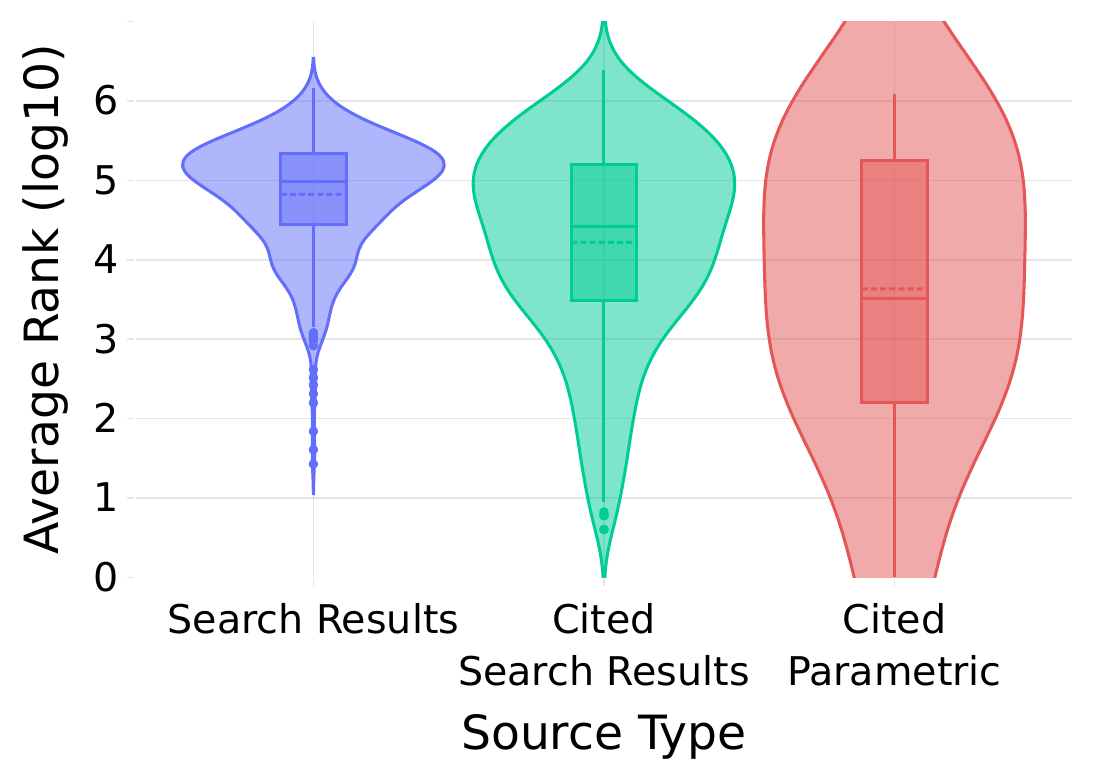}
        \caption{Grok}
    \label{fig:source_rank_violinplot_split_cited_grok_insitu}
    \end{subfigure}
    \caption{Tranco ranks of Search Results and Cited domains \invitro.}
    \vspace{-1mm}
    \label{fig:source_rank_violinplot_tranco_insitu_rev}
\end{figure*}

Figure~\ref{fig:source_rank_violinplot_tranco_insitu_rev} compares the Tranco ranks of search result URLs and cited URLs across different models. Consistent with the \invivo{} setting, cited URLs generally originate from higher-ranked domains than the overall search results, indicating that conversational agents preferentially rely on more authoritative sources when generating responses.

\subsection{Search Result Statistics}
\label{app:search-results_insitu}

\begin{figure}[t]
    \centering
    \begin{subfigure}[b]{0.48\columnwidth}
        \centering
        \includegraphics[width=\linewidth]{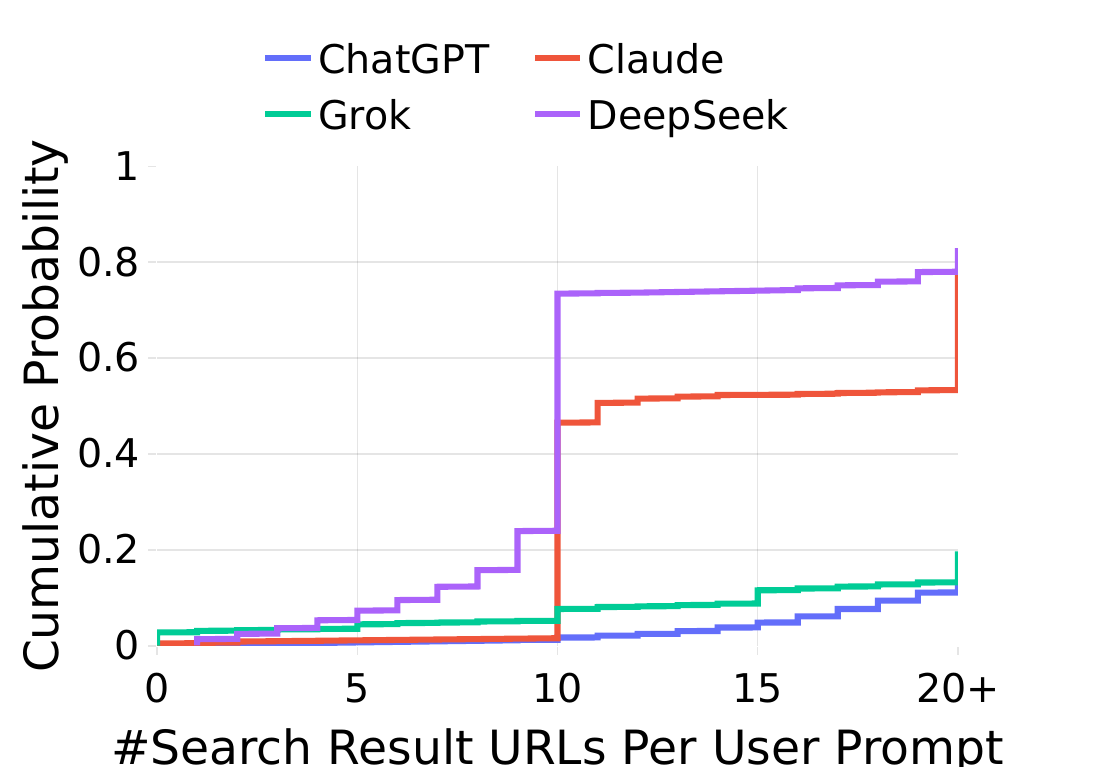}
        \caption{Per User Prompt}
        \label{fig:retrieved_urls_per_prompt_cdf}
    \end{subfigure}
    \begin{subfigure}[b]{0.48\columnwidth}
        \centering
        \includegraphics[width=\linewidth]{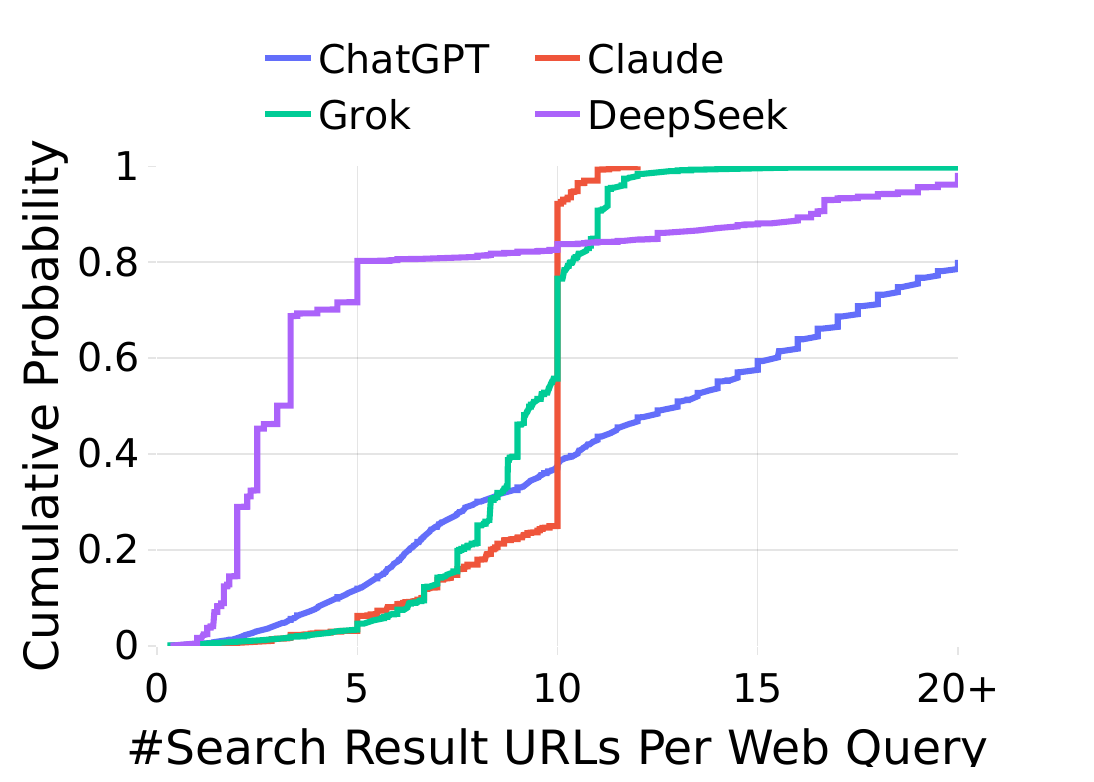}
        \caption{Per Web Query}
        \label{fig:retrieved_urls_per_web_query_cdf_invivo}
    \end{subfigure}
    \begin{subfigure}[b]{0.48\columnwidth}
        \centering
        \includegraphics[width=\linewidth]{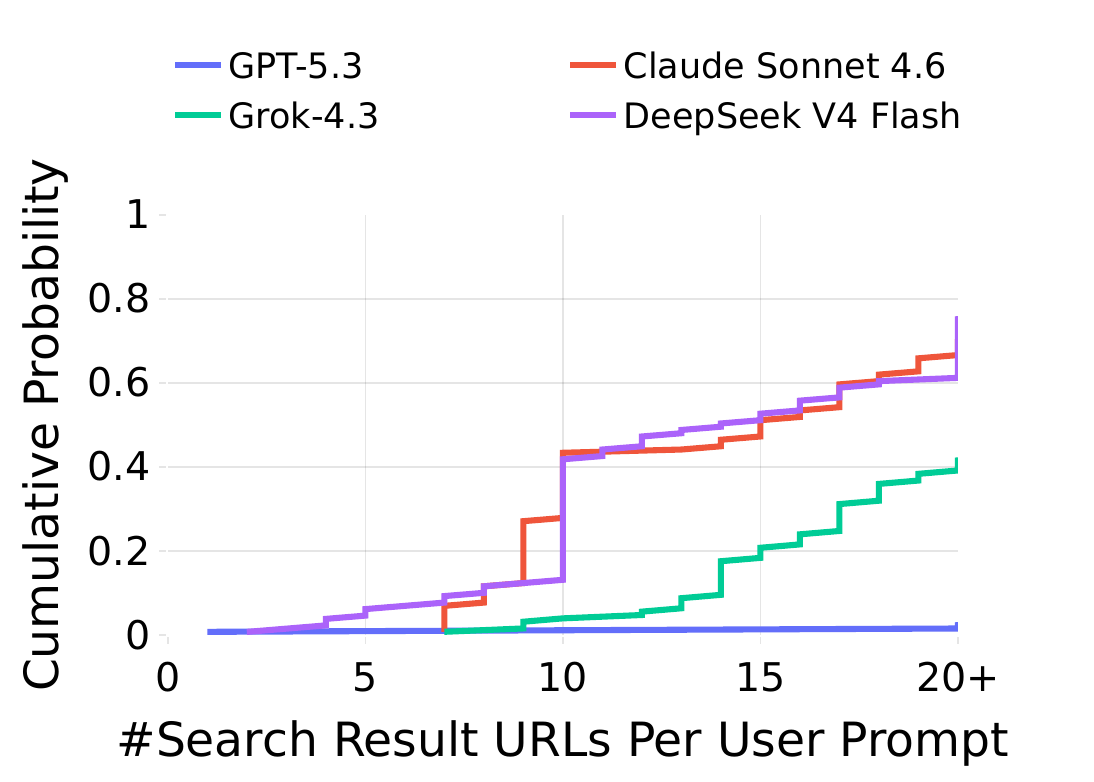}
        \caption{Per User Prompt}
        \label{fig:retrieved_urls_per_prompt_cdf_replay}
    \end{subfigure}
    \begin{subfigure}[b]{0.48\columnwidth}
        \centering
        \includegraphics[width=\linewidth]{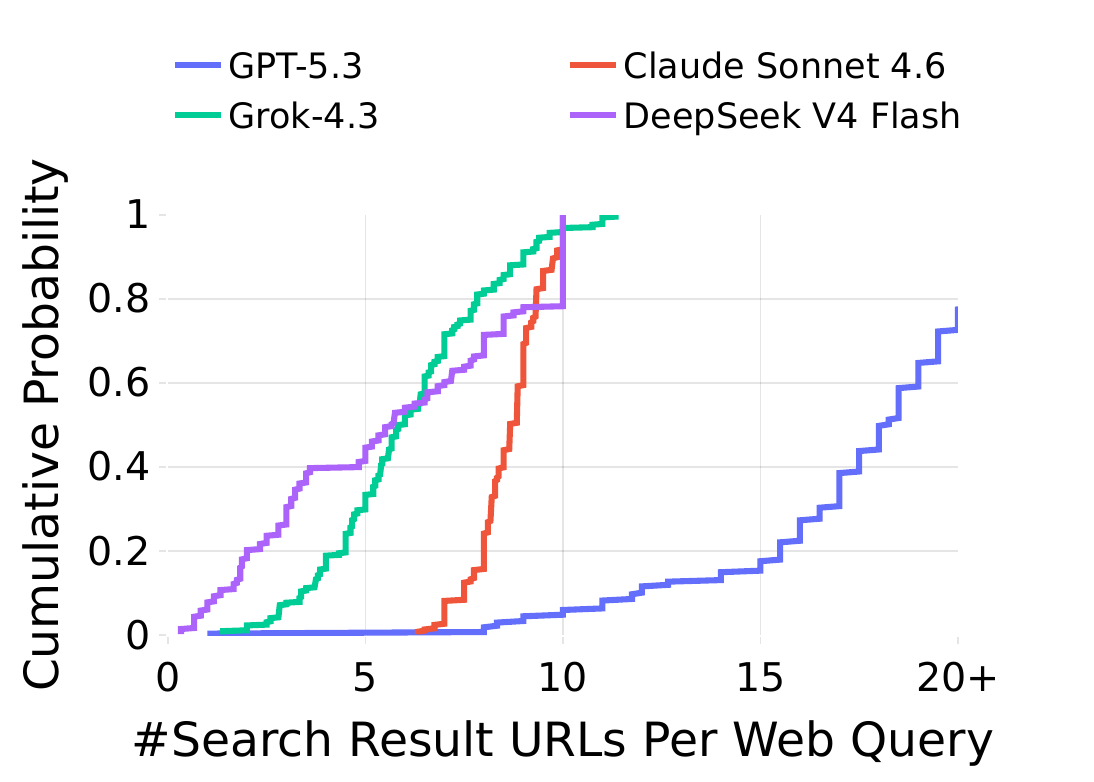}
        \caption{Per Web Query}
        \label{fig:retrieved_urls_per_web_query_cdf_replay}
    \end{subfigure}
  
    \caption{Distribution of search result URLs per user and Web query across platforms in \invivo{} and \invitro{} setting. The search results are vastly different exposing the difference in underlying search systems.}
    \label{fig:retrieved_urls_per_web_query_cdf}
\end{figure}

Table~\ref{tab:queries-retrieved_insitu} and Figure~\ref{fig:retrieved_urls_per_web_query_cdf} summarize the amount of evidence from search results by each model. \invitro{} (all) denotes results over all 1000 samples, while \invitro{} (common) denotes results over all common Web calling samples across all platforms. Despite receiving identical user prompts, these models' search results lead to substantially different numbers of URLs per query and per response. These differences mirror the platform-specific search behavior observed in the \invivo{} dataset and further demonstrate that search strategies are determined not only by user prompts but also by the deployed search system.

\begin{table}[t]
\centering
\begin{adjustbox}{max width=\columnwidth}
\begin{tabular}{p{1.5cm}lrrrr}
\toprule
\textbf{Category} &
\textbf{Model} &
\makecell{\textbf{\#Web} \\ \textbf{Queries}} &
\makecell{\textbf{Avg. \#Web Queries} \\ \textbf{/User Prompt} \\
\textbf{(95\% CI)}} &
\makecell{\textbf{Avg. \#URLs} \\ \textbf{/Web Query} \\
\textbf{(95\% CI)}} &
\makecell{\textbf{Avg. \#URLs} \\ \textbf{/User Prompt} \\
\textbf{(95\% CI)}} \\
\midrule
\multirow{4}{*}{\makecell{\invitro \\ (all)}}
& GPT-5.3
& 279
& 2.11 {\color{black}\scriptsize(2.04--2.20)}
& 17.71 {\color{black}\scriptsize(17.27--18.14)}
& 37.42 {\color{black}\scriptsize(36.14--38.71)}
 \\

& Claude Sonnet 4.6
& 2,000
& 2.42 {\color{black}\scriptsize(2.27--2.59)}
& 8.54 {\color{black}\scriptsize(8.49--8.60)}
& 20.71 {\color{black}\scriptsize(19.39--22.09)}
 \\

& Grok-4.3
& 2,885
& 3.80 {\color{black}\scriptsize(3.68--3.91)}
& 5.92 {\color{black}\scriptsize(5.85--6.00)}
& 22.49 {\color{black}\scriptsize(21.78--23.22)}
 \\

& DeepSeek V4 Flash
& 1,657
& 2.84 {\color{black}\scriptsize(2.67--3.01)}
& 6.39 {\color{black}\scriptsize(6.23--6.55)}
& 18.12 {\color{black}\scriptsize(17.22--19.04)}
 \\

\midrule

\multirow{4}{*}{\makecell{\invitro \\ (common)}}
& GPT-5.3
& 267
& 2.12 {\color{black}\scriptsize(2.04--2.21)}
& 17.61 {\color{black}\scriptsize(17.17--18.05)}
& 37.32 {\color{black}\scriptsize(36.01--38.60)} \\

& Claude Sonnet 4.6
& 368
& 2.85 {\color{black}\scriptsize(2.35--3.40)}
& 8.63 {\color{black}\scriptsize(8.55--8.72)}
& 24.63 {\color{black}\scriptsize(20.26--29.36)}
 \\

& Grok-4.3
& 518
& 4.14 {\color{black}\scriptsize(3.86--4.42)}
& 6.09 {\color{black}\scriptsize(5.91--6.28)}
& 25.25 {\color{black}\scriptsize(23.40--27.11)}
 \\

& DeepSeek V4 Flash
& 410
& 3.18 {\color{black}\scriptsize(2.79--3.57)}
& 5.54 {\color{black}\scriptsize(5.21--5.86)}
& 17.60 {\color{black}\scriptsize(15.70--19.63)}
 \\
\bottomrule
\end{tabular}
\end{adjustbox}
\caption{Web-search querying and search results across model \invitro.}
\label{tab:queries-retrieved_insitu}
\end{table}


%% file: human_validation.tex
\section{Human Annotation and Judge Validation}
\label{sec:human_validation}

To validate the reliability of the judge-based evaluations used throughout this work, we conducted a human annotation study with 3 authors as annotators across multiple tasks, including web-trigger identification, query categorization, query reformulation analysis, query specificity, replay evaluation, and response entailment. 

We use the same set of 30 sampled user prompts across the web-triggers, query-type, replay-evaluation, and entailment-validation tasks to ensure consistency across evaluations. For query reformulation analysis, we annotate the first-iteration web queries generated from these prompts, resulting in 46 reformulation instances. For entailment evaluation, annotators additionally assess whether cited response chunks are supported by their associated retrieved sources.

Three independent annotators participated in the study. For categorical tasks, we aggregate annotations using majority voting. For replay evaluation metrics (factuality, relevance, and completeness), which use 5-point Likert scales, we aggregate annotations using the median score.

Table~\ref{tab:manual_annotation_agreement} summarizes agreement statistics across tasks. We report pairwise human agreement, Fleiss’ $\kappa$ for categorical annotations, and judge agreement with aggregated human labels. For replay evaluation metrics, we report pairwise within-1 agreement between annotators, average pairwise quadratic weighted $\kappa$ as the reliability measure, and within-1 agreement between the judge and the human median score.

\begin{table*}[ht]
\centering
\small
\begin{tabular}{lccccc}
\toprule
\textbf{Task} & \textbf{\#Items} & \textbf{Aggregation} & \textbf{Human Agreement} & \textbf{Reliability} & \textbf{Judge Agreement} \\
\midrule
Policy Trigger Identification & 30 & Majority & 0.778 & 0.705 & 0.964 \\
Query Specificity (Temporal) & 30 & Median  & 0.956  & 0.686  &  0.967 \\
Query Specificity (Geographic) & 30 & Median  & 0.933   &  0.527 & 1.00  \\
Query Specificity (Entity) & 30 & Median  & 0.856  &  0.416 &  0.900 \\
Query Type Classification & 30 & Majority & 0.778 & 0.566 & 0.929 \\
Query Reformulation Reason & 46 & Majority & 0.825 & 0.590 & 0.881 \\
Replay Factuality & 30 & Median & 0.867 & 0.005 & 0.667 \\
Replay Relevance & 30 & Median & 0.833 & 0.223 & 0.767 \\
Replay Completeness & 30 & Median & 0.833 & 0.332 & 0.733 \\
Response Entailment (NLI) & 30 & Majority & 0.782 & 0.167 & 0.931 \\
\bottomrule
\end{tabular}
\caption{
Human annotation agreement and judge validation across evaluation tasks. Human agreement denotes pairwise annotator agreement. Reliability corresponds to Fleiss’ $\kappa$ for categorical tasks and average pairwise quadratic weighted $\kappa$ for replay evaluation metrics; for the latter, lower $\kappa$ values can arise even when within-1 agreement is high if annotators use the ordinal scale with consistent offsets. Judge agreement measures agreement between the automated judge and aggregated human annotations.
}
\label{tab:manual_annotation_agreement}
\end{table*}

Table~\ref{tab:manual_annotation_agreement} illustrates the high agreement among our human annotators and the judge. These results provide additional evidence that the judge model produces evaluations that closely align with human assessments, supporting its use throughout this work.

%% file: prompts.tex
\section{Prompts}
\label{sec:prompts}

In this section, we have provided all the prompts used in our evaluation.
To prevent cross-platform data leakage, we used a different judge model for each platform, selecting models of comparable capability: \texttt{GPT-4o-mini} for ChatGPT, \texttt{Claude-Haiku-3.5} for Claude, \texttt{Grok-3-mini} for Grok, and \texttt{Deepseek-chat} for DeepSeek.

\subsection{PII Identification}
\label{app:pii_id_prompt}

The PII identification prompt is shown in Fig.~\ref{fig:pii_prompt}.

\begin{figure*}[t]
\centering
\tiny
\begin{tcolorbox}[colback=gray!5, width=\textwidth, colframe=gray!80, title= Prompt for PII Identification]
\begin{verbatim}
You are an annotator that identifies the personal data or sensitive data in a two-party conversation (User A and User B). 
Return ONLY valid JSON; do not include explanations outside JSON.

## GDPR Reference

- **GDPR Article 4(1)** defines personal data as any information relating to an identified or identifiable natural person, including identifiers such as
a name, an identification number, location data, an online identifier, or to one or more factors specific to the physical, physiological, genetic,
mental, economic, cultural or social identity of that person.
- **GDPR Article 9(1)** defines special category personal data as data revealing racial or ethnic origin, political opinions, religious or philosophical
beliefs, or trade union membership, and the processing of genetic data, biometric data, data concerning health, or data concerning a natural person’s sex
life or sexual orientation of that person. 

--- 

## Questions to Answer

We are providing only the list of **User A’s messages**. For each message, answer the following:

1) Does the attached conversation have any personal information of **User A** as defined by GDPR Article 4(1)?  
   - Answer "Yes" or "No".  
   - If "Yes", specify the **type(s) of personal data** and the **exact instance(s) from the message** in a list. Choose the type of data from below 
   (exact strings):
     Name | Birth Information | Phone Number | Email Address | Location | Online Identifiers | Economic or Financial Information | Educational Information
     | Employment Information | Social Identity | Business or Project Information | Physical Identity (Hair, body weight, height..etc) | Cultural Identity
     | Social Identity | Travel history | Animals/Pet Information | Family/Friends Information (including name, health, relationship..etc) | Device 
     Information | Other Identification numbers
   - If Other, please specify the type.

2) Does the attached conversation have any special categories of personal data of **User A** as defined by GDPR Article 9(1)?  
   - Answer "Yes" or "No".  
   - If "Yes", specify the **type(s) of special category data** and the **exact instance(s) from the message** in a list. Choose the type of data from 
   below (exact strings):
     Political opinions | Racial or ethnic origin | Sex life | Sexual Orientation | Relationship details | Mental Health | Other Health Information |
     Biometric data | Genetic data | Religious beliefs | Philosophical beliefs | Personal views and feelings
   - If Other, please specify the type.

--- 

## Important Notes

-  Only analyze **User A’s messages** (the list provided).  
-  Use the schema exactly as provided.  
-  Do not include explanations outside the JSON object.

---

## Output JSON schema (and nothing else):
{
  "data_per_turn": [
    {
      "turn_index": <int, 0-based>,
      "personal_data": {
            "present": "Yes|No",
            "types": [
                {
                    "type": "<type of personal data>",
                    "instance": "<exact instance from message>"
                }
                ....
            ]
        },
        "special_category_data": {
            "present": "Yes|No",
            "types": [
                {
                    "type": "<type of special category data>",
                    "instance": "<exact instance from message>"
                }
                ....
            ]
        }
    }
    ...
  ]
}
\end{verbatim}
\end{tcolorbox}
\caption{Prompt for PII identification.}
\label{fig:pii_prompt}
\end{figure*}

\subsection{Factuality, Completeness, and Relevance Evaluations}
\label{sec:prompts_metrics}
The system prompts for the factuality, completeness, and relevance evaluations are shown in Figs.~\ref{fig:factuality_prompt}, \ref{fig:completeness_prompt}, and \ref{fig:relevance_prompt}, respectively. The user prompt is shared across all three evaluations and is shown in Fig.~\ref{fig:user_prompt_three}.

\begin{figure*}[t]
\centering
\begin{tcolorbox}[colback=gray!5, width=\textwidth, colframe=gray!80, title=System Prompt for Factuality Evaluation]
\begin{verbatim}
You are an evaluator assessing the factual correctness of an AI-generated response 
to a user query.

Evaluate:
- Is the response factually correct and free from hallucinations or false claims?
- Is the information up-to-date and not outdated when recency matters?

Return JSON:

{
"score": 1-5,
"reasoning": "<1-2 sentence explanation>"
}

Scoring guide:
1 = Mostly incorrect or clearly hallucinated; core claims are wrong
2 = More incorrect than correct; contains significant factual errors that undermine 
the answer, even if some parts are right
3 = Mixed accuracy; contains both correct and incorrect claims of similar importance
4 = Mostly correct; minor inaccuracies or slightly outdated details that do not 
change the overall answer
5 = Fully correct, precise, and up-to-date; no meaningful errors

Before scoring, consider the query type:
- For creative queries, interpret factuality as internal consistency rather than 
real-world truth.
\end{verbatim}
\end{tcolorbox}
\caption{System prompt for factuality evaluation.}
\label{fig:factuality_prompt}
\end{figure*}

\begin{figure*}[t]
\centering
\begin{tcolorbox}[colback=gray!5, 
width=\textwidth, colframe=gray!80, title=System Prompt for Completeness Evaluation]
\begin{verbatim}
You are an evaluator assessing the completeness of an AI-generated response to a 
user query.

Evaluate:
- Does the response fully address and cover all parts of the user’s question?

Return JSON:

{
  "score": 1-5,
  "reasoning": "<1-2 sentence explanation>"
}

Scoring guide:
1 = Very incomplete; misses most parts of the question or fails to address the main 
request
2 = Partially incomplete; addresses some parts but omits major components of the 
question
3 = Moderately complete; covers the main request but misses some secondary aspects 
or details
4 = Mostly complete; addresses nearly all parts with only minor omissions
5 = Fully complete; covers all aspects of the question thoroughly

Before scoring, consider the query type:
- For open-ended queries, interpret completeness as reasonable coverage of key 
aspects, not exhaustiveness.
\end{verbatim}
\end{tcolorbox}
\caption{System prompt for completeness evaluation.}
\label{fig:completeness_prompt}
\end{figure*}

\begin{figure*}[t]
\centering
\begin{tcolorbox}[colback=gray!5, width=\textwidth, colframe=gray!80, title=System Prompt for Relevance Evaluation]
\begin{verbatim}
You are an evaluator assessing how relevant an AI-generated response is to a user 
query.

Evaluate:
- Does the response directly address the user's question or intent?
- Is the response concise, to the point, and free from off-topic or unnecessary 
information?

Return JSON:

{
"score": 1-5,
"reasoning": "<1-2 sentence explanation>"
}

Scoring guide:
1 = Irrelevant; does not address the user’s question or intent at all
2 = Weakly relevant; touches on the topic but largely misses the user’s intent 
or includes substantial off-topic content
3 = Partially relevant; addresses the main intent but includes noticeable 
irrelevance or digressions
4 = Mostly relevant; well-aligned with the intent with only minor off-topic or 
unnecessary details
5 = Fully relevant; directly and precisely addresses the user’s intent with no 
unnecessary content
\end{verbatim}
\end{tcolorbox}
\caption{System prompt for relevance evaluation.}
\label{fig:relevance_prompt}
\end{figure*}

\begin{figure*}[t]
\centering
\begin{tcolorbox}[colback=gray!5, width=\textwidth, colframe=gray!80, title=User Prompt for Factuality/Completeness/Relevance Evaluation]
\begin{verbatim}
Evaluate the following.

User Query:
{user_query}

AI Response:
{response}

Return ONLY valid JSON in this exact format:
{
"score": <integer 1-5>,
"reasoning": "<string>"
}

Rules:
- Do not include any text outside the JSON
- Do not add explanations before or after
- Ensure the JSON is valid
\end{verbatim}
\end{tcolorbox}
\caption{User prompt for the three evaluations: factuality, completeness, and relevance.}
\label{fig:user_prompt_three}
\end{figure*}

\subsection{System Prompts of GPT models}
\label{sec:system_prompt_gpt_models}
The system prompts for the GPT models, including GPT-5.3-Chat, GPT-4.1-mini, and o4-mini, are shown in Figs.~\ref{fig:prompt_extracted_systemprompt_gpt5.3chat}, \ref{fig:prompt_extracted_systemprompt_gpt4.1mini}, and \ref{fig:prompt_extracted_systemprompt_o4mini}, respectively.

\begin{figure*}[t]
\centering
\tiny
\begin{tcolorbox}[colback=gray!5, width=\textwidth, colframe=gray!80, title=Extracted System Prompt from GPT-5.3-Chat]
\begin{verbatim}
# Web Tool Usage

Use the `web` tool according to the following triggers.

## (1) Volatile/Temporal Information — Time-sensitive or frequently changing info (e.g., news, weather, prices, sports, policies, releases).

- Information that are fresh, current, or time-sensitive.
- Information that are could change over time and must be verified by web searches at the time of the request.
- Contemporary people info. celebrities, politicians, LinkedIn profiles, recent works.
- Requests for Opinions, Reviews, Recommendations, and information that often rely on changing trends or community sentiment.

## (2) Unfamiliar Term/Typo — Rare, ambiguous, or possibly misspelled terms requiring lookup.

- Requests for information about named Entities, Public Figures, Companies, Brands, Products, Services, Places, etc.

## (3) High-Investment Recommendation — Decisions involving significant time, money, or commitment (e.g., travel, purchases, services).

- Information in domains that require fresh and accurate data, including:
  - Local or travel queries. For example: restaurants near me, shops, hotels, operating hours, itineraries, localized time, etc.
- Requests related to physical retail products (e.g. Fashion, Clothing, Apparel, Electronics, Home & Living, Food & Beverage, Auto Parts), including 
(but not limited to) product searches, recommendation or comparisons, price look-ups, general information about products, etc.

## (4) Attribution/Sourcing Needed — Requires verifiable sources, citations, quotes, or links.

- Requests for online resources, such as tools, tutorials, courses, manuals, documentations, reference materials, social updates, etc.

## (5) External Reference — Mentions a specific external resource not included in the prompt (e.g., URL, paper, dataset).

- Navigational queries, where the user is requesting links to particular site or page. For example, queries that are just short names of websites,
brands, and entities, such as "instagram", "openai", "apple", "wiki", "booking", "white house".
- Data retrieval tasks, such as accessing specific external websites, pages, documents, or summarizing information from a given URL.

## (6) Low Confidence/Niche Fact — Obscure, highly specific, or emerging topics with high hallucination risk.

- Requests for deep / comprehensive research into a subject.
- Difficult questions where you might be able to improve by drawing on external sources.

## (7) High-Stakes Accuracy — Medical, legal, or financial queries where errors could cause harm.

- High stakes queries. You must use the web for verification if factual inaccuracies in your response 
could lead to serious consequences, e.g. legal matters, regulations, policies, financial matters, medical matters, election results, government 
office-holders, etc.

## (8) User Verification — User asks to confirm, validate, or fact-check information.

- Information that should be specific, accurate, verifiable, and trustworthy. Fact-checking using the web are required for such information even if the 
information are considered not changing over time.

## (9) Explicit Command — User explicitly asks to search, browse, or check online.

- If the user makes an explicit request to search the internet, find latest information, look up, etc, you must obey their request.
- If the user asks you to not access the web, then you must not use this tool.

Do **not** use the `web` tool when web information would not help answer the user's request. Examples include:

- Greetings, pleasantries, and other casual chatting.
- Non-informational requests.
- Creative writing when no references are required.
- Requests to rewrite, summarize, or translate text that is already provided.
- Requests towards other tools other than the `web` tool.
- Questions about yourself, your own opinions, or purely internal analysis.
\end{verbatim}
\end{tcolorbox}
\caption{Extracted system prompt from GPT-5.3-Chat}
\label{fig:prompt_extracted_systemprompt_gpt5.3chat}
\end{figure*}

\begin{figure*}[t]
\centering
\small
\begin{tcolorbox}[colback=gray!5, width=\textwidth, colframe=gray!80, title=Extracted System Prompt from GPT-4.1-mini]
\begin{verbatim}
# Web Tool Usage

Use the `web` tool according to the following triggers.

## (1) Volatile/Temporal Information — Time-sensitive or frequently changing info (e.g., news, 
weather, prices, sports, policies, releases).

- **Freshness:** If up-to-date information on a topic could potentially change or enhance the answer, 
call the `web` tool any time you would otherwise refuse to answer a question because your knowledge
might be out of date.

## (2) High-Investment Recommendation — Decisions involving significant time, money, or commitment 
(e.g., travel, purchases, services).

- **Local Information:** Use the `web` tool to respond to questions that require information about 
the user's location, such as the weather, local businesses, or events.

## (3) Low Confidence/Niche Fact — Obscure, highly specific, or emerging topics with high 
hallucination risk.

- **Niche Information:** If the answer would benefit from detailed information not widely known 
or understood (such as details about a small neighborhood, a less well-known company, or arcane
regulations), use web sources directly rather than relying on the distilled knowledge from 
pretraining.

## (4) High-Stakes Accuracy — Medical, legal, or financial queries where errors could cause harm.

- **Accuracy:** If the cost of a small mistake or outdated information is high (e.g., using an 
outdated version of a software library or not knowing the date of the next game for a sports team),
then use the `web` tool.
\end{verbatim}
\end{tcolorbox}
\caption{Extracted system prompt from GPT-4.1-mini}
\label{fig:prompt_extracted_systemprompt_gpt4.1mini}
\end{figure*}

\begin{figure*}[t]
\centering
\small
\begin{tcolorbox}[colback=gray!5, width=\textwidth, colframe=gray!80, title=Extracted System Prompt from o4-mini]
\begin{verbatim}
# Web Tool Usage

Use the `web` tool according to the following triggers.

## (1) Volatile/Temporal Information — Time-sensitive or frequently changing info (e.g., news, 
weather, prices, sports, policies, releases).

- You *must* browse the web for *any* query that could benefit from up-to-date information.
- Example topics include but are not limited to politics, current events, weather, sports, scientific 
developments, cultural trends, recent media or entertainment developments, general news, or many many 
other types of questions.
- It's absolutely critical that you browse, using the `web` tool, *any* time you are remotely
uncertain if your knowledge is up-to-date and complete.
- If the user asks about the "latest" anything, you should likely be browsing.
- If the user makes any request that requires information after your knowledge cutoff, that requires
browsing.
- Further, you *must* also browse for high-level, generic queries about topics that might plausibly
be in the news (e.g. "Apple", "large language models", etc.).
- If you are asked to do something that requires up-to-date knowledge as an intermediate step, it's
also CRUCIAL you browse in this case.
- For example, if the user asks to generate a picture of the current president, you still must browse 
with the `web` tool to check who that is; your knowledge is very likely out of date for this and many 
other cases!
- Remember, you MUST browse (using the `web` tool) if the query relates to current events in politics, 
sports, scientific or cultural developments, or ANY other dynamic topics.

## (2) External Reference — Mentions a specific external resource not included in the prompt (e.g., 
URL, paper, dataset).

- You *must* also browse for navigational queries (e.g. "YouTube", "Walmart site").

## (3) Low Confidence/Niche Fact — Obscure, highly specific, or emerging topics with high 
hallucination risk.

- You *must* browse the web for *any* query that could benefit from niche information.
- Example topics include esoteric topics and deep research questions.

## (4) High-Stakes Accuracy — Medical, legal, or financial queries where errors could cause harm.

- Incorrect or out-of-date information can be very frustrating (or even harmful) to users.

## (5) Explicit Command — User explicitly asks to search, browse, or check online.

- Unless the user explicitly asks you not to browse the web.
- Err on the side of over-browsing, unless the user tells you not to browse.
\end{verbatim}
\end{tcolorbox}
\caption{Extracted system prompt from o4-mini}
\label{fig:prompt_extracted_systemprompt_o4mini}
\end{figure*}

\subsection{Query Specificity}
\label{sec:query_specificity_prompt}
We evaluate query specificity along three dimensions: temporal, geographic, and entity. The system and user prompts for the temporal evaluation are shown in Figs.~\ref{fig:system_prompt_temporal} and \ref{fig:user_prompt_temporal}, respectively. The corresponding prompts for the geographic evaluation are shown in Figs.~\ref{fig:system_prompt_geographic} and \ref{fig:user_prompt_geographic}. The corresponding prompts for the entity evaluation are shown in Figs.~\ref{fig:system_prompt_entity} and \ref{fig:User_prompt_entity}.

\begin{figure*}[t]
\centering
\small
\begin{tcolorbox}[colback=gray!5, width=\textwidth, colframe=gray!80, title=System Prompt for Temporal Specificity]
\begin{verbatim}
You are an impartial evaluator whose only task is to measure the temporal 
specificity of a grounding query.

Temporal specificity measures how precisely the query specifies WHEN the requested 
information is relevant.

Assign exactly one score from 1 to 5.

Scoring rubric:

1 — No time reference.
Examples:
- laptop price
- weather
- best restaurants

2 — Broad or vague timeframe.
Examples:
- latest
- recent
- upcoming
- modern

3 — Relative time.
Examples:
- today
- yesterday
- this week
- last month
- next weekend

4 — Specific date or month/year.
Examples:
- June 2026
- March 15
- 2025 Q4

5 — Exact date and time.
Examples:
- June 3, 2026
- June 3, 2026 9:00 AM EST
- 2025-11-15T14:30 UTC

Evaluation rules:
- Judge only the text of the query.
- Do not infer missing temporal information.
- Ignore all non-temporal aspects.

Return only valid JSON.

{{
  "score": <1-5>,
  "reason": "<one concise sentence>"
}}
\end{verbatim}
\end{tcolorbox}
\caption{System prompt for temporal specificity}
\label{fig:system_prompt_temporal}
\end{figure*}

\begin{figure*}[t]
\centering
\begin{tcolorbox}[colback=gray!5, width=\textwidth, colframe=gray!80, title=User Prompt for Temporal Specificity]
\begin{verbatim}
Evaluate the temporal specificity of the following grounding query.

Grounding Query:
{QUERY}

Use the temporal specificity rubric provided in the system instructions.

Return ONLY valid JSON in the following format:

{{
  "score": <integer between 1 and 5>,
  "reason": "<one concise sentence explaining the score>"
}}
\end{verbatim}
\end{tcolorbox}
\caption{User prompt for temporal specificity}
\label{fig:user_prompt_temporal}
\end{figure*}

\begin{figure*}[t]
\centering
\small
\begin{tcolorbox}[colback=gray!5, width=\textwidth, colframe=gray!80, title=System Prompt for Geographic Specificity]
\begin{verbatim}
You are an impartial evaluator whose only task is to measure the geographic 
specificity of a grounding query.

Geographic specificity measures how precisely the query specifies WHERE the 
requested information applies.

Assign exactly one score from 1 to 5.

Scoring rubric:

1 — No location.
Examples:
- best restaurants
- weather
- laptop price

2 — Large region or continent.
Examples:
- Europe
- North America
- Southeast Asia

3 — Country, state, or province.
Examples:
- Japan
- California
- Germany

4 — City or locality.
Examples:
- Seattle
- Tokyo
- Manhattan

5 — Exact place.
Examples:
- Pike Place Market
- Tokyo Station
- 1600 Pennsylvania Avenue
- GPS coordinates

Evaluation rules:
- Judge only the wording of the query.
- Do not infer any location.
- Ignore all non-geographic information.

Return only valid JSON.

{{
  "score": <1-5>,
  "reason": "<one concise sentence>"
}}
\end{verbatim}
\end{tcolorbox}
\caption{System prompt for geographic specificity}
\label{fig:system_prompt_geographic}
\end{figure*}

\begin{figure*}[t]
\centering
\begin{tcolorbox}[colback=gray!5, width=\textwidth, colframe=gray!80, title=User Prompt for Geographic Specificity]
\begin{verbatim}
Evaluate the geographic specificity of the following grounding query.

Grounding Query:
{QUERY}

Use the geographic specificity rubric provided in the system instructions.

Return ONLY valid JSON in the following format:

{{
  "score": <integer between 1 and 5>,
  "reason": "<one concise sentence explaining the score>"
}}
\end{verbatim}
\end{tcolorbox}
\caption{User prompt for geographic specificity}
\label{fig:user_prompt_geographic}
\end{figure*}

\begin{figure*}[t]
\centering
\small
\begin{tcolorbox}[colback=gray!5, width=\textwidth, colframe=gray!80, title=System Prompt for Entity Specificity]
\begin{verbatim}
You are an impartial evaluator whose only task is to measure the entity specificity 
of a grounding query.

Entity specificity measures how precisely the query identifies the object, person, 
organization, product, document, or item being requested.

Assign exactly one score from 1 to 5.

Scoring rubric:

1 — Generic category only.
Examples:
- laptop
- restaurant
- phone

2 — Category with descriptive qualifiers.
Examples:
- gaming laptop
- Italian restaurant
- electric SUV

3 — Named brand, company, organization, or person.
Examples:
- Dell laptop
- Apple Watch
- Starbucks

4 — Product family, model, or uniquely named item.
Examples:
- Dell XPS 15
- iPhone 16 Pro
- Tesla Model Y

5 — Exact model, SKU, identifier, or uniquely identifiable entity.
Examples:
- Dell XPS 15 9530 i7-13700H
- Samsung QE65S95D
- ISBN 9780135957059

Evaluation rules:
- Judge only the wording of the query.
- Do not infer missing identifiers.
- Ignore temporal, geographic, and numeric information.

Return only valid JSON.

{{
  "score": <1-5>,
  "reason": "<one concise sentence>"
}}
\end{verbatim}
\end{tcolorbox}
\caption{System prompt for entity specificity}
\label{fig:system_prompt_entity}
\end{figure*}

\begin{figure*}[t]
\centering
\begin{tcolorbox}[colback=gray!5, width=\textwidth, colframe=gray!80, title=User Prompt for Entity Specificity]
\begin{verbatim}
Evaluate the entity specificity of the following grounding query.

Grounding Query:
{QUERY}

Use the entity specificity rubric provided in the system instructions.

Return ONLY valid JSON in the following format:

{{
  "score": <integer between 1 and 5>,
  "reason": "<one concise sentence explaining the score>"
}}
\end{verbatim}
\end{tcolorbox}
\caption{User prompt for entity specificity}
\label{fig:User_prompt_entity}
\end{figure*}

\subsection{Type of Query}
\label{sec:prompts_types}
The system and user prompts for user \& Web query type classification are shown in Figs.~\ref{fig:System_prompt_user_webquery_type_classification} and \ref{fig:User_prompt_user_webquery_type_classification}, respectively.

\begin{figure*}[t]
\centering
\small
\begin{tcolorbox}[colback=gray!5, width=\textwidth, colframe=gray!80, title=System Prompt for User \& Web Query Type Classification]
\begin{verbatim}
You are an evaluator classifying the intent of a query.

Classify the query into exactly ONE of the following categories:

- "informational":
  The query is primarily seeking information, explanations, facts, answers, or learning content about
  a topic.

- "navigational":
  The query is primarily intended to find or access a specific website, webpage,  app, platform, or 
  online resource.

- "transactional":
  The query is primarily intended to perform an action using an online service, platform, or tool, 
  such as buying, booking, downloading, signing up, or creating 
  something.

- "commercial":
  The query is primarily intended to research or compare products/services with potential purchase 
  intent, but without explicitly attempting to complete a transaction.

Guidelines:
- Choose the SINGLE best category
- Focus on the primary intent of the query

Examples:

Query: "I am shopping for a flight from KUL to Guadalajara..."
→ transactional

Query: "Open the Stanford CS229 course website"
→ navigational

Query: "How can I reset my Instagram password if I don’t have access to my original email account?"
→ informational

Query: "best noise cancelling headphones"
→ commercial

Return ONLY valid JSON:

{{
  "type": "transactional" | "navigational" | "informational" | "commercial",
  "reasoning": "<1-2 sentence explanation>"
}}

Rules:
- Output must be valid JSON
- Do not include any extra text
\end{verbatim}
\end{tcolorbox}
\caption{System prompt for user \& Web query type classification}
\label{fig:System_prompt_user_webquery_type_classification}
\end{figure*}

\begin{figure*}[t]
\centering
\begin{tcolorbox}[colback=gray!5, width=\textwidth, colframe=gray!80, title=User Prompt for User \& Web Query Type Classification]
\begin{verbatim}
Classify the following query into exactly one intent type.

Query:
{user_query}

Return ONLY valid JSON:
{{
  "type": "transactional" | "navigational" | "informational" | "commercial,
  "reasoning": "<string>"
}}

Rules:
- Choose exactly one type
- Focus on the primary intent
- Output valid JSON only
\end{verbatim}
\end{tcolorbox}
\caption{User prompt for user \& Web query type classification}
\label{fig:User_prompt_user_webquery_type_classification}
\end{figure*}

\subsection{Query Reformulation Behaviour}
\label{sec:prompts_reasons}
The system and user prompts for query reformulation behaviour are shown in Figs.~\ref{fig:system_prompt_query_reformulation_behavior} and \ref{fig:user_prompt_query_reformulation_behavior}, respectively.

\begin{figure*}[t]
\centering
\small
\begin{tcolorbox}[colback=gray!5, width=\textwidth, colframe=gray!80, title=System Prompt for Query Reformulation Behaviour]
\begin{verbatim}
You are an expert evaluator of conversational search behavior, specializing in 
query reformulation.

Your task is to label the relationship between specified query transitions based on 
whether the reformulated query improves upon the original query in 1 of the 2 
following ways:

You must choose exactly one of the following categories:

1. Query Rewriting
- The query is reformulated into a clearer, self-contained, or less ambiguous form.
- Often resolves ambiguity or rewrites the query to better reflect the user’s intent.

2. Query Expansion
- The query is augmented with additional terms or context.
- Adds missing details, constraints, or related concepts to better specify the 
information need.

3. Hybrid
- Combines both rewriting and expansion.
- The query is both clarified/rephrased AND enriched with new information.

4. Other
- The reformulated query is neither more clarified nor enriched with new information.
- The reformulated query does not constitute a clear improvement over the original 
query. So, it cannot be labeled either as query rewriting or query expansion.

Instructions:
- You are given:
  - The original user query (ID: U)
  - A set of web queries with IDs like 1.1, 2.1, etc.
  - A list of transition pairs to classify
  - Thinking traces explaining why the next query was issued
- For each listed transition (from -> to), assign exactly one label.
- Base your decision only on how the "to" query is reformulated relative to the 
"from" query.
- Treat U as the original user query text.
- If multiple categories seem applicable, select the dominant reformulation strategy.
- Provide a short reasoning (1–2 sentences) grounded in these definitions.
- Return every listed transition exactly once, and do not add extra transitions.

Output format (STRICT JSON):
{{
  "transitions": [
    {{
      "from": "U",
      "to": "1.1",
      "label": "Query Rewriting | Query Expansion | Hybrid | Other",
      "reasoning": "1-2 sentence explanation"
    }}
  ]
}}
\end{verbatim}
\end{tcolorbox}
\caption{System prompt for query reformulation behaviour}
\label{fig:system_prompt_query_reformulation_behavior}
\end{figure*}

\begin{figure*}[t]
\centering
\scriptsize
\begin{tcolorbox}[colback=gray!5, width=\textwidth, colframe=gray!80, title=User Prompt for Query Reformulation Behaviour]
\begin{verbatim}
Classify the listed transitions using conversational search query reformulation terminology.

Example:

User Query (U):
Best laptops for programming

Web Queries:
(1.1) best laptops for programmers
(2.1) best lightweight laptops for programming students
(2.2) macbook air m3 student programming battery life

Thinking Traces:
(1.1) "I should rephrase this into a direct benchmark-style web query."
(2.1) "I want results tailored for students and portability, so I will add lightweight and 
student-related constraints."
(2.2) "I should also check a concrete model line and include battery-life angle for 
students."

Transitions to classify:
(U -> 1.1)
(1.1 -> 2.1)
(1.1 -> 2.2)

Output:
{{
  "transitions": [
    {{
      "from": "U",
      "to": "1.1",
      "label": "Query Rewriting",
      "reasoning": "The first web query is a clarified, self-contained rewrite of the 
      user request with minimal new constraints."
    }},
    {{
      "from": "1.1",
      "to": "2.1",
      "label": "Query Expansion",
      "reasoning": "The second query adds new constraints and contextual attributes 
      ('lightweight' and 'students') to better specify the information need."
    }},
    {{
      "from": "1.1",
      "to": "2.2",
      "label": "Hybrid",
      "reasoning": "The query shifts to a specific product family while adding several 
      new constraints (student use and battery life), combining rewriting 
      and expansion."
    }}
  ]
}}

Now classify this:

User Query (U):
{user_query}

Web Queries:
{web_queries}

Thinking Traces:
{thinking_traces}

Transitions to classify:
{transition_candidates}

Return ONLY the JSON.
\end{verbatim}
\end{tcolorbox}
\caption{User prompt for query reformulation behaviour}
\label{fig:user_prompt_query_reformulation_behavior}
\end{figure*}

\subsection{Claim Extraction \& NLI Evaluations}
\label{sec:prompts_claim_nli}
To evaluate the grounding and factuality of model responses, we first extract atomic claims, then determine whether each claim is entailed by evidence from cited or retrieved URLs using NLI, and finally assess the factuality of each claim. The corresponding system and user prompts for each stage are shown below. Specifically, the prompts for claim extraction are shown in Figs.~\ref{fig:system_prompt_claimextraction} and \ref{fig:user_prompt_claimextraction}, respectively.  The system and user prompts for NLI-based grounding evaluation are shown in Figs.~\ref{fig:system_prompt_response-source-nli} and \ref{fig:user_prompt_response-source-nli}. Finally, the system and user prompts for factuality evaluation are shown in Figs.~\ref{fig:system_prompt_claim_Factuality_Eval} and \ref{fig:user_prompt_claim_Factuality_Eval}.
\begin{figure*}[t]
\centering
\begin{tcolorbox}[colback=gray!5, width=\textwidth, colframe=gray!80, title=System Prompt for Claim Extraction]
\begin{verbatim}
You are an expert claim extraction system.

Your task is to identify and extract claims from text.

Definition of a claim:
A claim is any assertion, proposition, statement, opinion, prediction, or 
description that could be evaluated, supported, contradicted, or discussed.

Rules:
- Extract all meaningful claims expressed in the text.
- Rewrite claims as standalone declarative sentences.
- Resolve pronouns and references where possible.
- Split compound sentences into atomic claims whenever appropriate.
- Preserve the original meaning, including:
  - negation
  - modality
  - uncertainty
  - comparisons
  - quantities
  - temporal information
- Do not infer unstated information.
- Avoid duplicate claims.
- Keep claims concise and self-contained.

Output requirements:
- Return ONLY a valid JSON array of strings.
- Do not include explanations or additional text.
\end{verbatim}
\end{tcolorbox}
\caption{System prompt for claim extraction}
\label{fig:system_prompt_claimextraction}
\end{figure*}

\begin{figure*}[t]
\centering
\begin{tcolorbox}[colback=gray!5, width=\textwidth, colframe=gray!80, title=User Prompt for Claim Extraction]
\begin{verbatim}
Extract all claims from the following text.

Text:
{text}
\end{verbatim}
\end{tcolorbox}
\caption{User prompt for claim extraction}
\label{fig:user_prompt_claimextraction}
\end{figure*}

\begin{figure*}[t]
\centering
\begin{tcolorbox}[colback=gray!5, width=\textwidth, colframe=gray!80, title=System Prompt for Response-Source NLI Evaluation]
\begin{verbatim}
You are an NLI (Natural Language Inference) judge.

Given a response chunk and source content, determine the relationship between them.

Labels:
- entailment: the source content supports or expresses the same meaning as the 
response chunk
- contradiction: the source content conflicts with the response chunk on a 
meaningful point
- neutral: the source content does not provide enough information to support or 
contradict the response chunk

Rules:
- Treat the response as a chunk or partial segment, not necessarily a complete 
standalone answer.
- Evaluate only the claims explicitly present in the response chunk.
- Use only the provided source content.
- Do not use external knowledge.
- Base your decision on semantic meaning, not exact wording.
- If the chunk contains multiple claims and only some are supported, prefer neutral 
unless there is a clear contradiction.
- Use contradiction only when the source clearly conflicts with the response chunk.
- Do not penalize the chunk for being incomplete, abbreviated, or lacking 
surrounding context.

Output JSON only:
{{
  "label": "entailment" | "contradiction" | "neutral",
  "reason": "<1-2 sentence explanation>",
  "score": 1-5
}}
\end{verbatim}
\end{tcolorbox}
\caption{System prompt for response-source NLI evaluation}
\label{fig:system_prompt_response-source-nli}
\end{figure*}

\begin{figure*}[t]
\centering
\begin{tcolorbox}[colback=gray!5, width=\textwidth, colframe=gray!80, title=User Prompt for Response-Source NLI Evaluation]
\begin{verbatim}
Response Chunk:
{response_text}

Source Content:
{source}

Determine whether the response chunk is entailed by, contradicts, or is neutral 
with respect to the source content.

Return ONLY valid JSON:
{{
  "label": "entailment|contradiction|neutral",
  "reason": "<string>",
  "score": <integer 1-5>
}}
\end{verbatim}
\end{tcolorbox}
\caption{User prompt for response-source NLI evaluation}
\label{fig:user_prompt_response-source-nli}
\end{figure*}

\begin{figure*}[t]
\centering
\begin{tcolorbox}[colback=gray!5, width=\textwidth, colframe=gray!80, title=System Prompt for Claim Factuality Evaluation]
\begin{verbatim}
You are an evaluator assessing the factual correctness of individual factual claims
made in an AI-generated response to a user query.

Evaluate each factual claim:
- Is the claim factually correct and free from hallucinations or false information?
- Is the claim up-to-date and not outdated when recency matters?

Return JSON:

{{
"score": 1-5,
"reasoning": "<1-2 sentence explanation>"
}}

Scoring guide:
1 = Clearly incorrect or hallucinated
2 = More incorrect than correct; contains significant factual errors
3 = Mixed or uncertain accuracy; contains both correct and incorrect aspects
4 = Mostly correct; minor inaccuracies or slightly outdated details
5 = Fully correct, precise, and up-to-date

Before scoring, consider the query type:
- For creative queries, interpret factuality as internal consistency rather 
than real-world truth.
\end{verbatim}
\end{tcolorbox}
\caption{System prompt for claim factuality evaluation}
\label{fig:system_prompt_claim_Factuality_Eval}
\end{figure*}

\begin{figure*}[t]
\centering
\begin{tcolorbox}[colback=gray!5, width=\textwidth, colframe=gray!80, title=User Prompt for Claim Factuality Evaluation]
\begin{verbatim}
Evaluate the following.

User Query:
{user_query}

Claim from AI Response:
{claim}

Return ONLY valid JSON in this exact format:
{{
"score": 1-5,
"reasoning": "<1-2 sentence explanation>"
}}

Rules:
- Do not include any text outside the JSON
- Do not add explanations before or after
- Ensure the JSON is valid
\end{verbatim}
\end{tcolorbox}
\caption{User prompt for claim factuality evaluation}
\label{fig:user_prompt_claim_Factuality_Eval}
\end{figure*}

%% file: disclaimer.tex
\section{Disclaimer to Annotators}
\label{app:disclaimer}
\begin{figure}
    \centering
    \includegraphics[width=1\linewidth]{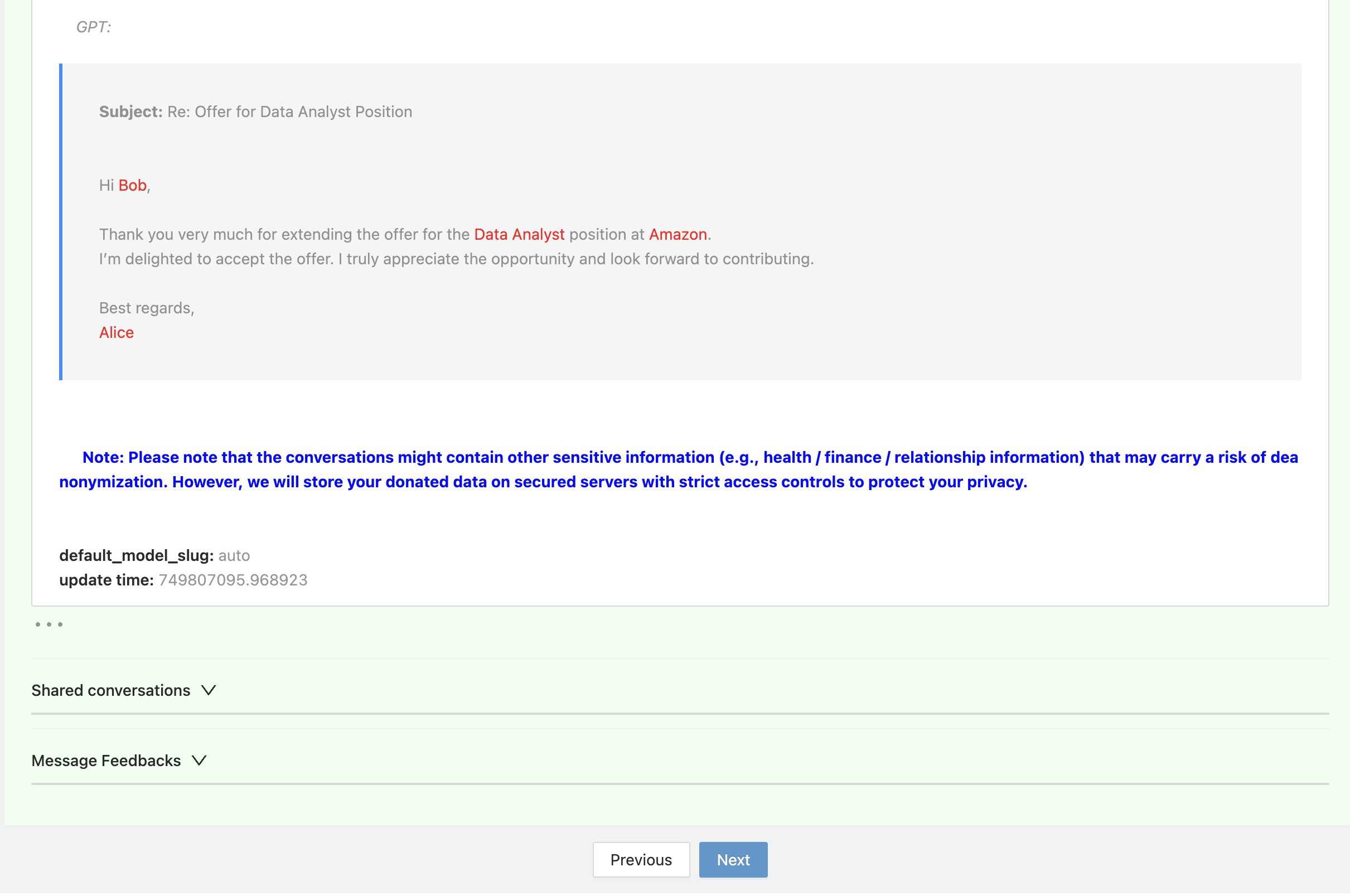}
    \caption{Disclaimer given to human annotators.}
    \label{fig:screenshot}
\end{figure}

Figure~\ref{fig:screenshot} shows the screenshot of a disclaimer given to human annotators.

